\documentclass{article} %
\usepackage{iclr2025_conference,times}

\usepackage{amsmath,amsfonts,bm}

\def\eqref#1{equation~\ref{#1}}

\def\1{\bm{1}}

\DeclareMathAlphabet{\mathsfit}{\encodingdefault}{\sfdefault}{m}{sl}
\SetMathAlphabet{\mathsfit}{bold}{\encodingdefault}{\sfdefault}{bx}{n}

\usepackage[pagebackref=true,breaklinks=true,letterpaper=true,colorlinks,bookmarks=false,citecolor=cyan]{hyperref}
\usepackage{url}

\usepackage{subfiles} %
\usepackage{makecell} %
\usepackage{multirow} %
\usepackage{subcaption}
\usepackage{color,xcolor}
\usepackage{graphicx}
\usepackage{booktabs}
\usepackage{authblk}
\usepackage{pifont}%

\newcommand{\name}{JudgeLM}

\newcommand{\add}[1]{\textcolor{black}{#1}}
\newcommand{\authorskip}{\hspace{12mm}}
\newcommand{\revision}[1]{\textcolor{black}{{#1}}}

\title{\name{}: Fine-tuned Large Language Models are Scalable Judges}

\newcommand{\myparagraph}[1]{{\bf #1}}

\author{
Lianghui Zhu$^{1,2}$ \thanks{This work was done when Lianghui Zhu was an intern at Beijing Academy of Artificial Intelligence. $^{\dagger}$Corresponding authors: \url{xgwang@hust.edu.cn} and \url{wangxinlong@baai.ac.cn}.
}
\authorskip Xinggang Wang$^{1\dagger}$ 
\authorskip Xinlong Wang$^{2\dagger}$   \\
{
\fontsize{10.4pt}{9.84pt}\selectfont
\textsuperscript{1} School of EIC, Huazhong University of Science \& Technology \\
\textsuperscript{2} Beijing Academy of Artificial Intelligence \\
}
{
\fontsize{9.4pt}{9.84pt}\selectfont 
\vspace{0.2cm}
Code \& Models: {\url{https://github.com/baaivision/JudgeLM}}
}
}

\iclrfinalcopy %
\begin{document}

\maketitle

\begin{abstract}
Evaluating Large Language Models (LLMs) in open-ended scenarios is challenging because existing benchmarks and metrics can not measure them comprehensively.
To address this problem, we propose to fine-tune L\textbf{LM}s as scalable \textbf{judge}s (\name) to evaluate LLMs efficiently and effectively in open-ended benchmarks. 
We first propose a comprehensive, large-scale, high-quality dataset containing task seeds, LLMs-generated answers, and GPT-4-generated judgments for fine-tuning high-performance judges, as well as a new benchmark for evaluating the judges. 
We train \name{} at different scales from 7B, 13B, to 33B parameters, and conduct a systematic analysis of its capabilities and behaviors. 
We then analyze the key biases in fine-tuning LLM as a judge and consider them as position bias, knowledge bias, and format bias.
To address these issues, \name{} introduces a bag of techniques including swap augmentation, reference support, and reference drop, which clearly enhance the judge's performance. 
\name{} obtains the state-of-the-art judge performance on both the existing PandaLM benchmark and our proposed new benchmark.
Our \name{} is efficient and the \name-7B only needs 3 minutes to judge 5K samples with 8 A100 GPUs.
\name{} obtains high agreement with the teacher judge, achieving an agreement exceeding 90\% that even surpasses human-to-human agreement\footnote{As a reference, the max agreement among humans in MT-bench~\citep{zheng2023chatbot-arena} is 82\%.}. 
\name{} also demonstrates extended capabilities in being judges of the single answer, multimodal models, multiple answers, multi-turn chat, etc. 
\end{abstract}

\section{Introduction}
\label{sec:intro}

Recent advancements in large language models (LLMs) have fostered significant interest due to their remarkable performance in following instructions and their broad capabilities in dealing with open-ended scenarios. 
Based on the open-source LLMs, including OPT~\citep{zhang2022opt}, Flan-T5~\citep{chung2022flant5}, LLaMA~\citep{touvron2023llama}, and Pythia~\citep{biderman2023pythia}, researchers propose numerous methods to align these models with human preferences through instruction fine-tuning. These aligned LLMs demonstrate enhanced abilities in comprehending human instructions and generating more coherent responses. Nonetheless, existing benchmarks~\citep{hendrycks2020mmlu, liang2022helm} and traditional metrics~\citep{lin2004rouge, papineni2002bleu, zhang2019bertscore, sellam2020bleurt, yuan2021bartscore} do not adequately estimate the capabilities of LLMs in open-ended scenarios. Therefore, a new benchmark method that could evaluate LLMs comprehensively in open-ended tasks is needed.

Concurrent works are making efforts to explore various methods for evaluating the performance of LLM. The arena-format~\citep{zheng2023chatbot-arena} methods leverage crowdsourced platforms to extract anonymous LLM competition results. While evaluations by humans are trustworthy, they are also time-consuming and financially demanding. Some approaches~\citep{chiang2023vicuna} utilize GPT-4 as a judge. Nevertheless, these methods grapple with challenges of potential data exposure and volatile API model transitions, potentially compromising the judge's reproducibility. PandaLM~\citep{wang2023pandalm} attempts to fine-tune open-source LLMs for evaluating answers. However, limitations stemming from the 
training data quality, and inherent LLM biases, undermine the effectiveness of such fine-tuned models in the role of a judge.

\vspace{-0.05in}

\begin{figure}[ht]
    \vspace{-1.0em}
    \begin{center}
    \begin{subfigure}{1.0 \linewidth}
        \includegraphics[width=1.0\linewidth]{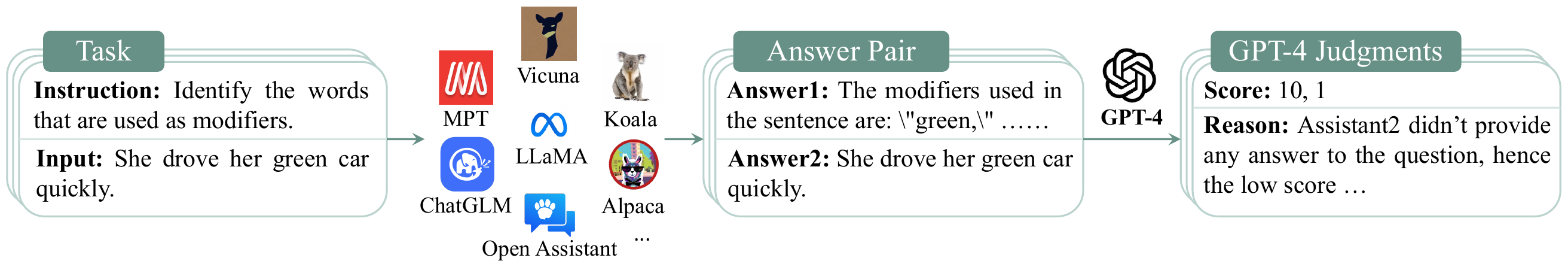}

        \vspace{-0.10in}
        \caption{Data generation pipeline of our \name{}. We first collect 105K seed tasks as questions. Then, we extract answers from 11 LLMs and randomly sample a pair of answers from the answer set. Last, we input the tasks, the sampled answer pairs, and optionally reference answers to GPT-4, which generates scores and detailed reasons as a judge teacher.}
        \label{fig: data_gen}
    \end{subfigure}
    \hfill
    \begin{subfigure}{1.0 \linewidth}
        \includegraphics[width=1.0\linewidth]{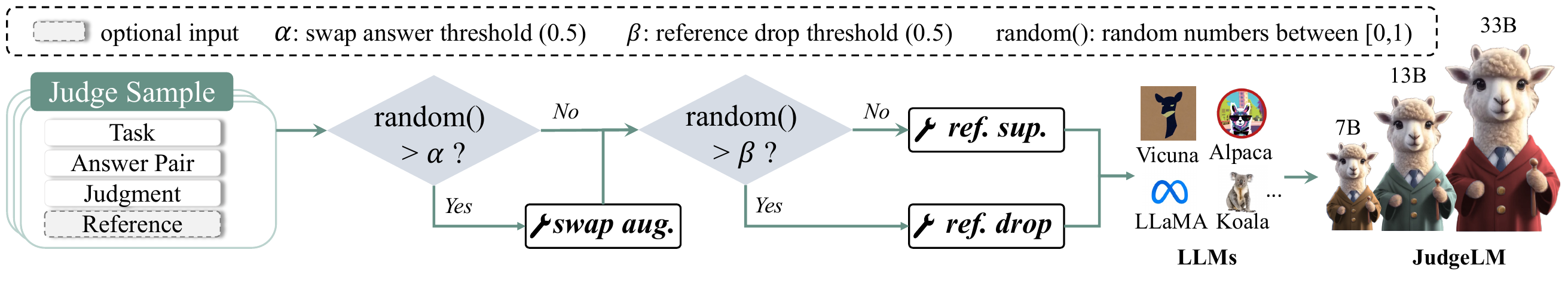}

        \vspace{-0.12in}
        \caption{An illustration of the \name{}'s fine-tuning and various functions. We use generated judge samples to fine-tune LLMs as scalable judges. When fine-tuning LLMs as judges, we also propose swap augmentation, reference support, and reference drop to address the position bias, knowledge bias, and format bias, respectively. }
        \label{fig: train_and_serving}
    \end{subfigure}
    \hfill
    \begin{subfigure}{1.0 \linewidth}
        \includegraphics[width=1.0\linewidth]{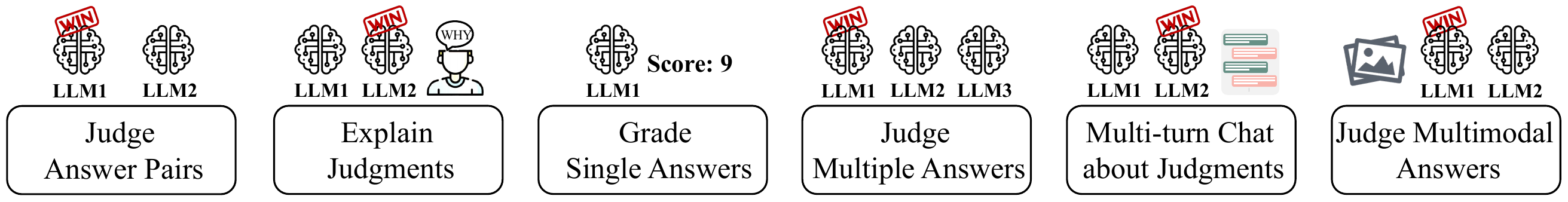}

        \vspace{-0.10in}
        \caption{An illustration of various functions of our \name{}.}
        \label{fig: functions}
    \end{subfigure}
    \end{center}
    \label{fig: main_figure}
    \vspace{-0.15in}
    \caption{An overview of our scalable \name{} including data generation, fine-tuning, and various functions.}
\end{figure}

In this paper, we propose to evaluate LLMs through fine-tuned open-source L\textbf{LM}s, which serve as scalable \textbf{judge}s (\name{}) achieving satisfactory agreement with the teacher judge. Our methodology incorporates scalable judges as evaluators in open-ended tasks, coupled with a high-quality dataset conducive to both training and evaluating the judge models. Within our framework, we adapt open-source LLMs to serve as judges and analyze their scaling ability in relation to model size (ranging from 7B to 33B) and volume of training data (extending from 3.5K to 100K). Our curated dataset comprises 105K seed questions, LLM answer pairs, and judgments from the teacher judge, GPT-4, as shown in Fig.~\ref{fig: data_gen}. Note that we generated two judgments for each seed task with and without reference answers. This dataset is partitioned, with 100K seed questions allocated for training (2 $\times$ larger than PandaLM) and the remainder for validation (29 $\times$ larger than PandaLM).

Utilizing LLMs as judges inevitably introduces biases such as position bias (favoring answers in specific positions), knowledge bias (over-reliance on pre-trained knowledge), and format bias (optimal performance only under specific prompt formats) as shown in Fig.~\ref{fig: position_bias}, ~\ref{fig: knowledge_bias}, ~\ref{fig: mismatch_wo}, ~\ref{fig: mismatch_w}. 
When fine-tuning is not possible, GPT-4-API-based judge~\citep{zheng2023chatbot-arena} tries to alleviate this by well-designed prompt methods, i.e., Chain-of-thought, few-shot judge, and judging multiple times with different positions. 
JudgeLM presents a new way that can address these biases in the fine-tuning stage, skipping the complicated prompt methods and multi-turn API calling.
Moreover, our \name{} system presents extended capabilities as shown in Fig.~\ref{fig: train_and_serving}, including grading single answers, judging multiple answers, judging multimodal models, multi-turn chat, etc.

In contrast to arena-format methods, our approach is rapid and has a low cost. For instance, \name-7B requires only 8 A100 GPUs and can evaluate 5000 response pairs in just 3 minutes. In comparison to closed-source LLM judges, \name{} ensures reproducibility and protects user privacy. When compared to concurrent open-source LLM judges, our system explores both the scaling ability and biases in LLM fine-tuning. Furthermore, \name{} dataset stands as the most diverse and high-quality one, significantly benefitting subsequent research in judge model investigations.

Our main contributions can be summarized as follows:
\begin{itemize}
    \item We introduce a high-quality, large-scale dataset for judge models, enriched with diverse seed tasks, LLMs-generated answers, and detailed judgments from GPT-4, laying the foundation for future LLMs evaluating research.
    \item We propose \name{}, a scalable language model judge, designed for evaluating LLMs in open-ended scenarios. It achieves an agreement exceeding 90\% that surpasses the human-to-human agreement. 
    Our \name{} can also generalize to many extended tasks.
    \item We analyze the biases inherent to \textbf{LLM judge fine-tuning} and introduce a series of methods to address them. Our methods significantly improve the consistency of the model in different cases, making the \name{} more reliable and flexible.
\end{itemize}

\section{Related Work}
\label{sec:relatedw}
\subsection{Instruction Fine-tuning of Large Language Models}
With the development of large language models (LLMs), researchers find that fine-tuning pre-trained LLMs such as GPT-3~\citep{brown2020gpt3}, T5~\citep{raffel2020T5}, OPT~\citep{zhang2022opt}, and PaLM~\citep{chowdhery2022palm} enable LLMs to follow human instructions and help with open-ended tasks. The instruction fine-tuned LLMs such as InstructGPT~\citep{ouyang2022instructgpt}, ChatGPT~\citep{openai2022chatgpt}, FLAN-T5~\citep{chung2022flant5}, FLAN-PaLM~\citep{chung2022flant5}, OPT-IML~\citep{iyer2022opt-iml}, and GPT-4~\citep{openai2023gpt4} exhibit stronger ability in zero-shot or few-shot tasks than their base models. After Meta released the powerful open-source LLM LLaMA~\citep{touvron2023llama} and LLaMA2~\citep{touvron2023llama2}, lots of instruction fine-tuning works based on LLaMA or LLaMA2 were proposed in the natural language generation or multimodal generation domain, such as Alpaca, Vicuna~\citep{chiang2023vicuna}, OpenFlamingo~\citep{awadalla2023openflamingo}, LLaMA-Adapter~\citep{zhang2023llama-adapter}, and Emu~\citep{sun2023emu}. Our \name{} also belongs to the LLaMA family and takes the Vicuna series as base models.
Our \name{} follows the instruction fine-tuning manner to create LLM judges and proposes to model the judgment-generation task as ``grading, judging, and reasoning''. We further collect a high-quality, large-scale dataset for research in judging the performance of LLMs.

\subsection{Evaluation of Large Language Models}
As many open-source large language models (LLMs) and their fine-tuned variants are proposed and present remarkable performance on various tasks, evaluating the capabilities of LLMs becomes a popular and challenging task. To address this problem, Chatbot Arena~\citep{zheng2023chatbot-arena} aims to build a crowdsourced platform that ranks the LLMs through pairwise comparison and Elo rating. The crowdsourced way to evaluate LLMs has more reliable results but faces high costs and low efficiency. 
Vicuna~\citep{chiang2023vicuna} uses GPT-4 as a judge to select the better answer. Although the GPT-4-based method can judge LLMs like a human expert, the API-based methods have potential risks of data leakage and unstable performance. Zeno Build~\citep{Cabrera2023zeno-chatbot} proposes to evaluate LLMs at a customer service dataset, but using traditional metrics such as ChrF~\citep{popovic-2015-chrf} and BERTScore~\citep{zhang2019bertscore} can not fully evaluate the answers of LLMs in open-ended tasks. Besides, PandaLM~\citep{wang2023pandalm} and Auto-J~\cite{li2023autoj} developed judge models based on LLaMA~\citep{touvron2023llama} or LLaMA2~\citep{touvron2023llama2} to compare answers produced by LLMs. When serving as judges, PandaLM achieves an accuracy close to ChatGPT but \revision{ignoring the inherent LLM biases} limits its performance further.
Our \name{} contains scalable judges from 7B-parameter to 33B-parameter and achieves state-of-the-art performance in both PandaLM and our benchmarks. Furthermore, researchers can use the proposed \name{} locally which ensures reproducibility and data security.

\section{Dataset}
\label{sec:data}

High-quality, large-scale datasets are crucial for effectively fine-tuning large language models (LLMs) to act as evaluative judges. However, the concurrent datasets, such as the one by PandaLM~\citep{wang2023pandalm}, present limitations in terms of diversity and the granularity of judgment criteria. To address this, we introduce a novel dataset replete with a rich variety of seed tasks, comprehensive answers from modern LLMs, answers' grades from the teacher judge, and detailed reasons for judgments. Section~\ref{subsec: data_gen} elucidates the data generation process, while Section~\ref{subsec: train_and_eval} delineates the methods adopted for training and evaluation using our dataset.

\subsection{Data Generation}
\label{subsec: data_gen}

The primary objective of our data generation is to create a large-scale and diversified dataset that maximizes the evaluative capabilities of judge models. 
We sample 105K instruction seed tasks from a large-scale set that contains Alpaca-GPT4~\citep{peng2023alpaca-gpt4}, Dolly-15K~\citep{DatabricksBlog2023DollyV2}, GPT4All-LAION~\citep{gpt4all}, and ShareGPT.
To enhance the heterogeneity of the dataset, answers are collated from 11 leading open-source LLMs including, but not limited to, LLaMA~\citep{touvron2023llama}, Alpaca, and Vicuna~\citep{chiang2023vicuna}. Following this, we amalgamate LLM-generated answers with the reference answer to create answer sets. Pairs are randomly selected from the sets, upon which, fine-grained scores and detailed reasons are assigned by the advanced teacher model, GPT-4. To ensure robust and comprehensive judgments, we utilize detailed templates as demonstrated in Fig.~\ref{fig: temp_wo_ref}. Additionally, to allow the model to judge with reference answers, the reference-inclusive template is employed as Fig.~\ref{fig: temp_w_ref}. This encourages the model to integrate external knowledge during the evaluative process.
Please note that all samples in the JudgeLM $val$ set are further checked and re-annotated by authors to ensure alignment with human preference.

\begin{figure}[t]
    \centering
        \includegraphics[width=.88\linewidth]{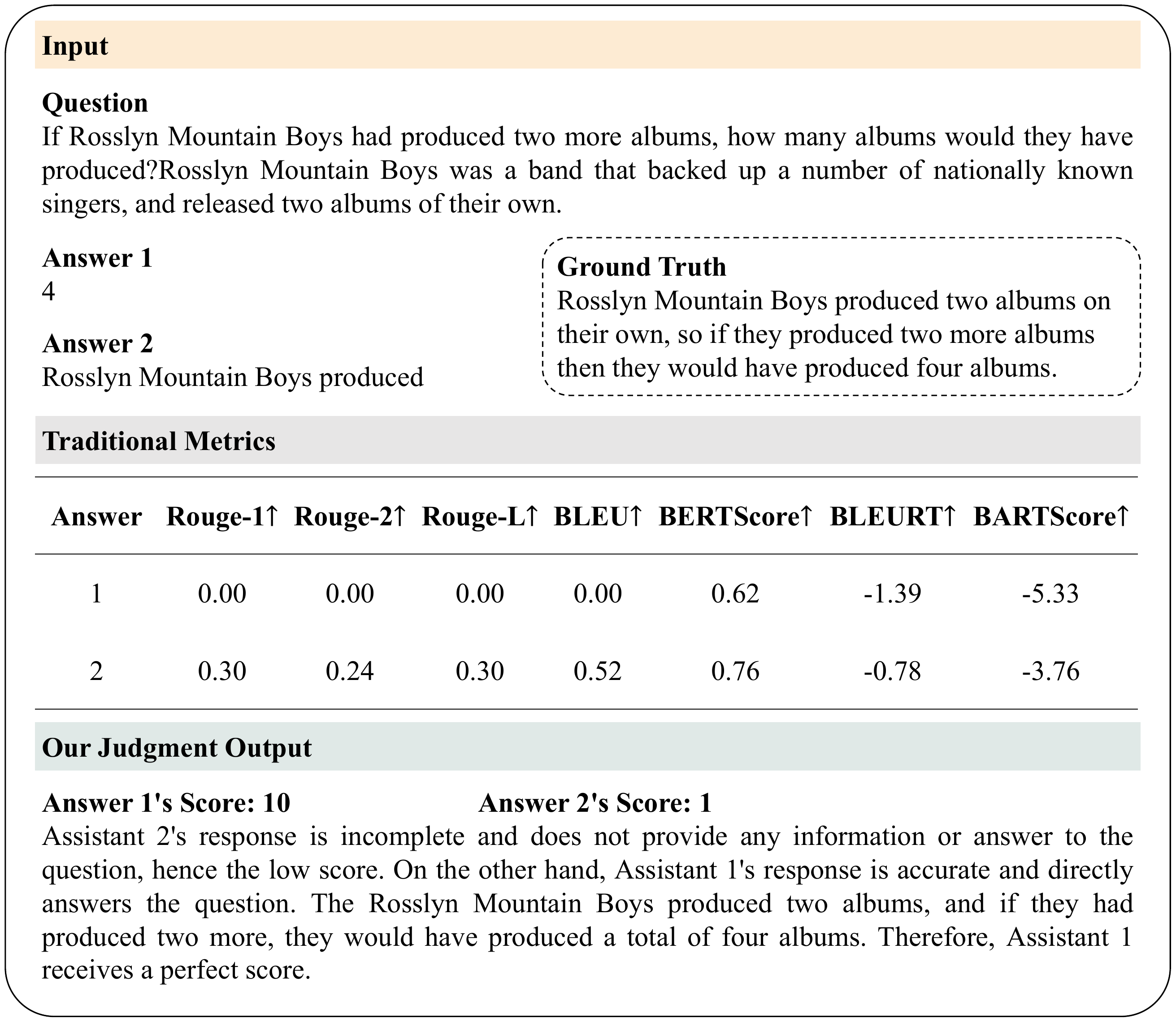}
    \vspace{-0.08in}
    \caption{The input and output of our \name{} data sample. 
    In open-ended scenarios, traditional metrics can not judge answers accurately by comparing the answers with ground truth.
    However, the LLM judges can understand the questions and answers and give accurate scores and reasons.}
    \label{fig: model_io}
    \vspace{-0.2in}
\end{figure}

\subsection{Training and Evaluating}
\label{subsec: train_and_eval}

To better utilize our dataset to train and evaluate the judge models, we partition it into a training split and a validation split. The training set contains 100K judge samples, while the validation set has 5K. We then introduce the way we use this dataset to train and evaluate, respectively.

\myparagraph{Training.} 
The training process of \name{} adheres to the instruction fine-tuning paradigm. As illustrated in Fig.~\ref{fig: model_io}, the model is fed a question alongside a pair of answers, and an optional reference answer, yielding outputs comprising scores and detailed reasons. It is imperative to note the significance of a detailed crafted prompt template to harness the full potential of \name{}'s instruction-following ability. Distinct input templates cater to scenarios with and without references, as depicted in Fig.~\ref{fig: temp_wo_ref} and Fig.~\ref{fig: temp_w_ref} respectively.

To further analyze the scaling ability of \name{}, we fine-tune \name{} with sizes of 7B, 13B, and 33B parameters. The specific hyperparameters are enumerated in Table~\ref{tab: ft_cfg}. As for the scaling analysis for dataset size, we also fine-tune \name{} on varying data scales from 3.5K to 100K samples. \name{} demonstrates scaling ability both in terms of model size and data volume.

\myparagraph{Evaluating.}
For the judge's result, we model it as ``grading, judging, and reasoning''. The judge model first generates scores for answer pairs. Subsequently, we can get the judge result from three situations: ``Answer 1 wins'' if the answer 1's score is higher than the answer 2's, ``Answer 2 wins'' if the answer 2's score is higher, or ``Tie'' if the scores of two answers are the same. Last, the model generates detailed reasons if needed. The advantage of this modeling is that the judge model just needs little time to grade and judge, and generates time-consuming reasoning optionally.

For the metrics, we employ the objective metrics and reliability metrics to evaluate the judge models comprehensively.
For the objective metrics, we compute the agreement, precision, recall, and F1-score between the model's judge results and those of the teacher. This provides insights into the alignment of judge models with established benchmarks, such as GPT-4 or human experts. 
As for reliability metrics, we first compare the results before and after swapping LLM answers. Then we calculate the self-consistency to measure the judge model's reliability. Last, we further calculate the metrics like ``bias toward 1st'', ``bias toward 2nd'', and ``delta bias'' to get insights from specific position biases and their variance.

\vspace{-.1 in}

\section{Inherent Bias}
\label{sec:bias}
In this paper, we also study the inherent biases that influence the reliability of fine-tuned LLM judges through reliability metrics and visualizations.

\myparagraph{Position Bias.}
Position bias means that the LLM judges prefer answers in a certain position and it widely exists in natural language processing tasks~\citep{ko2020look1st, wang2018position_bias} and decision-making of humans~\citep{blunch1984human_position_bias, raghubir2006human_decision_making}. The powerful LLMs, ChatGPT and GPT-4, also face this challenge when working as judges~\citep{wang2023pandalm, zheng2023chatbot-arena, li2023split}. As the qualitative and quantitative results shown in Fig.~\ref{fig: position_bias} and Table~\ref{tab: swap_aug}, \name{} also faces the position bias and prefers the first answer when swapping the positions of answers.

\myparagraph{Knowledge Bias.}
Knowledge bias arises when the pre-trained data lacks the knowledge of some seed tasks or induces possibly undesirable knowledge~\citep{ko2020look1st, zheng2023chatbot-arena} that could degenerate the generative capabilities of LLMs.
Fig.~\ref{fig: knowledge_bias} provides an example that LLM judges can not give correct judgments to open-ended tasks if they lack related truth. 

\myparagraph{Format Bias.}
Researchers expect that the judge model can make judgments based on pre-trained knowledge when the reference is not available and can make judgments following the reference when it is available.
However, our experiments revealed that judge models fine-tuned without reference perform poorly in judging with reference, and vice versa, as shown in Fig.~\ref{fig: mismatch_wo}, Fig.~\ref{fig: mismatch_w}, and Table~\ref{tab: ref_drop}.
We hypothesize fine-tuning with references encourages the judge model to make judgments based on external knowledge and fine-tuning without references pushes the judge model to make judgments through its pre-trained knowledge.
We name the situation that a judge fine-tuned without reference but validated with reference as a mismatched format, and vice versa.
Such a format bias limits the further generalization of the judge model in other domains.
\vspace{-.1 in}

\section{Method}
\label{sec:methods}

In evaluating LLM-generated answers for a seed question, the LLM judge aims to determine the superior answer from a pair of candidates. Motivated by recent methods~\citep{touvron2023llama, chiang2023vicuna, ouyang2022instructgpt}, we present \name{}, a scalable judge model, and address inherent biases in such models. Our methodology is depicted in Fig.~\ref{fig: train_and_serving}. The subsequent sections provide a detailed breakdown of our approach.

\subsection{Swap Augmentation}
\label{subsec: swap_aug}

MT-bench~\citep{zheng2023chatbot-arena} and PandaLM~\citep{wang2023pandalm} alleviate the position bias by judging twice with original and reverse order. These methods regard the result as a tie if the judgments are not the same. This kind of method ignoring the inherent position bias and casting double time to evaluate, can be regarded as a compromise and does not improve the reliability of LLM judges.

Intuitively, swapping the positions at the fine-tuning stage could push the judge model to pay more attention to the contents of answers rather than positions. Leveraging our structured judge data, we can easily swap the positions of answers to generate a new input sample. Correspondingly, we also swap the scores and question indexes of the judgment from the teacher (i.e., GPT4) to get the new ground truth. As shown in Fig.~\ref{fig: swap_aug_io}, the augmented judge sample keeps the same results but exchanges the positions of answers. Overall, it is simple but effective to augment the training data and address position bias. The \name{}-with-swap-augmentation can give good judgment to the same judge sample as shown in Fig.~\ref{fig: swap_aug}. 

\subsection{Reference Support}
\label{subsec: ref_sup}

Introducing external knowledge in the fine-tuning stage is an intuitive way to make up for the lack of related pre-trained knowledge. To do so, we propose the reference support method to teach the model to judge with the help of reference answers. Following \cite{zheng2023chatbot-arena}, we collect reference answers for all judge samples and re-generate reference-guided judgments by GPT-4. Please note that GPT-4 also gives different scores and judgments for most judge samples with or without references. This proves that the differences between pre-trained knowledge and reference answers greatly impact judgments. As shown in Fig.~\ref{fig: reference_sup}, the \name{} with reference support can avoid factual errors and give reliable judgments. Furthermore, introducing reference support to LLM judges can simply insert judge preferences. \name{} with reference support training can flexibly set reference answers with different preferences for different scenarios and needs. As shown in Fig.~\ref{fig: change_reference}, changing reference answers does not need extra training and makes \name{} more flexible to different preferences.

\subsection{Reference Drop}
\label{subsec: ref_drop}

To address the format bias, we introduce a method, named reference drop, in which we randomly drop the training sample with reference and use the corresponding sample without reference. 
As shown in Fig.~\ref{fig: ref_drop}, judge models with reference drop can alleviate the overfitting for fine-tuning formats and 
make judgments based on external reference or pre-trained knowledge when given reference or not, respectively.
Furthermore, the reference drop method also makes the judge model easy to use and decreases the cost of fitting into different formats.

\section{Experiment}
\label{sec:exps}

We study the performance of \name{} as follows: Section~\ref{subsec: main_res} presents the main results of \name{} comparing with concurrent methods, Section~\ref{subsec: scaling_ana} analyzes the scaling ability of \name{} from both model sizes and data scales, and Section~\ref{subsec: ablation} shows ablation studies of proposed methods in detail. Detailed settings are shown in Section~\ref{subsec: setting_more}.

\subsection{Main Results}
\label{subsec: main_res}

\begin{table*}[t!]
\caption{Main results for our \name{} and concurrent methods on our $val$ set, which uses GPT-4 annotation results as ground truth.}
\vspace{-0.16in}
\begin{center}
\begin{tabular}{lccccccc}
\toprule
Methods & \begin{tabular}[c]{@{}c@{}}Agreement ↑ \\  (w/ GPT-4) \end{tabular} & \begin{tabular}[c]{@{}c@{}}Precision ↑\\  (w/ GPT-4) \end{tabular} & \begin{tabular}[c]{@{}c@{}}Recall ↑\\  (w/ GPT-4) \end{tabular} & \begin{tabular}[c]{@{}c@{}}F1 ↑\\  (w/ GPT-4) \end{tabular} & \begin{tabular}[c]{@{}c@{}}Consistency ↑\\  (w/ swap.) \end{tabular} \\
\midrule
\multicolumn{6}{l}{\textbf{\textit{Judge w/o reference.}}} \\
GPT-3.5 & 73.83 & 70.70 & 52.80 & 52.85 & 68.89 \\
Vicuna-13B & - & - & - & - & - \\
PandaLM-7B & 68.61 & 40.75 & 38.82 & 39.41 & 74.78 \\ 
Auto-J-13B & 74.86 & 61.65  & 57.53 & 58.14 & 84.34 \\
\midrule
\multicolumn{6}{l}{\textbf{\textit{Judge w/o reference (Ours).}}} \\
\name-7B & 81.11 & 69.67 & 78.39 & 72.21 & 83.57 \\
\name-13B & 84.33 & 73.69 & 80.51 & 76.17 & 85.01 \\
\name-33B & 89.03 & 80.97 & 84.76 & 82.64  & 91.36 \\
\midrule
\multicolumn{6}{l}{\textbf{\textit{Judge w/ reference.}}} \\
GPT-3.5 & 71.46 & 56.86 & 51.12 & 51.14 & 62.94 \\
Vicuna-13B & - & - & - & - & - \\
PandaLM-7B & 63.77 & 39.79 & 34.82 & 35.18 & 55.39 \\ 
Auto-J-13B & 72.90 & 58.80  & 56.12 & 56.59 & 82.84 \\
InstructScore-7B & 55.80 & 58.74 & 56.84 & 53.72 & - \\
\midrule
\multicolumn{6}{l}{\textbf{\textit{Judge w/ reference (Ours).}}} \\
\name-7B & 84.08 & 75.92 & 82.55 & 78.28 & 84.46 \\ 
\name-13B & 85.47 & 77.71 & 82.90 & 79.77 & 87.23\\
\name-33B & 89.32 & 84.00 & 86.21 & 84.98 & 92.37 \\
\bottomrule
\end{tabular}
\end{center}
\label{tab: main_our}
\end{table*}

\begin{table*}[t!]
\vspace{-1.0em}
\caption{\name{} zero-shot evaluation results on PandaLM $test$ set, which uses human annotation results as ground truth. ``${}^*$'' means the results are reported in PandaLM~\citep{wang2023pandalm}}
\vspace{-0.16in}
\begin{center}
\begin{tabular}{lcccc}
\toprule
Methods & \begin{tabular}[c]{@{}c@{}}Agreement ↑ \\  (w/ Human) \end{tabular} & \begin{tabular}[c]{@{}c@{}}Precision ↑\\  (w/ Human) \end{tabular} & \begin{tabular}[c]{@{}c@{}}Recall ↑\\  (w/ Human) \end{tabular} & \begin{tabular}[c]{@{}c@{}}F1 ↑\\  (w/ Human) \end{tabular} \\
\midrule
\multicolumn{5}{l}{\textbf{\textit{zero-shot methods.}}} \\
GPT-3.5$^*$                        & 62.96 & 61.95 & 63.59 & 58.20  \\
GPT-4$^*$                          & 66.47 & 66.20  & 68.15 & 61.80  \\
\midrule
\multicolumn{5}{l}{\textbf{\textit{Fine-tuned on PandaLM $train$ set. }}} \\
PandaLM-7B$^*$                     & 59.26 & 57.28 & 59.23 & 54.56 \\
\midrule
\multicolumn{5}{l}{\textbf{\textit{Ours (zero-shot).}}} \\
JudgeLM-7B                & 65.07 & 66.89 & 71.95 & 61.92 \\
JudgeLM-13B               & 68.97 & 68.21 & 74.15 & 65.12 \\
JudgeLM-33B               & 75.18 & 69.30 & 74.93 & 69.73 \\
\bottomrule
\end{tabular}
\end{center}
\label{tab: main_panda}
\vspace{-0.25in}
\end{table*}

\myparagraph{Comparison on \name{} Benchmark.} 
We first evaluate the proposed \name{} on our $val$ set. Note that \name{} $val$ set is further checked and re-annotated by authors to ensure alignment with human preference. As shown in Table~\ref{tab: main_our}, we give the quantitative results of GPT-3.5, Vicuna-13B, PandaLM-7B, Auto-J-13B~\citep{li2023autoj}, InstructScore-7B~\citep{xu2023instructscore}, and our \name{} with three model sizes. Among them, GPT-3.5 is used in the form of APIs with the help of templates in Fig.~\ref{fig: temp_wo_ref} and Fig.~\ref{fig: temp_w_ref}. PandaLM-7B and Auto-J-13B are deployed with the released checkpoints and templates. These methods could be regarded as zero-shot methods because they are not fine-tuned by the \name{} dataset. 
On JudgeLM $val$ set, the vanilla Vicuna-13B fails 77\% of questions. Specifically, the vanilla Vicuna-13B can not even output a pair of scores in judgments in the failed cases. But the finetuned version, i.e., JudgeLM, would not fail any questions in the JudgeLM $val$ set. 
Our \name{}s are fine-tuned with proposed methods, i.e., swap augmentation, reference support, and reference drop. So, they can handle situations with or without references simultaneously.
It can be observed that our \name{}-7B outperforms PandaLM-7B, Auto-J, and InstructScore in all metrics, and even surpasses GPT-3.5. Furthermore, the proposed \name{}-33B exhibits the most powerful judge ability.

\begin{table*}[t!]
\caption{Efficiency comparison for our \name{} and PandaLM on our $val$ set. We use a machine with 8 Nvidia-A100 GPUs with 40G memory to evaluate their efficiency.}
\vspace{-0.16in}
\begin{center}
\begin{tabular}{lccccccr}
\toprule
Methods & model size & GPUs per model  & parallel judge? & generate reason? & total time \\
\midrule
PandaLM & 7B & 1 & \ding{55} & \ding{51} & 6 hrs 40 mins \\
\midrule
\multicolumn{6}{l}{\textbf{\textit{Ours.}}} \\
\add{\name{}} & \add{7B} & \add{1} & \add{\ding{55}} & \add{\ding{51}} & \add{6 hrs 40 mins} \\
\add{\name{}} & \add{7B} & \add{1} & \add{\ding{55}} & \add{\ding{55}} & \add{24 mins} \\
\name{} & 7B & 1 & \ding{51} & \ding{51} & 50 mins \\ %
\name{} & 7B & 1 & \ding{51} & \ding{55} & 3 mins \\ %
\name{} & 13B & 1 & \ding{51} & \ding{55} & 5 mins \\ %
\name{} & 33B & 2 & \ding{51} & \ding{55} & 15 mins \\ %
\bottomrule
\end{tabular}
\end{center}
\label{tab: efficiency}
\vspace{-0.15in}
\end{table*}

\begin{table*}[t!]
\caption{Performance analysis for the scaling \name{} on our $val$ set.}
\vspace{-0.16in}
\begin{center}
\begin{tabular}{lccccccc}
\toprule
\begin{tabular}[l]{@{}l@{}}Judge  Size \end{tabular} & \begin{tabular}[c]{@{}c@{}}Data  Scale \end{tabular} & \begin{tabular}[c]{@{}c@{}}Agreement ↑\\  (w/ GPT-4) \end{tabular}  & \begin{tabular}[c]{@{}c@{}}Consistency ↑\\  (w/ swap.) \end{tabular} & \begin{tabular}[c]{@{}c@{}}Bias ↓\\  toward 1st \end{tabular}  & \begin{tabular}[c]{@{}c@{}}Bias ↓\\  toward 2nd \end{tabular} & \multicolumn{1}{l}{Delta bias ↓} \\ %
\midrule
7B  & 3.5k & 75.87                         & 73.45                         & 19.83                         & 6.72                          & 13.11                         \\
7B  & 10k  & 78.89                         & 78.25                         & 17.30                         & 4.45                          & 12.85                         \\
7B  & 30k  & 81.43                         & 80.89                         & 14.54                         & 4.57                          & 9.97                          \\
7B  & 100k & 83.71                         & 82.62                         & 12.31                         & 5.07                          & 7.24                          \\
\midrule
13B & 3.5k & 80.61                         & 78.91                         & 14.68                         & 6.41                          & 8.27                          \\
13B & 10k  & 83.19                         & 81.90                         & 13.42                         & 4.68                          & 8.74                          \\
13B & 30k  & 84.39                         & 82.99                         & 11.96                         & 5.05                          & 6.91                          \\
13B & 100k &  85.87  & 83.01                         &  11.53  &  5.46   &  6.07   \\
\midrule
33B & 3.5k & 85.38                         & 85.16                         & 9.34                          & 5.50                          & 3.84                          \\
33B & 10k  & 87.49                         & 86.40                         & 8.32                          & 5.28                          & 3.04                          \\
33B & 30k  & 88.84                         & 87.34                         & 7.57                          & 5.09                          & 2.48                          \\
33B & 100k & 90.06                         & 87.93                         & 6.85                          & 5.22                          & 1.63                          \\
\bottomrule
\end{tabular}
\end{center}
\label{tab: scaling}
\vspace{-0.1in}
\end{table*}

\begin{table*}[t!]
\caption{Ablation study for the swap augmentation on our $val$ set.}
\vspace{-0.16in}
\begin{center}
\begin{tabular}{lccccccc}
\toprule
Methods & \begin{tabular}[c]{@{}c@{}}Agreement ↑\\  (w/ GPT-4) \end{tabular}  & \begin{tabular}[c]{@{}c@{}}Consistency ↑\\  (w/ swap.) \end{tabular} & \begin{tabular}[c]{@{}c@{}}Bias ↓\\  toward 1st \end{tabular}  & \begin{tabular}[c]{@{}c@{}}Bias ↓\\  toward 2nd \end{tabular} & \multicolumn{1}{l}{Delta Bias ↓} \\ %
\midrule
baseline & 75.87 & 73.45 & 19.83 & 6.72 & 13.11 \\ %
+ swap aug. & 76.51 & 78.89 & 15.34 & 5.77 & 9.57 \\ %
\bottomrule
\end{tabular}
\end{center}
\label{tab: swap_aug}
\vspace{-0.25in}
\end{table*}

\myparagraph{Comparison on Other Human Evaluation Benchmarks.}
We further evaluate our \name{} on other Human evaluation benchmarks, i.e., PandaLM $test$ set and human-annotated MM-Vet.
PandaLM's $train$ and $val$ sets are annotated by GPT-3.5 and humans, respectively.
Following the manner of the PandaLM $val$ set, we present the zero-shot results of \name{} in Table~\ref{tab: main_panda}. 
It can be observed that the \name{}-7B outperforms GPT-3.5 and PandaLM-7B. When compared with GPT-4, \name{}-7B has lower accuracy and higher Precision, Recall, and F1-score than GPT-4. Furthermore, \name{}-33B achieves higher results than GPT-4, which demonstrates that fine-tuned \name{} can outperform its teacher in this specific task. Besides, we also propose a human-annotated multimodal judging benchmark to evaluate our \name{} as shown in Table~\ref{tab: mmvet}.

\myparagraph{Efficiency comparison.}
To further compare the efficiency between our \name{} and PandaLM, we conduct experiments on our $val$ set to display the time cost using the same machine with 8 NVIDIA-A100 (40G) GPUs. As shown in Table~\ref{tab: efficiency}, we display the methods and model sizes in the first and second columns. The third column shows the needed GPUs for each judge model. The models with 7B or 13B parameters run on 1 A100 GPU with 40G memory while the 33B-parameter needs 2 GPUs. The fourth column shows whether the methods can judge answers in parallel. 
The fifth column indicates whether judge reasons are generated at runtime. The sixth column presents the total time cost.
We use PandaLM-7B and \name{}-7B as efficiency baselines, which do not use parallel judging and generate detailed reasons for all questions. 
Thanks to the modeling of JudgeLM, i.e., ``grading, judging, and reasoning", \name{} can skip the reasoning phase and only requires 24 minutes, which is 16.65 $\times$ faster than baselines. 
When we enable the engineering optimization of parallel judging, the \name{} can make full use of the 8 GPUs and cost only 50 minutes, which is 8 $\times$ faster than the baseline running on a single GPU.
When we enable parallel judging and skip the reasoning phase for \name{}, the \name{}-7B only consumes 3 minutes to judge 5000 response pairs, which is 133.3 $\times$ faster than baselines.
The largest judge model, \name{}-33B, can also complete the validation within 15 minutes.
\name{}'s high efficiency can significantly reduce the time spent on evaluating LLMs, allowing researchers and developers to boost the pace of advancements.

\subsection{Scaling Analysis of \name{}} %
\label{subsec: scaling_ana}

In this section, we analyze the scaling ability of the plain \name{} \add{(without the proposed methods)} on our $val$ set without reference as illustrated in Table~\ref{tab: scaling}. As we increase the model size and data scale, we can observe the metrics increase. It demonstrates that the proposed \name{} is scalable and can reach up to 90.06\% agreement and 87.93\% consistency with 33B-parameter and 100K fine-tuning data.

\begin{table*}[t]
\caption{Ablation study for the reference support and reference drop on our $val$ set.}
\vspace{-0.16in}
\begin{center}
\begin{tabular}{lccccccccc}
\toprule
Methods &  \begin{tabular}[c]{@{}c@{}}$ft$\\  w/ ref? \end{tabular} &  \begin{tabular}[c]{@{}c@{}}$val$\\  w/ ref? \end{tabular} & \begin{tabular}[c]{@{}c@{}}Agreement ↑\\  (w/ GPT-4) \end{tabular}  & \begin{tabular}[c]{@{}c@{}}Consistency ↑\\  (w/ swap.) \end{tabular} & \begin{tabular}[c]{@{}c@{}}Bias ↓\\  toward 1st \end{tabular}  & \begin{tabular}[c]{@{}c@{}}Bias ↓\\  toward 2nd \end{tabular} & \begin{tabular}[c]{@{}c@{}}Delta\\  Bias ↓ \end{tabular} \\
\midrule
\multicolumn{8}{l}{\textbf{\textit{matching format.}}} \\
baseline & \ding{55} & \ding{55} & 75.87 & 73.45 & 19.83 & 6.72 & 13.11 \\
baseline & \ding{51} & \ding{51} & 80.15 & 81.23 & 11.55 & 7.22 & 4.33 \\
\multicolumn{8}{l}{\textbf{\textit{mismatched format.}}} \\
baseline & \ding{55} & \ding{51} & 73.09 & 67.75 & 29.44 & 2.81 & 26.63 \\
baseline & \ding{51} & \ding{55} & 75.69 & 73.40 & 20.89 & 5.71 & 15.18 \\ %
\multicolumn{8}{l}{\textbf{\textit{w/ ref. drop.}}} \\
baseline & ref. drop & \ding{55} & 76.86 & 77.13 & 17.30 & 5.57 & 11.73 \\
baseline & ref. drop & \ding{51} & 80.35 & 81.24 & 11.48 & 7.28 & 4.20 \\ %
\bottomrule
\end{tabular}
\end{center}
\label{tab: ref_drop}
\vspace{-0.1in}
\end{table*}

\begin{table}[t!]
\centering
\caption{\add{Performance of JudgeLM-7B with explanation-first (CoT) or score-first (Ours) on JudgeLM $val$ set.}}
\vspace{-0.10in}
\begin{tabular}{lccccc}
\toprule
\add{Methods} & \begin{tabular}[c]{@{}c@{}}\add{Agreement ↑}\\  \add{(w/ GPT-4)} \end{tabular}  & \begin{tabular}[c]{@{}c@{}}\add{Consistency ↑}\\  \add{(w/ swap.)} \end{tabular} & \begin{tabular}[c]{@{}c@{}}\add{Bias ↓}\\  \add{toward 1st} \end{tabular}  & \begin{tabular}[c]{@{}c@{}}\add{Bias ↓}\\  \add{toward 2nd} \end{tabular} & \multicolumn{1}{l}{\add{Delta Bias ↓}} \\ %
\midrule
\add{score-first (Our)}       & \add{75.87}       & \add{73.45}         & \add{19.83}             & \add{6.72}              & \add{13.11}        \\ 
\add{explanation-first (CoT)} & \add{75.54}       & \add{74.39}         & \add{15.05}             & \add{10.56}             & \add{4.50}         \\ 
\bottomrule
\end{tabular}
\label{tab: cot}
\vspace{-0.1 in}
\end{table}

\subsection{Ablation Study}
\label{subsec: ablation}

In this section, we present the ablation studies of the proposed methods. For all ablation studies, we use \name{}-7B as the base model and 3.5K data for fine-tuning. Based on this baseline, we analyze the improvements brought by swap augmentation, reference support, and reference drop.

\myparagraph{Improvements of Swap Augmentation.} 
As shown in Table~\ref{tab: swap_aug}, swap augmentation can improve the baseline model comprehensively. It improves consistency by 5.44\%, which demonstrates that swap augmentation can reduce the influence of position bias and push the judge to pay more attention to the contents of answers.

\myparagraph{Improvements of Reference Support.}
As shown in the rows with the matching format of Table~\ref{tab: ref_drop}, \name{} fine-tuned with reference support exhibits superior performance on every metric. It demonstrates that the introduction of reference answers induces the judge to rely on external knowledge and addresses the limitation of pre-trained knowledge.

\myparagraph{Improvements of Reference Drop.}
As shown in Table~\ref{tab: ref_drop}, baselines can not reach satisfactory performance when facing mismatched formats. With the help of the reference drop, the \name{} can handle both the format with or without reference and achieve higher agreement and consistency. It demonstrates that reference drop can address the format bias and avoid the \name{} overfitting to a single format.

\myparagraph{Ablation of Judging Form}
We further evaluate the performance of JudgeLM-7B with explanation-first (Chain of Thought, CoT~\citep{wei2022cot}) or score-first (Ours) in Table~\ref{tab: cot}. JudgeLM with CoT performs similar agreement with our score-first baseline but with higher consistency, which means that explanation-first form, i.e., CoT, can alleviate the position bias of fine-tuned judges, but not bring significant agreement improvement. As a result, we choose the score-first method for \name{}, which has slightly less consistency but more flexible usage.

\begin{table}[ht]
\caption{\add{Comparison between GPT-4 teacher and JudgeLM-33B on JudgeLM $val$ set.}}
\vspace{-0.08in}
\begin{center}
\begin{tabular}{lccccccc}
\toprule
\add{Methods} & \begin{tabular}[c]{@{}c@{}}\add{Agreement ↑}\\  \add{(w/ GPT-4)} \end{tabular}  & \begin{tabular}[c]{@{}c@{}}\add{Consistency ↑}\\  \add{(w/ swap.)} \end{tabular} & \begin{tabular}[c]{@{}c@{}}\add{Bias ↓}\\  \add{toward 1st} \end{tabular}  & \begin{tabular}[c]{@{}c@{}}\add{Bias ↓}\\  \add{toward 2nd} \end{tabular} & \multicolumn{1}{l}{\add{Delta Bias ↓}} \\ %
\midrule
\add{GPT-4} & \add{--} & \add{85.82} & \add{6.10} & \add{8.10} & \add{2.00} \\ %
\add{JudgeLM-33B} & \add{89.03} & \add{91.36} & \add{5.55} & \add{3.09} & \add{2.46} \\ %
\bottomrule
\end{tabular}
\end{center}
\label{tab: comp_with_gpt4}
\vspace{-0.1in}
\end{table}

\subsection{Additional Experiment}
\label{subsec: exp_more}

\myparagraph{Comparison with GPT-4 Teacher}
As shown in Table~\ref{tab: comp_with_gpt4}, we further list the metrics except for the agreement with GPT-4 itself. JudgeLM-33B achieves higher consistency than GPT-4, which demonstrates that fine-tuning judges like JudgeLM-33B can achieve higher consistency through the proposed techniques.

The learning paradigm of JudgeLM is similar to knowledge distillation, as JudgeLM learns from the expert judgments provided by GPT-4. In the knowledge distillation domain~\citep{gou2021knowlededist}, the student model (JudgeLM) mimics the teacher model (GPT-4) to achieve competitive or even superior performance. Additionally, our JudgeLM employs three key methods and utilizes large-scale training data specifically for the judging task to enhance its agreement and consistency. As mentioned in ``Comparison on Other Human Evaluation Benchmarks" under Section~\ref{subsec: main_res}, our experiments demonstrate that a fine-tuned specialist judge model can surpass its generalist teacher on some judging benchmarks, i.e., JudgeLM $val$ set and PandaLM $test$ set.

\subsection{Details of Dataset}

For the proposed dataset and benchmark, we also provide the explanation of usage scope, details of metric calculations, dataset quality, question category \& distribution (among 19 categories), and comparison with UltraFeedback~\citep{cui2023ultrafeedback} in Sec.~\ref{subsec: data_more}. We hope the dataset can help researchers build more robust evaluation tools in the future.

\subsection{Generalization Ability of \name{}}

Not only judging answer pairs, but our \name{} can also generalize to various judging tasks (including math problems and code generation), unseen judging benchmarks (human-annotated benchmark, multimodal judging benchmark, retrieval-format benchmark, multiple-format benchmark, toxic chat benchmark, reward model benchmark), and other various judging extensions (grading single answer, multi-turn chat).
We leave the detailed analysis in Sec.~\ref{subsec: ext} of the appendix.

\subsection{More Discussion}

Due to the limitation of pages, we leave more discussion in Sec.~\ref{subsec: diss_more} of the appendix.

\section{Conclusion}
\label{sec:conclu}
In this paper, we first introduce a high-quality, large-scale dataset for LLM evaluation, that provides a robust foundation for future research. Next, the proposed \name{} as scalable judges for evaluating LLMs in open-ended tasks efficiently, achieving state-of-the-art judge performance on two benchmarks. 
Then, we \revision{analyze two} key biases \revision{and introduce a new format bias} in fine-tuning LLMs as judges, and address them with the proposed techniques. 
We hope our work can motivate more studies to explore the judge models for LLMs in open-ended tasks and build more powerful LLMs with guidance from judge models.

\textbf{Limitations.}
Although the proposed \name{} achieves encouraging performance and efficiency, the cost of the judge dataset limits further scaling up in the judge dataset. Currently, we spend about 4000 dollars to provide 100K high-quality GPT-4-generated judge data to the public. 
We expect to further improve the performance of judge models with the help of synthetic judge data.

\section*{Acknowledgments and Disclosure of Funding}
This work was partially supported by the National Natural Science Foundation of China (NSFC) under
Grant No. 62276108.

\bibliography{judgelm}

\begin{thebibliography}{57}
\providecommand{\natexlab}[1]{#1}
\providecommand{\url}[1]{\texttt{#1}}
\expandafter\ifx\csname urlstyle\endcsname\relax
  \providecommand{\doi}[1]{doi: #1}\else
  \providecommand{\doi}{doi: \begingroup \urlstyle{rm}\Url}\fi

\bibitem[Alex \& Graham(2023)Alex and Graham]{Cabrera2023zeno-chatbot}
Cabrera Alex and Neubig Graham.
\newblock Zeno chatbot report, 2023.
\newblock URL \url{https://github.com/zeno-ml/zeno-build/tree/main/examples/chatbot/report\#zeno-chatbot-report}.

\bibitem[Anand et~al.(2023)Anand, Nussbaum, Duderstadt, Schmidt, and Mulyar]{gpt4all}
Yuvanesh Anand, Zach Nussbaum, Brandon Duderstadt, Benjamin Schmidt, and Andriy Mulyar.
\newblock Gpt4all: Training an assistant-style chatbot with large scale data distillation from gpt-3.5-turbo.
\newblock \url{https://github.com/nomic-ai/gpt4all}, 2023.

\bibitem[Awadalla et~al.(2023)Awadalla, Gao, Gardner, Hessel, Hanafy, Zhu, Marathe, Bitton, Gadre, Sagawa, et~al.]{awadalla2023openflamingo}
Anas Awadalla, Irena Gao, Josh Gardner, Jack Hessel, Yusuf Hanafy, Wanrong Zhu, Kalyani Marathe, Yonatan Bitton, Samir Gadre, Shiori Sagawa, et~al.
\newblock Openflamingo: An open-source framework for training large autoregressive vision-language models.
\newblock \emph{arXiv preprint arXiv:2308.01390}, 2023.

\bibitem[Bai et~al.(2023)Bai, Bai, Chu, Cui, Dang, Deng, Fan, Ge, Han, Huang, et~al.]{bai2023qwen}
Jinze Bai, Shuai Bai, Yunfei Chu, Zeyu Cui, Kai Dang, Xiaodong Deng, Yang Fan, Wenbin Ge, Yu~Han, Fei Huang, et~al.
\newblock Qwen technical report.
\newblock \emph{arXiv preprint arXiv:2309.16609}, 2023.

\bibitem[Biderman et~al.(2023)Biderman, Schoelkopf, Anthony, Bradley, O’Brien, Hallahan, Khan, Purohit, Prashanth, Raff, et~al.]{biderman2023pythia}
Stella Biderman, Hailey Schoelkopf, Quentin~Gregory Anthony, Herbie Bradley, Kyle O’Brien, Eric Hallahan, Mohammad~Aflah Khan, Shivanshu Purohit, USVSN~Sai Prashanth, Edward Raff, et~al.
\newblock Pythia: A suite for analyzing large language models across training and scaling.
\newblock In \emph{International Conference on Machine Learning}, pp.\  2397--2430. PMLR, 2023.

\bibitem[Blunch(1984)]{blunch1984human_position_bias}
Niels~J Blunch.
\newblock Position bias in multiple-choice questions.
\newblock \emph{Journal of Marketing Research}, 21\penalty0 (2):\penalty0 216--220, 1984.

\bibitem[Brown et~al.(2020)Brown, Mann, Ryder, Subbiah, Kaplan, Dhariwal, Neelakantan, Shyam, Sastry, Askell, et~al.]{brown2020gpt3}
Tom Brown, Benjamin Mann, Nick Ryder, Melanie Subbiah, Jared~D Kaplan, Prafulla Dhariwal, Arvind Neelakantan, Pranav Shyam, Girish Sastry, Amanda Askell, et~al.
\newblock Language models are few-shot learners.
\newblock \emph{Advances in neural information processing systems}, 33:\penalty0 1877--1901, 2020.

\bibitem[Chiang et~al.(2023)Chiang, Li, Lin, Sheng, Wu, Zhang, Zheng, Zhuang, Zhuang, Gonzalez, et~al.]{chiang2023vicuna}
Wei-Lin Chiang, Zhuohan Li, Zi~Lin, Ying Sheng, Zhanghao Wu, Hao Zhang, Lianmin Zheng, Siyuan Zhuang, Yonghao Zhuang, Joseph~E Gonzalez, et~al.
\newblock Vicuna: An open-source chatbot impressing gpt-4 with 90\%* chatgpt quality.
\newblock \emph{See https://vicuna. lmsys. org (accessed 14 April 2023)}, 2023.

\bibitem[Chowdhery et~al.(2022)Chowdhery, Narang, Devlin, Bosma, Mishra, Roberts, Barham, Chung, Sutton, Gehrmann, et~al.]{chowdhery2022palm}
Aakanksha Chowdhery, Sharan Narang, Jacob Devlin, Maarten Bosma, Gaurav Mishra, Adam Roberts, Paul Barham, Hyung~Won Chung, Charles Sutton, Sebastian Gehrmann, et~al.
\newblock Palm: Scaling language modeling with pathways.
\newblock \emph{arXiv preprint arXiv:2204.02311}, 2022.

\bibitem[Chung et~al.(2022)Chung, Hou, Longpre, Zoph, Tay, Fedus, Li, Wang, Dehghani, Brahma, et~al.]{chung2022flant5}
Hyung~Won Chung, Le~Hou, Shayne Longpre, Barret Zoph, Yi~Tay, William Fedus, Eric Li, Xuezhi Wang, Mostafa Dehghani, Siddhartha Brahma, et~al.
\newblock Scaling instruction-finetuned language models.
\newblock \emph{arXiv preprint arXiv:2210.11416}, 2022.

\bibitem[Conover et~al.(2023)Conover, Hayes, Mathur, Xie, Wan, Shah, Ghodsi, Wendell, Zaharia, and Xin]{DatabricksBlog2023DollyV2}
Mike Conover, Matt Hayes, Ankit Mathur, Jianwei Xie, Jun Wan, Sam Shah, Ali Ghodsi, Patrick Wendell, Matei Zaharia, and Reynold Xin.
\newblock Free dolly: Introducing the world's first truly open instruction-tuned llm, 2023.
\newblock URL \url{https://www.databricks.com/blog/2023/04/12/dolly-first-open-commercially-viable-instruction-tuned-llm}.

\bibitem[Cui et~al.(2023)Cui, Yuan, Ding, Yao, Zhu, Ni, Xie, Liu, and Sun]{cui2023ultrafeedback}
Ganqu Cui, Lifan Yuan, Ning Ding, Guanming Yao, Wei Zhu, Yuan Ni, Guotong Xie, Zhiyuan Liu, and Maosong Sun.
\newblock Ultrafeedback: Boosting language models with high-quality feedback.
\newblock \emph{arXiv preprint arXiv:2310.01377}, 2023.

\bibitem[Gou et~al.(2021)Gou, Yu, Maybank, and Tao]{gou2021knowlededist}
Jianping Gou, Baosheng Yu, Stephen~J Maybank, and Dacheng Tao.
\newblock Knowledge distillation: A survey.
\newblock \emph{International Journal of Computer Vision}, 129\penalty0 (6):\penalty0 1789--1819, 2021.

\bibitem[Hendrycks et~al.(2020)Hendrycks, Burns, Basart, Zou, Mazeika, Song, and Steinhardt]{hendrycks2020mmlu}
Dan Hendrycks, Collin Burns, Steven Basart, Andy Zou, Mantas Mazeika, Dawn Song, and Jacob Steinhardt.
\newblock Measuring massive multitask language understanding.
\newblock \emph{arXiv preprint arXiv:2009.03300}, 2020.

\bibitem[Imani et~al.(2023)Imani, Du, and Shrivastava]{imani2023mathprompter}
Shima Imani, Liang Du, and Harsh Shrivastava.
\newblock Mathprompter: Mathematical reasoning using large language models.
\newblock \emph{arXiv preprint arXiv:2303.05398}, 2023.

\bibitem[Ivison et~al.(2023)Ivison, Wang, Pyatkin, Lambert, Peters, Dasigi, Jang, Wadden, Smith, Beltagy, et~al.]{ivison2023tulu2}
Hamish Ivison, Yizhong Wang, Valentina Pyatkin, Nathan Lambert, Matthew Peters, Pradeep Dasigi, Joel Jang, David Wadden, Noah~A Smith, Iz~Beltagy, et~al.
\newblock Camels in a changing climate: Enhancing lm adaptation with tulu 2.
\newblock \emph{arXiv preprint arXiv:2311.10702}, 2023.

\bibitem[Iyer et~al.(2022)Iyer, Lin, Pasunuru, Mihaylov, Simig, Yu, Shuster, Wang, Liu, Koura, et~al.]{iyer2022opt-iml}
Srinivasan Iyer, Xi~Victoria Lin, Ramakanth Pasunuru, Todor Mihaylov, Daniel Simig, Ping Yu, Kurt Shuster, Tianlu Wang, Qing Liu, Punit~Singh Koura, et~al.
\newblock Opt-iml: Scaling language model instruction meta learning through the lens of generalization.
\newblock \emph{arXiv preprint arXiv:2212.12017}, 2022.

\bibitem[Jiang et~al.(2023)Jiang, Li, Zhang, Huang, Lin, and Chen]{jiang2023tigerscore}
Dongfu Jiang, Yishan Li, Ge~Zhang, Wenhao Huang, Bill~Yuchen Lin, and Wenhu Chen.
\newblock Tigerscore: Towards building explainable metric for all text generation tasks.
\newblock \emph{Transactions on Machine Learning Research}, 2023.

\bibitem[Kim et~al.(2023)Kim, Shin, Cho, Jang, Longpre, Lee, Yun, Shin, Kim, Thorne, et~al.]{kim2023prometheus}
Seungone Kim, Jamin Shin, Yejin Cho, Joel Jang, Shayne Longpre, Hwaran Lee, Sangdoo Yun, Seongjin Shin, Sungdong Kim, James Thorne, et~al.
\newblock Prometheus: Inducing fine-grained evaluation capability in language models.
\newblock In \emph{The Twelfth International Conference on Learning Representations}, 2023.

\bibitem[Kingma \& Ba(2014)Kingma and Ba]{adam}
Diederik~P Kingma and Jimmy Ba.
\newblock Adam: A method for stochastic optimization.
\newblock \emph{arXiv preprint arXiv:1412.6980}, 2014.

\bibitem[Ko et~al.(2020)Ko, Lee, Kim, Kim, and Kang]{ko2020look1st}
Miyoung Ko, Jinhyuk Lee, Hyunjae Kim, Gangwoo Kim, and Jaewoo Kang.
\newblock Look at the first sentence: Position bias in question answering.
\newblock In \emph{2020 Conference on Empirical Methods in Natural Language Processing, EMNLP 2020}, pp.\  1109--1121. Association for Computational Linguistics (ACL), 2020.

\bibitem[Lambert et~al.(2024)Lambert, Pyatkin, Morrison, Miranda, Lin, Chandu, Dziri, Kumar, Zick, Choi, et~al.]{lambert2024rewardbench}
Nathan Lambert, Valentina Pyatkin, Jacob Morrison, LJ~Miranda, Bill~Yuchen Lin, Khyathi Chandu, Nouha Dziri, Sachin Kumar, Tom Zick, Yejin Choi, et~al.
\newblock Rewardbench: Evaluating reward models for language modeling.
\newblock \emph{arXiv preprint arXiv:2403.13787}, 2024.

\bibitem[Lee et~al.(2024)Lee, Kim, Park, Kim, and Seo]{lee2024prometheusvision}
Seongyun Lee, Seungone Kim, Sue~Hyun Park, Geewook Kim, and Minjoon Seo.
\newblock Prometheusvision: Vision-language model as a judge for fine-grained evaluation.
\newblock \emph{arXiv preprint arXiv:2401.06591}, 2024.

\bibitem[Li et~al.(2023{\natexlab{a}})Li, Sun, Yuan, Fan, Zhao, and Liu]{li2023autoj}
Junlong Li, Shichao Sun, Weizhe Yuan, Run-Ze Fan, Hai Zhao, and Pengfei Liu.
\newblock Generative judge for evaluating alignment.
\newblock \emph{CoRR}, abs/2310.05470, 2023{\natexlab{a}}.
\newblock URL \url{https://doi.org/10.48550/arXiv.2310.05470}.

\bibitem[Li et~al.(2023{\natexlab{b}})Li, Wang, Ma, Wu, Wang, Gao, and Liu]{li2023split}
Zongjie Li, Chaozheng Wang, Pingchuan Ma, Daoyuan Wu, Shuai Wang, Cuiyun Gao, and Yang Liu.
\newblock Split and merge: Aligning position biases in large language model based evaluators.
\newblock \emph{arXiv preprint arXiv:2310.01432}, 2023{\natexlab{b}}.

\bibitem[Liang et~al.(2022)Liang, Bommasani, Lee, Tsipras, Soylu, Yasunaga, Zhang, Narayanan, Wu, Kumar, et~al.]{liang2022helm}
Percy Liang, Rishi Bommasani, Tony Lee, Dimitris Tsipras, Dilara Soylu, Michihiro Yasunaga, Yian Zhang, Deepak Narayanan, Yuhuai Wu, Ananya Kumar, et~al.
\newblock Holistic evaluation of language models.
\newblock \emph{arXiv preprint arXiv:2211.09110}, 2022.

\bibitem[Lin(2004)]{lin2004rouge}
Chin-Yew Lin.
\newblock Rouge: A package for automatic evaluation of summaries.
\newblock In \emph{Text summarization branches out}, pp.\  74--81, 2004.

\bibitem[Lin et~al.(2023)Lin, Wang, Tong, Wang, Guo, Wang, and Shang]{lin2023toxicchat}
Zi~Lin, Zihan Wang, Yongqi Tong, Yangkun Wang, Yuxin Guo, Yujia Wang, and Jingbo Shang.
\newblock Toxicchat: Unveiling hidden challenges of toxicity detection in real-world user-ai conversation.
\newblock \emph{arXiv preprint arXiv:2310.17389}, 2023.

\bibitem[Liu et~al.(2023)Liu, Li, Wu, and Lee]{liu2023llava}
Haotian Liu, Chunyuan Li, Qingyang Wu, and Yong~Jae Lee.
\newblock Visual instruction tuning.
\newblock \emph{arXiv preprint arXiv:2304.08485}, 2023.

\bibitem[Loshchilov \& Hutter(2019)Loshchilov and Hutter]{Loshchilov2019adamw}
Ilya Loshchilov and Frank Hutter.
\newblock Decoupled weight decay regularization.
\newblock In \emph{ICLR}, 2019.

\bibitem[Luo et~al.(2023)Luo, Sun, Xu, Zhao, Lou, Tao, Geng, Lin, Chen, and Zhang]{luo2023wizardmath}
Haipeng Luo, Qingfeng Sun, Can Xu, Pu~Zhao, Jianguang Lou, Chongyang Tao, Xiubo Geng, Qingwei Lin, Shifeng Chen, and Dongmei Zhang.
\newblock Wizardmath: Empowering mathematical reasoning for large language models via reinforced evol-instruct.
\newblock \emph{arXiv preprint arXiv:2308.09583}, 2023.

\bibitem[Ni et~al.(2024{\natexlab{a}})Ni, Song, Ghosal, Li, Zhang, Yue, Xue, Zheng, Zhang, Shah, et~al.]{ni2024mixevalx}
Jinjie Ni, Yifan Song, Deepanway Ghosal, Bo~Li, David~Junhao Zhang, Xiang Yue, Fuzhao Xue, Zian Zheng, Kaichen Zhang, Mahir Shah, et~al.
\newblock Mixeval-x: Any-to-any evaluations from real-world data mixtures.
\newblock \emph{arXiv preprint arXiv:2410.13754}, 2024{\natexlab{a}}.

\bibitem[Ni et~al.(2024{\natexlab{b}})Ni, Xue, Yue, Deng, Shah, Jain, Neubig, and You]{ni2024mixeval}
Jinjie Ni, Fuzhao Xue, Xiang Yue, Yuntian Deng, Mahir Shah, Kabir Jain, Graham Neubig, and Yang You.
\newblock Mixeval: Deriving wisdom of the crowd from llm benchmark mixtures.
\newblock \emph{arXiv preprint arXiv:2406.06565}, 2024{\natexlab{b}}.

\bibitem[OpenAI(2022)]{openai2022chatgpt}
OpenAI.
\newblock Chatgpt, 2022.
\newblock URL \url{https://openai.com/blog/chatgpt/}.

\bibitem[OpenAI(2023)]{openai2023gpt4}
OpenAI.
\newblock Gpt-4 technical report, 2023.

\bibitem[Ouyang et~al.(2022)Ouyang, Wu, Jiang, Almeida, Wainwright, Mishkin, Zhang, Agarwal, Slama, Ray, et~al.]{ouyang2022instructgpt}
Long Ouyang, Jeffrey Wu, Xu~Jiang, Diogo Almeida, Carroll Wainwright, Pamela Mishkin, Chong Zhang, Sandhini Agarwal, Katarina Slama, Alex Ray, et~al.
\newblock Training language models to follow instructions with human feedback.
\newblock \emph{Advances in Neural Information Processing Systems}, 35:\penalty0 27730--27744, 2022.

\bibitem[Papineni et~al.(2002)Papineni, Roukos, Ward, and Zhu]{papineni2002bleu}
Kishore Papineni, Salim Roukos, Todd Ward, and Wei-Jing Zhu.
\newblock Bleu: a method for automatic evaluation of machine translation.
\newblock In \emph{Proceedings of the 40th annual meeting of the Association for Computational Linguistics}, pp.\  311--318, 2002.

\bibitem[Peng et~al.(2023)Peng, Li, He, Galley, and Gao]{peng2023alpaca-gpt4}
Baolin Peng, Chunyuan Li, Pengcheng He, Michel Galley, and Jianfeng Gao.
\newblock Instruction tuning with gpt-4.
\newblock \emph{arXiv preprint arXiv:2304.03277}, 2023.

\bibitem[Popovi{\'c}(2015)]{popovic-2015-chrf}
Maja Popovi{\'c}.
\newblock chr{F}: character n-gram {F}-score for automatic {MT} evaluation.
\newblock In \emph{Proceedings of the Tenth Workshop on Statistical Machine Translation}, pp.\  392--395, Lisbon, Portugal, September 2015. Association for Computational Linguistics.
\newblock \doi{10.18653/v1/W15-3049}.
\newblock URL \url{https://aclanthology.org/W15-3049}.

\bibitem[Raffel et~al.(2020)Raffel, Shazeer, Roberts, Lee, Narang, Matena, Zhou, Li, and Liu]{raffel2020T5}
Colin Raffel, Noam Shazeer, Adam Roberts, Katherine Lee, Sharan Narang, Michael Matena, Yanqi Zhou, Wei Li, and Peter~J Liu.
\newblock Exploring the limits of transfer learning with a unified text-to-text transformer.
\newblock \emph{The Journal of Machine Learning Research}, 21\penalty0 (1):\penalty0 5485--5551, 2020.

\bibitem[Raghubir \& Valenzuela(2006)Raghubir and Valenzuela]{raghubir2006human_decision_making}
Priya Raghubir and Ana Valenzuela.
\newblock Center-of-inattention: Position biases in decision-making.
\newblock \emph{Organizational Behavior and Human Decision Processes}, 99\penalty0 (1):\penalty0 66--80, 2006.

\bibitem[Ramesh et~al.(2021)Ramesh, Pavlov, Goh, Gray, Voss, Radford, Chen, and Sutskever]{ramesh2021zero}
Aditya Ramesh, Mikhail Pavlov, Gabriel Goh, Scott Gray, Chelsea Voss, Alec Radford, Mark Chen, and Ilya Sutskever.
\newblock Zero-shot text-to-image generation.
\newblock \emph{arXiv preprint arXiv:2102.12092}, 2021.

\bibitem[Sellam et~al.(2020)Sellam, Das, and Parikh]{sellam2020bleurt}
Thibault Sellam, Dipanjan Das, and Ankur Parikh.
\newblock Bleurt: Learning robust metrics for text generation.
\newblock In \emph{Proceedings of the 58th Annual Meeting of the Association for Computational Linguistics}, pp.\  7881--7892, 2020.

\bibitem[Sun et~al.(2023)Sun, Yu, Cui, Zhang, Zhang, Wang, Gao, Liu, Huang, and Wang]{sun2023emu}
Quan Sun, Qiying Yu, Yufeng Cui, Fan Zhang, Xiaosong Zhang, Yueze Wang, Hongcheng Gao, Jingjing Liu, Tiejun Huang, and Xinlong Wang.
\newblock Generative pretraining in multimodality.
\newblock \emph{arXiv preprint arXiv:2307.05222}, 2023.

\bibitem[Touvron et~al.(2023{\natexlab{a}})Touvron, Lavril, Izacard, Martinet, Lachaux, Lacroix, Rozi{\`e}re, Goyal, Hambro, Azhar, et~al.]{touvron2023llama}
Hugo Touvron, Thibaut Lavril, Gautier Izacard, Xavier Martinet, Marie-Anne Lachaux, Timoth{\'e}e Lacroix, Baptiste Rozi{\`e}re, Naman Goyal, Eric Hambro, Faisal Azhar, et~al.
\newblock Llama: Open and efficient foundation language models.
\newblock \emph{arXiv preprint arXiv:2302.13971}, 2023{\natexlab{a}}.

\bibitem[Touvron et~al.(2023{\natexlab{b}})Touvron, Martin, Stone, Albert, Almahairi, Babaei, Bashlykov, Batra, Bhargava, Bhosale, et~al.]{touvron2023llama2}
Hugo Touvron, Louis Martin, Kevin Stone, Peter Albert, Amjad Almahairi, Yasmine Babaei, Nikolay Bashlykov, Soumya Batra, Prajjwal Bhargava, Shruti Bhosale, et~al.
\newblock Llama 2: Open foundation and fine-tuned chat models.
\newblock \emph{arXiv preprint arXiv:2307.09288}, 2023{\natexlab{b}}.

\bibitem[Wang et~al.(2018)Wang, Golbandi, Bendersky, Metzler, and Najork]{wang2018position_bias}
Xuanhui Wang, Nadav Golbandi, Michael Bendersky, Donald Metzler, and Marc Najork.
\newblock Position bias estimation for unbiased learning to rank in personal search.
\newblock In \emph{Proceedings of the eleventh ACM international conference on web search and data mining}, pp.\  610--618, 2018.

\bibitem[Wang et~al.(2023)Wang, Yu, Zeng, Yang, Wang, Chen, Jiang, Xie, Wang, Xie, Ye, Zhang, and Zhang]{wang2023pandalm}
Yidong Wang, Zhuohao Yu, Zhengran Zeng, Linyi Yang, Cunxiang Wang, Hao Chen, Chaoya Jiang, Rui Xie, Jindong Wang, Xing Xie, Wei Ye, Shikun Zhang, and Yue Zhang.
\newblock Pandalm: An automatic evaluation benchmark for llm instruction tuning optimization.
\newblock \emph{CoRR}, abs/2306.05087, 2023.
\newblock URL \url{https://doi.org/10.48550/arXiv.2306.05087}.

\bibitem[Wei et~al.(2022)Wei, Wang, Schuurmans, Bosma, Xia, Chi, Le, Zhou, et~al.]{wei2022cot}
Jason Wei, Xuezhi Wang, Dale Schuurmans, Maarten Bosma, Fei Xia, Ed~Chi, Quoc~V Le, Denny Zhou, et~al.
\newblock Chain-of-thought prompting elicits reasoning in large language models.
\newblock \emph{Advances in Neural Information Processing Systems}, 35:\penalty0 24824--24837, 2022.

\bibitem[Xu et~al.(2023)Xu, Wang, Pan, Song, Freitag, Wang, and Li]{xu2023instructscore}
Wenda Xu, Danqing Wang, Liangming Pan, Zhenqiao Song, Markus Freitag, William~Yang Wang, and Lei Li.
\newblock Instructscore: Explainable text generation evaluation with finegrained feedback.
\newblock \emph{arXiv preprint arXiv:2305.14282}, 2023.

\bibitem[Yu et~al.(2023{\natexlab{a}})Yu, Jiang, Shi, Yu, Liu, Zhang, Kwok, Li, Weller, and Liu]{yu2023metamath}
Longhui Yu, Weisen Jiang, Han Shi, Jincheng Yu, Zhengying Liu, Yu~Zhang, James~T Kwok, Zhenguo Li, Adrian Weller, and Weiyang Liu.
\newblock Metamath: Bootstrap your own mathematical questions for large language models.
\newblock \emph{arXiv preprint arXiv:2309.12284}, 2023{\natexlab{a}}.

\bibitem[Yu et~al.(2023{\natexlab{b}})Yu, Yang, Li, Wang, Lin, Liu, Wang, and Wang]{yu2023mmvet}
Weihao Yu, Zhengyuan Yang, Linjie Li, Jianfeng Wang, Kevin Lin, Zicheng Liu, Xinchao Wang, and Lijuan Wang.
\newblock Mm-vet: Evaluating large multimodal models for integrated capabilities, 2023{\natexlab{b}}.

\bibitem[Yuan et~al.(2021)Yuan, Neubig, and Liu]{yuan2021bartscore}
Weizhe Yuan, Graham Neubig, and Pengfei Liu.
\newblock Bartscore: Evaluating generated text as text generation.
\newblock \emph{Advances in Neural Information Processing Systems}, 34:\penalty0 27263--27277, 2021.

\bibitem[Zhang et~al.(2023)Zhang, Han, Zhou, Hu, Yan, Lu, Li, Gao, and Qiao]{zhang2023llama-adapter}
Renrui Zhang, Jiaming Han, Aojun Zhou, Xiangfei Hu, Shilin Yan, Pan Lu, Hongsheng Li, Peng Gao, and Yu~Qiao.
\newblock Llama-adapter: Efficient fine-tuning of language models with zero-init attention.
\newblock \emph{arXiv preprint arXiv:2303.16199}, 2023.

\bibitem[Zhang et~al.(2022)Zhang, Roller, Goyal, Artetxe, Chen, Chen, Dewan, Diab, Li, Lin, et~al.]{zhang2022opt}
Susan Zhang, Stephen Roller, Naman Goyal, Mikel Artetxe, Moya Chen, Shuohui Chen, Christopher Dewan, Mona Diab, Xian Li, Xi~Victoria Lin, et~al.
\newblock Opt: Open pre-trained transformer language models.
\newblock \emph{arXiv preprint arXiv:2205.01068}, 2022.

\bibitem[Zhang et~al.(2019)Zhang, Kishore, Wu, Weinberger, and Artzi]{zhang2019bertscore}
Tianyi Zhang, Varsha Kishore, Felix Wu, Kilian~Q Weinberger, and Yoav Artzi.
\newblock Bertscore: Evaluating text generation with bert.
\newblock \emph{arXiv preprint arXiv:1904.09675}, 2019.

\bibitem[Zheng et~al.(2023)Zheng, Chiang, Sheng, Zhuang, Wu, Zhuang, Lin, Li, Li, Xing, et~al.]{zheng2023chatbot-arena}
Lianmin Zheng, Wei-Lin Chiang, Ying Sheng, Siyuan Zhuang, Zhanghao Wu, Yonghao Zhuang, Zi~Lin, Zhuohan Li, Dacheng Li, Eric Xing, et~al.
\newblock Judging llm-as-a-judge with mt-bench and chatbot arena.
\newblock \emph{arXiv preprint arXiv:2306.05685}, 2023.

\end{thebibliography}
\bibliographystyle{iclr2025_conference}

\appendix

\section{Appendix / supplemental material}

\newlength\savewidth\newcommand\shline{\noalign{\global\savewidth\arrayrulewidth\global\arrayrulewidth 1pt}\hline\noalign{\global\arrayrulewidth\savewidth}}

\subsection{\add{More about Dataset}}
\label{subsec: data_more}
\add{
\myparagraph{Dataset Usage Scope}
We emphasize that the JudgeLM dataset is intended only for academic research and any commercial use is prohibited. Because the OpenAI's terms prohibit developing models that compete with OpenAI, the instruction-tuning datasets generated by the OpenAI's API, i.e., Alpaca, PandaLM, etc., all follow this rule.
}

\myparagraph{Details of Metric Calculations}
For objective metrics, we use the judgments annotated by humans or GPT-4 as ground truth labels, and the judgments generated by judge models as predicted labels. We use $TP$, $FP$, $TN$, and $FN$ to represent the true positive, false positive, true negative, and false negative, respectively. The calculation of agreement, precision, recall, and F1-score are as follows:

\begin{align}
    \text{Agreement} &= (TP + TN) / (TP + FP + TN + FN), \\
    \text{Precision} &= TP / (TP + FP),\\
    \text{Recall} &= TP / (TP + FN),\\
    \text{F1-score} &= (2 * TP) / (2 * TP + FP + FN).
\end{align}

For reliability metrics, we compare the results before and after swapping the order of the two answers (A and B), i.e., from ''the first answer is A, the second answer is B" to "the first answer is B, the second answer is A". When the judging results change, we mark it as a biased sample. Samples with a bias toward the first position consist of three situations, ''from Answer A wins to tie", "from Answer A wins to Answer B wins", and "from tie to Answer B wins". Similarly, samples with a bias toward the second position also consist of three situations, ''from Answer B wins to tie", "from Answer B wins to Answer A wins", and "from tie to Answer A wins". The calculation of metrics of "bias toward 1st", "bias toward 2nd", "delta bias" are defined as follows:

\begin{align}
\text{bias toward 1st} &= \frac{\text{Number of samples with a bias toward the first position}}{\text{Numbers of total samples}}, \\
\text{bias toward 2nd} &= \frac{\text{Number of samples with a bias toward the second position}} {\text{Numbers of total samples}}, \\
\text{delta bias} &= \lvert \text{bias toward 1st} - \text{bias toward 2nd}) \rvert.
\end{align}

\myparagraph{Dataset Quality}
To ensure the high quality of the proposed dataset, we filter the low-quality data samples in each step. For data samples (including seed tasks and references) in four public datasets, i.e., Alpaca-GPT4~\citep{peng2023alpaca-gpt4}, Dolly-15K~\citep{DatabricksBlog2023DollyV2}, GPT4All-LAION~\citep{gpt4all}, and ShareGPT, we first remove data samples containing obviously incorrect, irrelevant, or harmful reference answers through automated filtering scripts. Next, we randomly sample 105K samples from the filtered set and extract answers from 11 LLMs. Then, we input the tasks, randomly sampled answer pairs, and optionally reference answers to the GPT-4 teacher for judgment. 
Finally, the authors of this work are involved in 
\revision{a multi-step validation process}
to ensure the quality, accuracy, and reliability of the judge samples.
\revision{This process includes an initial annotation step where GPT-4 provides preliminary judgments, followed by independent human re-annotation where authors provide simple judgments (``Answer 1 wins,'' ``Answer 2 wins'', or ``Tie'') without exposure to GPT-4's annotations. The final step involves cross-validation and refinement, where human judgments are compared with GPT-4 annotations to thoroughly verify the judge results, scores, and reasoning quality.}
Please note that incorporating high-quality answers from closed-sourced models e.g., Qwen~\citep{bai2023qwen} and Claude, could enhance the diversity of the dataset. We leave it as a future work.

\add{
\myparagraph{Question Category \& Distribution of Validation Set }
We count the distribution of questions in the JudgeLM $val$ set as shown in Table~\ref{tab: val_set_categories}. 
Please note that the question categories included in the JudgeLM $val$ set and $train$ set are the same, but none of the data samples are identical.
}

\myparagraph{Comparison with UltraFeedback}
Furthermore, we compare the JudgeLM dataset with UltraFeedback~\citep{cui2023ultrafeedback}, which is an excellent dataset serving as a solid foundation for feedback-learning research. JudgeLM has nearly half more seeds than UltraFeedback, and an additional validation set containing 5K seeds. JudgeLM and UltraFeedback both provide scalar and text feedback, GPT-4 annotation, and fine-grained consideration. However, the JudgeLM dataset is further checked and re-annotated by humans, which provides double-checking on the quality of feedback. Moreover, JudgeLM supports judging with references, which can make up for the lack of pre-trained knowledge or insert specific judge preferences. Finally, JudgeLM clearly splits the seeds into 19 categories providing intuitive ability estimation for judges.

\begin{table}[htp]
\begin{tabular}{lccccccc}
\toprule
Dataset & \begin{tabular}[c]{@{}c@{}} \add{$train$} \\  \add{Seeds} \end{tabular} & \begin{tabular}[c]{@{}c@{}}\add{$val$} \\  \add{Seeds} \end{tabular} & \begin{tabular}[c]{@{}c@{}}\add{Feedback}\\  \add{Format} \end{tabular} & Annotator  & \begin{tabular}[c]{@{}c@{}}\add{fine}\\  \add{grained?} \end{tabular} & \begin{tabular}[c]{@{}c@{}}\add{with}\\  \add{Ref.?} \end{tabular} & \begin{tabular}[c]{@{}c@{}}\add{Seeds}\\  \add{Categories} \end{tabular} \\
\midrule
UltraFeedback & 64K            & 0                & Scalar \& Text  & GPT-4          & Y                & N                & -                \\
JudgeLM       & 100K           & 5K               & Scalar \& Text  & GPT-4 \& Human & Y                & Y                & 19        \\      
\bottomrule
\end{tabular}
\end{table}

\begin{table}[ht]
    \renewcommand\arraystretch{1.2}
    
    \centering
    \caption{\add{Distribution of question categories in JudgeLM $val$ set}}
    \begin{tabular}{ccc|ccc}
    \toprule
\textbf{}               & \add{count} & \add{percentage} &                    & {\add{count}} & {\add{percentage}} \\
\midrule
{\add{culture}}        & \add{233}   & \add{4.66\%}     & {\add{planning}}  & \add{309}   & \add{6.18\%}     \\
{\add{recommendation}} & \add{482}   & \add{9.64\%}     & {\add{roleplay}}  & \add{77}    & \add{1.54\%}     \\
{\add{finance}}        & \add{142}   & \add{2.84\%}     & {\add{coding}}    & \add{201}   & \add{4.02\%}     \\
{\add{science}}        & \add{393}   & \add{7.86\%}     & {\add{health}}    & \add{278}   & \add{5.56\%}     \\
{\add{technique}}      & \add{42}    & \add{0.84\%}     & {\add{writing}}   & \add{625}   & \add{12.50\%}    \\
{\add{common-sense}}   & \add{373}   & \add{7.46\%}     & {\add{hardware}}  & \add{130}   & \add{2.60\%}     \\
{\add{art}}            & \add{335}   & \add{6.70\%}     & {\add{history}}   & \add{243}   & \add{4.86\%}     \\
{\add{math}}           & \add{250}   & \add{5.00\%}     & {\add{geography}} & \add{199}   & \add{3.98\%}     \\
{\add{private-matter}} & \add{421}   & \add{8.42\%}     & {\add{others}}    & \add{63}    & \add{1.26\%}    \\
{\add{law}}            & \add{204}   & \add{4.08\%}     & {\add{total}}     & \add{5000}  & \add{100.00\%}  \\
    \bottomrule
    \end{tabular}
    \label{tab: val_set_categories}
\end{table}    

\myparagraph{\revision{Comparison with PandaLM Test Set}}
\revision{Furthermore, we compare the JudgeLM dataset with the PandaLM test set. 
An analysis of task distributions in Table~\ref{tab: val_set_diff} shows significant differences between the PandaLM test set and the JudgeLM benchmark. For example, business, fact-QA, summarizing, linguistics, emotion, entity-processing, explain, retrieval, document, and chat are well-represented in PandaLM but absent in JudgeLM, while writing and roleplay show a significant delta percentage (over 4\%). This confirms that the PandaLM test set includes 49\% unseen task samples that are out of distribution for JudgeLM.}

\begin{table}[ht]
\renewcommand\arraystretch{1.2}
\centering
\caption{\revision{Distribution of question categories in PandaLM test set. The $\Delta$ represents the percentage difference compared to the JudgeLM benchmark. We \textbf{bolded} categories that appear in the PandaLM test set but don't exist in the JudgeLM benchmark.}}
\begin{tabular}{cccccccc}
\toprule
\textbf{}         & count & percentage & $\Delta$ &           & count & percentage & $\Delta$ \\
\midrule
\textbf{business}          & 87    & 8.71\%     & 8.71\%           & writing   & 81    & 8.11\%     & -4.39\%          \\
\textbf{fact-QA}           & 70    & 7.01\%     & 7.01\%           & planning  & 57    & 5.71\%     & -0.47\%          \\
\textbf{summarizing}       & 64    & 6.41\%     & 6.41\%           & roleplay  & 57    & 5.71\%     & 4.17\%           \\
\textbf{linguistics}       & 45    & 4.50\%     & 4.50\%           & coding    & 54    & 5.41\%     & 1.39\%           \\
\textbf{emotion}           & 45    & 4.50\%     & 4.50\%           & art       & 44    & 4.40\%     & -2.30\%          \\
\textbf{entity-processing} & 42    & 4.20\%     & 4.20\%           & finance   & 40    & 4.00\%     & 1.16\%           \\
\textbf{explain}           & 41    & 4.10\%     & 4.10\%           & culture   & 38    & 3.80\%     & -0.86\%          \\
\textbf{retrieval}          & 40    & 4.00\%     & 4.00\%           & math      & 37    & 3.70\%     & -1.30\%          \\
\textbf{document}          & 30    & 3.00\%     & 3.00\%           & geography & 12    & 1.20\%     & -2.78\%          \\
\textbf{chat}              & 26    & 2.60\%     & 2.60\%           & others    & 5     & 0.50\%     & -0.76\%          \\
recommendation    & 84    & 8.41\%     & -1.23\%          & total     & 999   & 100.00\%   &     \\
\bottomrule
\end{tabular}
\label{tab: val_set_diff}
\end{table}

\subsection{Fine-tuning Setting}
\label{subsec: setting_more}
We list the hyper-parameters we used, as shown in Table~\ref{tab: ft_cfg}.
\begin{table}[htp!]
    \caption{\name{} fine-tuning setting.}
    \centering
    \begin{tabular}{l|c}
        config & \name{} / -7B / -13B / -33B \\
        \shline
        base model & Vicuna / -7B / -13B / -33B \\
        model max length & 2048\\
        
        fine-tuning data source & \name{}-100K \\
        
        learning rate & 2e-5 \\
        learning rate schedule & cosine decay \\
        
        optimizer & AdamW~\citep{adam,Loshchilov2019adamw} \\
        optimizer hyper-parameters & $\beta_1$, $\beta_2$, $\epsilon$ = 0.9, 0.999, 1e-8 \\
        weight decay & 0.0 \\

        GPU nums & 8 / 8 / 16 \\
        batch size & 128 \\
        training epochs & 3 \\
        warmup ratio & 0.003 \\
        
        numerical precision & bf16, tf32 \\
        ZeRO optimizer~\citep{ramesh2021zero} & stage 3 \\
        gradient checkpointing & True \\
        GPT-3.5 and GPT-4 version & 2023-03-15-preview \\
    
    \end{tabular}
    \vspace{-0.1 in}
    \label{tab: ft_cfg}
\end{table}

\subsection{Generalization Ability of \name{}}
\label{subsec: ext}

To validate the generalization ability of JudgeLM, we test JudgeLM on various judging tasks (including math problems and code generation), unseen judging benchmarks (human-annotated benchmark, multimodal judging benchmark, retrieval-format benchmark, multiple-format benchmark), and other various judging extensions (grading single answer, multi-turn chat).

\myparagraph{Generalize to Various Judging Tasks.} To further validate the judging performance on questions with specific categories, we present the judging results of JudgeLM-33B on these questions, i.e., coding, common-sense, math, roleplay, and writing. Table~\ref{tab: val_categories} shows that JudgeLM can handle the judging tasks of various categories. We also find that the judging performance of math questions is slightly lower than coding and common-sense questions. However we think it is a common problem of large language models~\citep{imani2023mathprompter}, and future advancement on base models~\citep{yu2023metamath, luo2023wizardmath} would alleviate this problem.

\begin{table}[htp]
\caption{\add{Performance of JudgeLM-33B with specific categories on JudgeLM $val$ set.}}
\vspace{-0.08in}
\begin{center}
\begin{tabular}{lccccccc}
\toprule
\add{ } & \begin{tabular}[c]{@{}c@{}} \add{Agreement ↑} \\  \add{(w/ GPT-4)} \end{tabular} & \begin{tabular}[c]{@{}c@{}}\add{Consistency ↑} \\  \add{(w/ swap.)} \end{tabular} & \begin{tabular}[c]{@{}c@{}}\add{Bias ↓}\\  \add{toward 1st} \end{tabular}  & \begin{tabular}[c]{@{}c@{}}\add{Bias ↓}\\  \add{toward 2nd} \end{tabular} & \multicolumn{1}{l}{\add{Delta Bias ↓}} \\
\midrule
\multicolumn{6}{l}{\textbf{\textit{\add{coding}}}} \\
val w/o ref      & \add{88.08}             & \add{88.60}              & \add{6.22}           & \add{5.18}             & \add{1.04}               \\
val w/ ref & \add{88.83}             & \add{91.37}              & \add{3.55}           & \add{5.08}             & \add{1.53}                \\
\midrule
\multicolumn{6}{l}{\textbf{\textit{\add{common-sense}}}} \\
val w/o ref      & \add{88.41}             & \add{90.43}              & \add{7.25}           & \add{2.32}             & \add{4.93}               \\
val w/ ref & \add{90.37}             & \add{92.35}              & \add{4.53}           & \add{3.12}             & \add{1.41}                \\
\midrule
\multicolumn{6}{l}{\textbf{\textit{\add{math}}}} \\
val w/o ref      & \add{86.45}             & \add{84.08}              & \add{7.76}           & \add{8.16}             & \add{0.40}               \\
val w/ ref & \add{86.81}             & \add{87.85}              & \add{4.45}           & \add{7.69}             & \add{3.24}                \\
\midrule
\multicolumn{6}{l}{\textbf{\textit{\add{croleplay}}}} \\
val w/o ref      & \add{88.00}             & \add{88.00}              & \add{6.67}           & \add{5.33}             & \add{1.34}               \\
val w/ ref & \add{88.72}             & \add{89.12}              & \add{5.01}           & \add{5.87}             & \add{0.86}                \\
\midrule
\multicolumn{6}{l}{\textbf{\textit{\add{writting}}}} \\
val w/o ref      & \add{87.33}             & \add{92.36}              & \add{3.25}           & \add{4.39}             & \add{1.14}               \\
val w/ ref & \add{89.11}             & \add{92.83}              & \add{3.67}           & \add{3.50}             & \add{0.17}                \\
\bottomrule
\end{tabular}
\end{center}
\label{tab: val_categories}
\end{table}

\myparagraph{Generalize to Multimodal Judging Benchmark.}
Traditional multimodal evaluation needs prediction to match the ground truth exactly. For some open-ended questions, a human-like evaluator is needed to determine whether the prediction is close to the ground truth range. 
Modern multimodal, such as MM-Vet~\citep{yu2023mmvet} and Prometheus-Vision~\citep{lee2024prometheusvision}, which use GPT-4V, GPT-4 or GPT-3.5 as judges. The API-based judge takes the question text, ground-truth text, the model's prediction, and optional input image as input, and makes judgments based on them. 
Our \name{} also provides good practice for such a multimodal evaluation by a slightly modified template as shown in Fig.~\ref{fig: temp_vqa}. Thanks to its capacity to judge open-ended answers, our \name{} can also perform well in judging multimodal models, as shown in Fig.~\ref{fig: vqa}. %

We further conduct experiments to evaluate \name{}' ability to judge multimodal models when compared with close-sourced LLM, i.e., GPT-3.5 and GPT-4. We first use GPT-4, GPT-3.5, and JudgeLM to judge the LLaVA's output~\citep{liu2023llava}, respectively. 
Then, we collect judgments from human annotators, whose judgments include three situations: completely correct, semi-correct, and completely wrong. Last, we compute the metrics between the LLM judges' judgments and human judgments, as shown in Table~\ref{tab: mmvet}. It can be observed that JudgeLM outperforms GPT-4 (0-shot) and GPT-3.5 (7-shot). Besides, JudgeLM achieves 2.5\% higher precision than GPT-4 (7-shot). 
The encouraging results demonstrate the generalization ability of \name{} in dealing with multimodal judging.
Furthermore, JudgeLM can use large multimodal models, e.g., LLaVA~\citep{liu2023llava}, as the backbone for better processing the multimodal judging. We leave it as a future work.

\begin{table}[htp]
\centering
\caption{\add{\name{} zero-shot evaluation results on human-annotated MM-Vet benchmark.}}
\begin{tabular}{lcccc}
\toprule
Methods & \begin{tabular}[c]{@{}c@{}}Agreement ↑ \\  (w/ Human) \end{tabular} & \begin{tabular}[c]{@{}c@{}}Precision ↑\\  (w/ Human) \end{tabular} & \begin{tabular}[c]{@{}c@{}}Recall ↑\\  (w/ Human) \end{tabular} & \begin{tabular}[c]{@{}c@{}}F1 ↑\\  (w/ Human) \end{tabular} \\
\midrule
\add{GPT-4 (7-shot)}       & \add{95.58}    & \add{88.63}     & \add{87.79}  & \add{88.04}    \\
\add{GPT-4 (0-shot)}       & \add{86.70}    & \add{79.75}     & \add{86.41}  & \add{81.81}    \\
\add{GPT-3.5 (7-shot)}     & \add{83.03}    & \add{76.14}     & \add{74.84}  & \add{73.62}    \\
\midrule
\add{JudgeLM-33B (0-shot)} & \add{91.74}    & \add{91.08}     & \add{85.58}  & \add{87.26}   \\
\bottomrule
\end{tabular}
\label{tab: mmvet}
\vspace{-0.1 in}
\end{table}

\myparagraph{Generalize to Out-of-distribution ToxicChat Benchmark.}
To further evaluate the generalization ability of the proposed \name{}, we selected an out-of-distribution benchmark, i.e., ToxicChat~\citep{lin2023toxicchat}, for evaluation. Following the guidelines provided in the ToxicChat dataset, we conducted experiments on the latest test set (0124) of ToxicChat, comparing OpenAI Moderation, the GPT-4 teacher, and our proposed JudgeLM. OpenAI Moderation is an API trained on publicly available toxicity datasets, primarily sourced from social media. For both the GPT-4 teacher and JudgeLM, we used the same templates and thresholds. As shown in Table~\ref{tab: comp_toxic}, our proposed JudgeLM achieves superior precision and comparable accuracy to the specialist model, i.e., OpenAI Moderation. These results demonstrate that JudgeLM can further generalize to out-of-distribution datasets such as ToxicChat.

\begin{table}[htp]
\centering
\caption{JudgeLM zero-shot evaluation results on toxic-chat test set.}
\begin{tabular}{lcccc}
\toprule
Methods                     & Accuracy↑ & Precision↑ & Recall↑ & F1↑   \\
\midrule
\textbf{\textit{Specialist API-based Method}} &           &            &         &       \\
OpenAI Moderation           & 89.70     & 54.76      & 69.89   & 61.41 \\
\midrule
\textbf{\textit{Generalist API-based Method}} &           &            &         &       \\
GPT-4                       & 88.08     & 52.17      & 73.20   & 60.92 \\
\midrule
\textbf{\textit{Open-sourced Method}}         &           &            &         &       \\
JudgeLM-33B                 & 89.66     & 58.79      & 61.88   & 60.30 \\
\bottomrule
\end{tabular}
\label{tab: comp_toxic}
\end{table}

\myparagraph{Generalize to Retrieval-format Benchmark.}
In real-world applications, we do not always have well-organized reference answers for judging. 
To evaluate the capability of \name{} in dealing with this situation, we inject the original reference answers into a randomly selected paragraph. As shown in Table~\ref{tab: ref_retrival}, we select paragraphs with different words, and evaluate \name{}-33B with the injected paragraphs as references in a zero-shot setting.
The results show that \name{} can retrieve the correct answers from the paragraphs and make judgments based on them. When the words of paragraphs increase to 400, \name{} faces a maximum drop of 3.73\% agreement and 3.37\% consistency. The results demonstrate that \name{} is promising for utilizing \name{} to deal with unstructured references or judge in the retrieved form.

\begin{table}[ht]
\caption{\add{Performance of JudgeLM-33B with injected paragraphs as references on JudgeLM $val$ set.}}
\vspace{-0.08in}
\begin{center}
\begin{tabular}{lccccccc}
\toprule
\begin{tabular}[c]{@{}c@{}}\add{Reference}\\  \add{Paragraph} \end{tabular} & \begin{tabular}[c]{@{}c@{}}\add{Agreement ↑}\\  \add{(w/ GPT-4)} \end{tabular}  & \begin{tabular}[c]{@{}c@{}}\add{Consistency ↑}\\  \add{(w/ swap.)} \end{tabular} & \begin{tabular}[c]{@{}c@{}}\add{Bias ↓}\\  \add{toward 1st} \end{tabular}  & \begin{tabular}[c]{@{}c@{}}\add{Bias ↓}\\  \add{toward 2nd} \end{tabular} & \multicolumn{1}{l}{\add{Delta Bias ↓}} \\ %
\midrule
\add{No}                  & \add{89.32}       & \add{92.37}         & \add{3.62}              & \add{4.01}              & \add{0.39}         \\
\add{50 words}            & \add{87.78}       & \add{92.35}         & \add{3.00}              & \add{4.65}              & \add{1.65}         \\
\add{100 words}           & \add{87.69}       & \add{91.84}         & \add{2.62}              & \add{5.54}              & \add{2.92}         \\
\add{200 words}           & \add{86.77}       & \add{90.48}         & \add{2.70}              & \add{6.82}              & \add{4.12}         \\
\add{300 words}           & \add{86.25}       & \add{89.26}         & \add{3.13}              & \add{7.61}              & \add{4.48}         \\
\add{400 words}           & \add{85.59}       & \add{89.00}         & \add{2.98}              & \add{8.02}              & \add{5.04}         \\
\bottomrule
\end{tabular}
\end{center}
\label{tab: ref_retrival}
\vspace{-0.2in}
\end{table}

\myparagraph{Generalize to Multiple-format benchmark.}
To get the optimal ranking for N answers from different LLMs, other judge models need to call the model $O(n^2)$ times to get the full matrix, which is a much less efficient solution. We attempt to resolve this limitation by extending our \name{} to process multiple answers at the same time. We first need to modify the template as shown in Fig.~\ref{fig: temp_single_ans}. As shown in Fig.~\ref{fig: rank_multi_ans}, \name{} can judge and rank the multiple answers within the context limit of LLM.

We further conduct experiments to evaluate the consistency in the judging form of answer pairs and multiple answers. 
We first generate answers on JudgeLM $val$ set through 3 LLMs, i.e., Vicuna-13B, LLaMA-7B, and alpaca-7B, for evaluation.
Then we use pairwise judging and multiple judging to grade answers and rank them, respectively. Last, we compute the consistency between the two ranking results. Please note that `Error Rate@2' indicates the position orderings of two answers are different between the result of paired judgment and the result of multiple judgment, and `Error Rate@3' means the position orderings of three answers are different. Table~\ref{tab: multiple-judging} shows that consistency between judging pairwise and judging multiple can reach 93.48\%, and only 0.14\% results are totally wrong. The results are impressive but the 6.38\% of `Error Rate@2' also shows room for improvement as well, which could be addressed by the further improvement of \name{}'s self-consistency.

\begin{table}[htp]
\centering
\caption{\add{Performance of JudgeLM-33B in judging multiple answers on JudgeLM $val$ set. We calculate the consistency between the pairwise judging results and multiple judging ones.}}
\begin{tabular}{lccc}
\toprule
            & \begin{tabular}[c]{@{}c@{}}Consistency ↑\\  (w/ pairwise) \end{tabular} & \begin{tabular}[c]{@{}c@{}}Error Rate@2 ↓\\  (w/ pairwise) \end{tabular} & \begin{tabular}[c]{@{}c@{}}Error Rate@3 ↓\\  (w/ pairwise) \end{tabular} \\
\midrule
\add{JudgeLM-33B multipile} & \add{93.48}     & \add{6.38}  & \add{0.14}  \\
\bottomrule
\end{tabular}
\label{tab: multiple-judging}
\vspace{-0.1 in}
\end{table}

\begin{table}[htp]
\centering
\caption{Comparison with advanced reward models on RewardBench.}
\begin{tabular}{lccccc}
\toprule
Model                        & Score & Chat & Hard & Safety & Reason \\
\midrule
\textbf{\textit{Closed-source Models}}         &       &      &      &        &        \\
GPT-3.5-turbo-0125           & 64.5  & 92.2 & 44.5 & 62.3   & 59.1   \\
\textbf{\textit{Powerful Reward Models.}}       &       &      &      &        &        \\
Prometheus-8$\times$7B-v2.0  & 75.3  & 93   & 47.1 & 83.5   & 77.4   \\
Prometheus-7B-v2.0   & 72.5  & 85.5 & 49.1 & 78.7   & 76.5   \\
UltraRM-13B          & 67.6  & 96.4 & 55.5 & 56.0     & 62.4   \\
TIGERScore-13B       & 35.6  & 35.2 & 32.9 & 41.5   & 32.7   \\
\midrule
\textbf{\textit{Llama-2-based Models.}}         &       &      &      &        &        \\
Tulu-2-dpo-70B       & 79.0    & 97.5 & 60.5 & 83.9   & 74.1   \\
Tulu-2-dpo-13B       & 76.4  & 95.8 & 58.3 & 78.2   & 73.2   \\
Tulu-2-dpo-7B        & 74.7  & 97.5 & 56.1 & 73.3   & 71.8   \\
\midrule
\textbf{\textit{Qwen1.5-based Models.}}         &       &      &      &        &        \\
Qwen1.5-72B-Chat     & 71.5  & 62.3 & 66.0   & 72.0     & 85.5   \\
Qwen1.5-14B-Chat     & 73.4  & 57.3 & 70.2 & 76.3   & 89.6   \\
Qwen1.5-7B-Chat      & 72.0    & 53.6 & 69.1 & 74.8   & 90.4   \\
\midrule
\textbf{\textit{Ours.}}                         &       &      &      &        &        \\
JudgeLM-7B                   & 78.5  & 92.2 & 56.1 & 83.2   & 82.3   \\
\bottomrule
\end{tabular}
\label{tab: comp_reward}
\end{table}

\myparagraph{Generalize to Single Answer Grading.} 
The Concurrent judge method~\citep{wang2023pandalm} usually judges a pair of answers to decide which one is better or tie but they lack the ability to evaluate a single answer.
Thanks to our judging mode of scoring first and then calculating the judging results, our \name{} provides an alternative practice to grade a single answer by slightly modifying the template as shown in Fig.~\ref{fig: temp_single_ans}.
Putting the reference answer in the first position and giving it a full grade as a prior, \name{} can give quantitative fine-grained evaluations as shown in Fig.~\ref{fig: grading_single_ans}.

The capability of grading a single answer is an important extension, which only relies the text-form prediction and ground truth to make judgments. For example, the following extension ``Judging multimodal models" is also based on this capability.

\myparagraph{Generalize to Reward Model.} 
We further compare the proposed JudgeLM with closed-source methods, powerful reward models, Llama-2-based reward methods~\citep{touvron2023llama2, ivison2023tulu2}, and Qwen-1.5-based reward methods~\citep{bai2023qwen} on the requested reward model benchmark as shown in Table~\ref{tab: comp_reward}. Following the evaluation manner of GPT-3.5-turbo-0125 and Prometheus series~\citep{kim2023prometheus} in RewardBench~\citep{lambert2024rewardbench}, we evaluate the models among 4 subsets, i.e., Chat, (Chat) Hard, Safety, and Reasoning, and present averaged score in the Score column. The proposed JudgeLM-7B outperforms the GPT-3.5-turbo-0125, the Prometheus series reward models~\citep{kim2023prometheus}, UltraRM-13B~\citep{cui2023ultrafeedback}, TIGERScore-13B~\citep{jiang2023tigerscore}, Tulu-2-dpo-13B~\citep{ivison2023tulu2}, Tulu-2-dpo-7B~\citep{ivison2023tulu2}, and Qwen1.5-based reward models~\citep{bai2023qwen}. Besides, JudgeLM-7B even achieves similar performance to Tulu-2-dpo-70B~\citep{ivison2023tulu2}. Noting the RewardBench paper mentions that involving Llama-3 as the base model can significantly improve the metrics on Hard and Reasoning, we leave it as future work.

\myparagraph{Multi-turn Chat about Judgments.}
It is worth noting that fine-tuning with judge samples does not compromise the multi-turn chat ability extended from base models. As illustrated in Fig.~\ref{fig: multi_turn_chat} \add{and Fig.~\ref{fig: multi_turn_chat_additional}}, our \name{} retains the capability to engage in meaningful dialogues with users, providing them with a richer context, detailed information, additional examples, and specific details.

\subsection{\add{More Discussion}}
\label{subsec: diss_more}
\begin{table}[htp]
\caption{\add{Comparison of different base models for JudgeLM-7B on JudgeLM $val$ set.}}
\vspace{-0.08in}
\begin{center}
\begin{tabular}{lccccccc}
\toprule
\add{\begin{tabular}[c]{@{}c@{}} \add{Base Models}\\  \add{(for JudgeLM-7B)} \end{tabular}} & \begin{tabular}[c]{@{}c@{}} \add{Agreement ↑} \\  \add{(w/ GPT-4)} \end{tabular} & \begin{tabular}[c]{@{}c@{}}\add{Precision ↑} \\  \add{(w/ GPT-4)} \end{tabular} & \begin{tabular}[c]{@{}c@{}} \add{Recall ↑}\\  \add{(w/ GPT-4)} \end{tabular} & \begin{tabular}[c]{@{}c@{}} \add{F1 ↑} \\  \add{(w/ GPT-4)} \end{tabular} & \begin{tabular}[c]{@{}c@{}}\add{Consistency ↑} \\  \add{(w/ swap.)} \end{tabular} \\
\midrule
\multicolumn{6}{l}{\textbf{\textit{\add{Judge w/o reference.}}}} \\
\add{Vicuna}      & \add{81.11}             & \add{69.67}              & \add{78.39}           & \add{72.21}             & \add{83.57}               \\
\add{LLaMA2-chat} & \add{83.87}             & \add{73.43}              & \add{80.06}           & \add{75.91}             & \add{85.17}                \\
\midrule
\multicolumn{6}{l}{\textbf{\textit{\add{Judge w/ reference.}}}} \\
\add{Vicuna}      & \add{84.08}             & \add{75.92}              & \add{82.55}           & \add{78.28}             & \add{84.46}                \\
\add{LLaMA2-chat} & \add{86.60}             & \add{79.47}              & \add{83.11}           & \add{81.02}             & \add{87.74}           \\  
\bottomrule
\end{tabular}
\end{center}
\label{tab: comp_llama2}
\end{table}

\add{
\myparagraph{Format Bias}
As shown in Table~\ref{tab: multiple-judging}, it can be seen that the judging of multiple answers does not receive a significant performance drop. We hold the viewpoint that judging multiple answers is an easy extension for JudgeLM, which does not change the basis for judging. As mentioned in `4 Inherent Biases - Format Bias', format bias means the model judging basis changes from pre-trained knowledge to reference, or vice versa. So, judging in mismatched situations faces format bias but judging multiple answers does not.
}

\add{
\myparagraph{Other Human-annotated benchmarks}
For a fair comparison, we also evaluate JudgeLM on the PandaLM $test$ set in a zero-shot setting. The PandaLM $train$ and $test$ sets are annotated by GPT3.5 and humans, respectively. 
As shown in Table~\textcolor{red}{2}, 
the zero-shot results of JudgeLM also outperform other judging methods, i.e., PandaLM, GPT-3.5, and GPT-4. Furthermore, JudgeLM also achieves a superior 0-shot judging performance on the multimodal benchmark with human annotation, i.e., MM-Vet, as shown in Table~\ref{tab: mmvet}.
}

\add{
\myparagraph{Reasoning Ability of LLM Judges}
Nowadays, NLP researchers are still struggling with proposing LLMs with superior reasoning abilities. JudgeLM also needs the proposed reference sup method to enhance the judging ability for out-of-domain or counterfactual tasks, as shown in Fig.~\ref{fig: reference_sup} and Fig.~\ref{fig: change_reference}. Notably, the proposed JudgeLM can benefit from stronger foundation LLMs, e.g., the LLaMA2-7B-Chat-based~\citep{touvron2023llama2} JudgeLM outperforms the original JudgeLM-7B on all metrics, as shown in Table~\ref{tab: comp_llama2}. The research of judge models is critical for the development of LLMs and can benefit from advanced LLMs, establishing a positive cycle.
}

\myparagraph{Critiques for Judgements}
Beyond using LLMs to compare answer pairs and generate reasons, the critique and correction of these reasons have become increasingly important topics. This approach allows LLMs to reassess their generated reasons in multiple rounds, thereby enhancing judging accuracy. Works like UltraCM, Auto-J, and Shepherd have reliably evaluated the quality of textual reasons. Recently, CriticBench has also provided a reliable benchmark to evaluate the reasons and evaluation abilities of LLMs.

For JudgeLM, its ability to generalize to multi-turn chat enables us to construct multi-turn judgement critique data by combining data samples without references and those with references. Through two rounds of judge Q\&A, i.e., without reference and with reference, JudgeLM can acquire the capability to critique its own judgments. We leave these experiments for future work.

\add{
\myparagraph{Reference Drop as an Independent Method}
We think the reference drop is an independent and significant method. At first, we argue that judging with or without references are two sub-benchmarks, which require judges to make judgments with internal knowledge or by comparing LLM-generated answers with a reference answer, respectively. The reference drop is not only a simple but effective hyper-parameter, but also an important method that bridges the two sub-benchmarks, which enables the JudgeLM to make judgments in different situations.
}

\begin{table}[htp]
\caption{Comparison between PandaLM and JudgeLM components in terms of datasets and methods.}
\begin{tabular}{lccccc}
\toprule
Base Model              & Data                & Method    & val w/ ref? & Agreement ↑ & Consistency ↑ \\
\midrule
\multicolumn{6}{l}{\textbf{\textit{PandaLM baseline.}}} \\
LLaMA                   & + 300K PandaLM data &           & N            & 68.61       & 74.78         \\
\midrule
\multicolumn{6}{l}{\textbf{\textit{change to JudgeLM data.}}} \\
LLaMA                   & +3.5k JudgeLM data  &           & N            & 71.30       & 69.59         \\
\midrule
\multicolumn{6}{l}{\textbf{\textit{w/ the proposed methods.}}} \\
LLaMA                   & +3.5k JudgeLM data  & +swap aug & N            & 72.41       & 73.50         \\
LLaMA                   & +3.5k JudgeLM data  & +ref sup  & Y            & 75.15       & 74.08         \\
LLaMA                   & +3.5k JudgeLM data  & +ref drop & Y            & 75.93       & 74.77         \\
\midrule
\multicolumn{6}{l}{\textbf{\textit{w/ all proposed methods.}}} \\
LLaMA                   & +3.5k JudgeLM data  & + all     & N            & 75.01       & 76.28         \\
LLaMA                   & +3.5k JudgeLM data  & + all     & Y            & 78.10       & 79.50         \\
\bottomrule
\end{tabular}
\label{tab: abla_on_panda}
\end{table}

\myparagraph{Differences between PandaLM and JudgeLM}
To fairly assess the impact of our proposed dataset and methods, we conducted experiments using the PandaLM baseline with LLaMA-7B as the base model. As shown in Table~\ref{tab: abla_on_panda}, our dataset and methods both provide significant improvements.

Compared with PandaLM, our method has these different novelties:
\begin{itemize}
    \item We introduce \textbf{a high-quality, large-scale dataset for judge models}, enriched with diverse seed tasks, LLMs-generated answers, and detailed judgments from GPT-4, laying the foundation for future LLMs evaluating research.
    \item We analyze the \textbf{biases inherent to LLM judge fine-tuning} and introduce \textbf{a series of methods} to address them. Our methods significantly improve the consistency of the model in different cases, making the JudgeLM more reliable and flexible.
    \item The proposed \textbf{judging pattern, i.e., grading, judging, and reasoning}, makes judging efficient, which only needs little time to grade and judge, and generates time-consuming reasons optionally.
\end{itemize}

Furthermore, other contributions which differ from PandaLM are as follows:
\begin{itemize}
    \item We analyze the \textbf{\revision{finer} scaling ability} of the language model judge, and \revision{the scales of training data}, i.e., JudgeLM-7B, JudgeLM-13B, JudgeLM-33B, \revision{3.5k-data, 10k-data, 30k-data, and 100k-data}, for evaluating LLMs in open-ended scenarios. The JudgeLM-33B\revision{-100K} even \textbf{exceeds the humans' judging agreement}.
    \item The proposed JudgeLM presents \textbf{extended capabilities} as shown in Fig. 1b, including grading single answers, judging multiple answers, judging multimodal models, multi-turn chat, and judging with injected paragraphs.
\end{itemize}
Specifically, the proposed JudgeLM presents generalization ability to seven different judging areas:
\begin{itemize}
\item Generalize to \textbf{19 various judging tasks}, including coding, common-sense, math, roleplay, writing, etc.

\item Generalize to \textbf{human-annotated benchmarks}, such as the PandaLM test set and MM-Vet benchmark.

\item Generalize to \textbf{multimodal judging benchmark}, such as MM-Vet benchmark.

\item Generalize to \textbf{retrieval-format benchmark} which injects original reference answers into paragraphs with different words.

\item Generalize to \textbf{multiple-format benchmark} which judges multiple answers at the same time.

\item Generalize to \textbf{single-answer grading}.

\item Generalize to \textbf{multi-turn chat about judgments}.
\end{itemize}

\myparagraph{GPT-4 Distilled Data}
Evaluating Large Language Models (LLMs) in open-ended scenarios is challenging because existing benchmarks and metrics can not measure them comprehensively. Because the GPT-4-based judge can judge LLMs like a human expert, much concurrent work delving into curating training data~\citep{li2023autoj, kim2023prometheus, cui2023ultrafeedback, xu2023instructscore, jiang2023tigerscore, wang2023pandalm} and benchmarks~\citep{lambert2024rewardbench, zheng2023chatbot-arena} with the help of the closed-source LLMs, such as GPT-4 and GPT-3.5.

However, the GPT-4 teacher also faces inherent biases. Firstly, GPT-4’s training data may contain cultural, societal, and linguistic biases. These biases may influence the GPT-4’s judgment and lead to skewed evaluations. To mitigate these possible biases, the authors of this work are involved in double-checking to ensure the judgments from the GPT-4 teacher are accurate, objective, and unbiased. Secondly, the GPT-4's judgments also contain position bias and knowledge bias~\citep{zheng2023chatbot-arena}. To address these problems, we introduce swap augmentation and reference support in fine-tuning LLMs as judges, which significantly improves the consistency and accuracy of fine-tuned judges.

\begin{figure}[htp]
    \begin{center}
       \includegraphics[width=.88\linewidth]{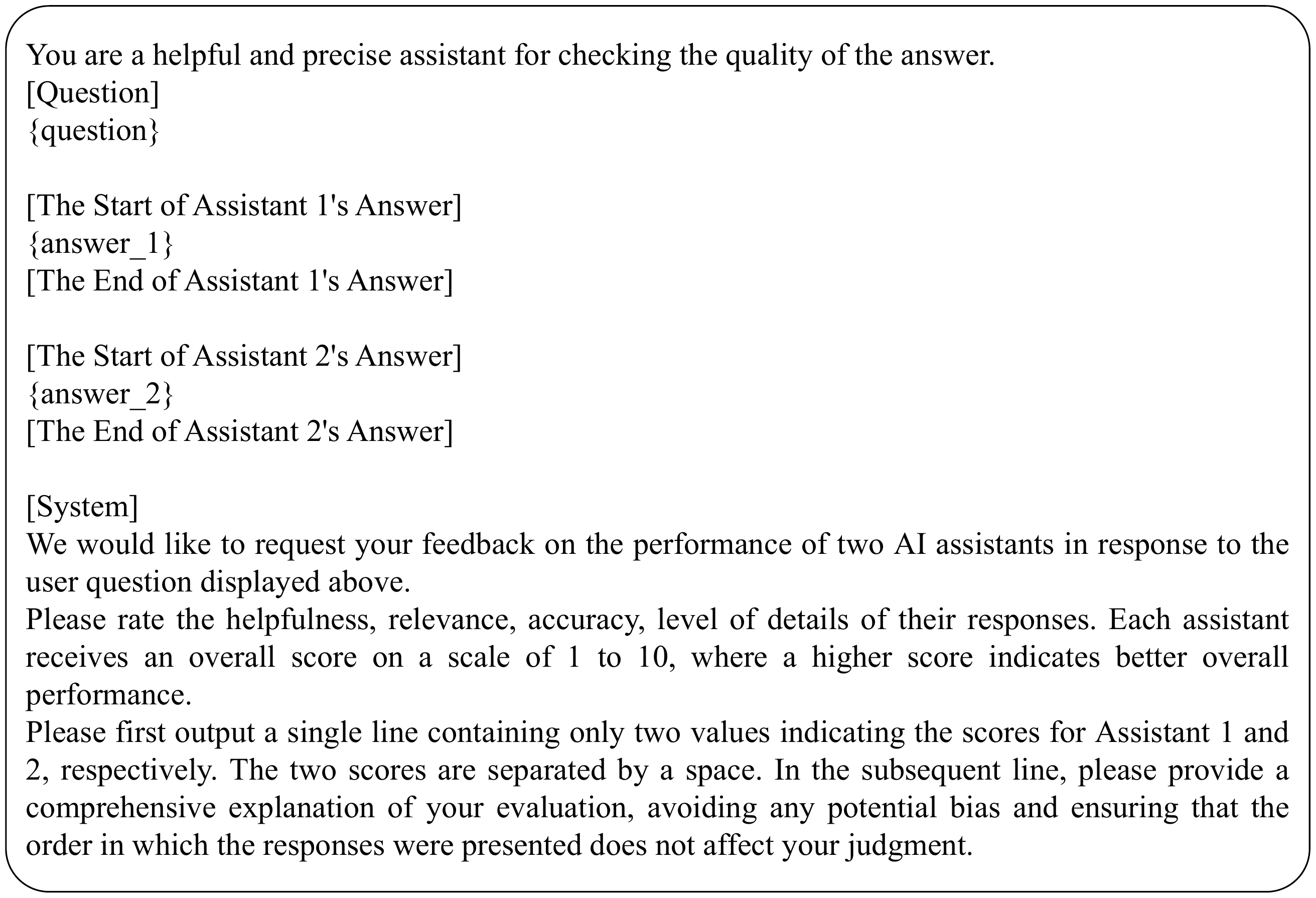}
    \end{center}
    \caption{The template for judging answers without the reference.}
    \label{fig: temp_wo_ref}
\end{figure}

\begin{figure}[htp]
    \begin{center}
       \includegraphics[width=.88\linewidth]{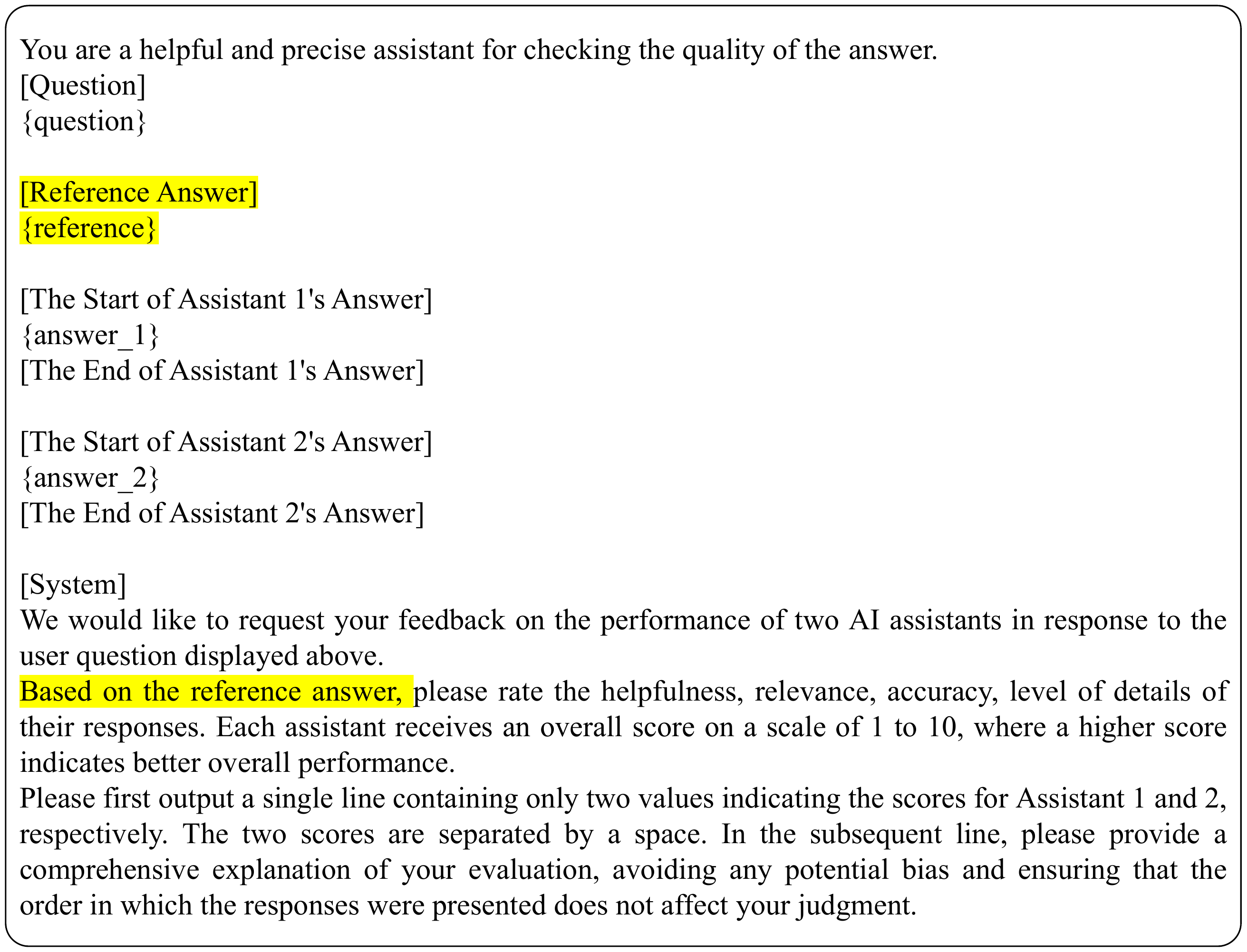}
    \end{center}
    \caption{The template for judging answers with the reference.}
    \label{fig: temp_w_ref}
\end{figure}

\begin{figure}[htp]
    \begin{center}
        \includegraphics[width=0.9\linewidth]{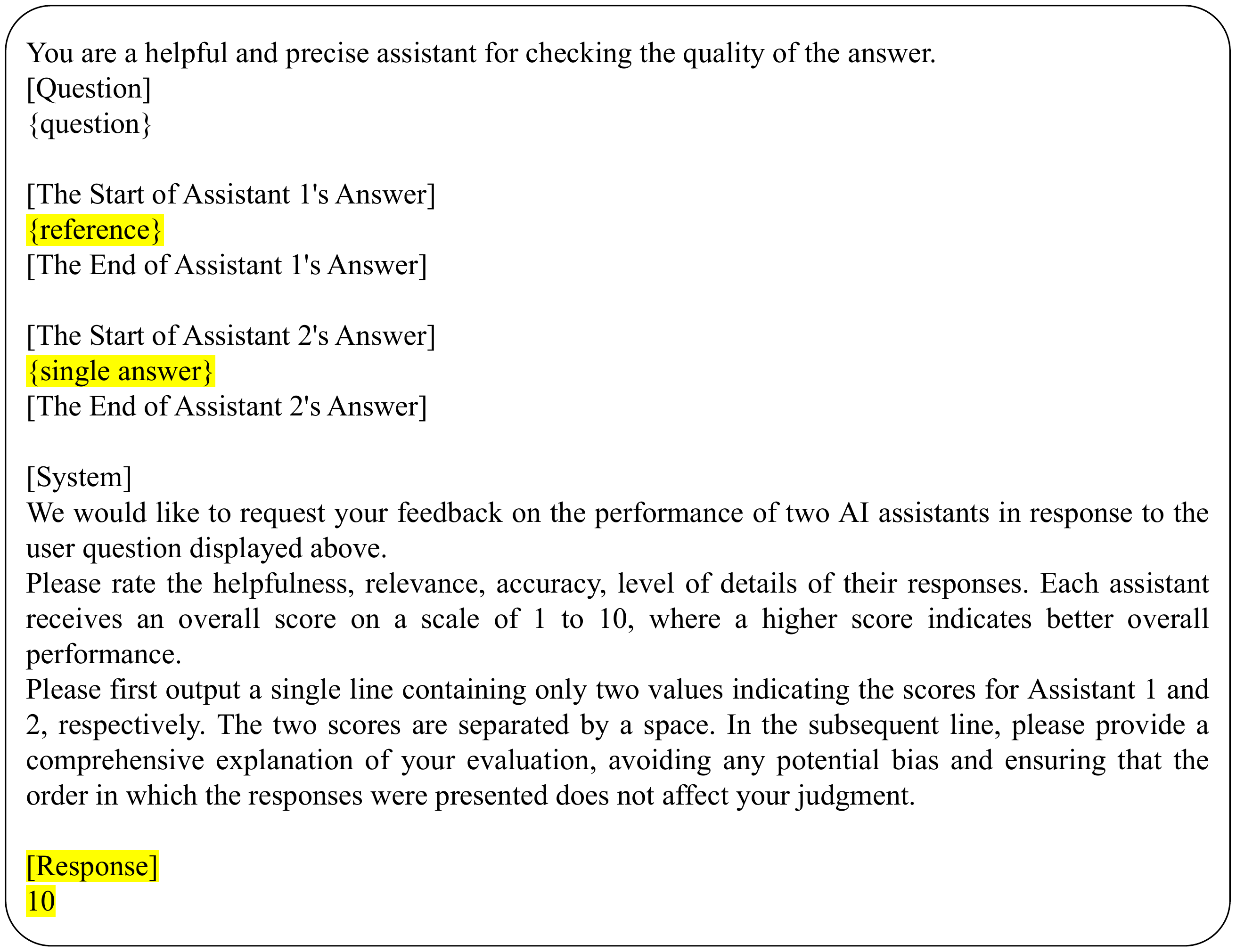}
    \end{center}
    \caption{The template for grading a single answer. We set the reference answer in the position of Answer 1. Then, we set the score of the reference answer to 10. Last, the \name{} outputs the score of the single answer with such a prior.}
    \label{fig: temp_single_ans}
\end{figure}

\begin{figure}[htp]
    \begin{center}
        \includegraphics[width=0.9\linewidth]{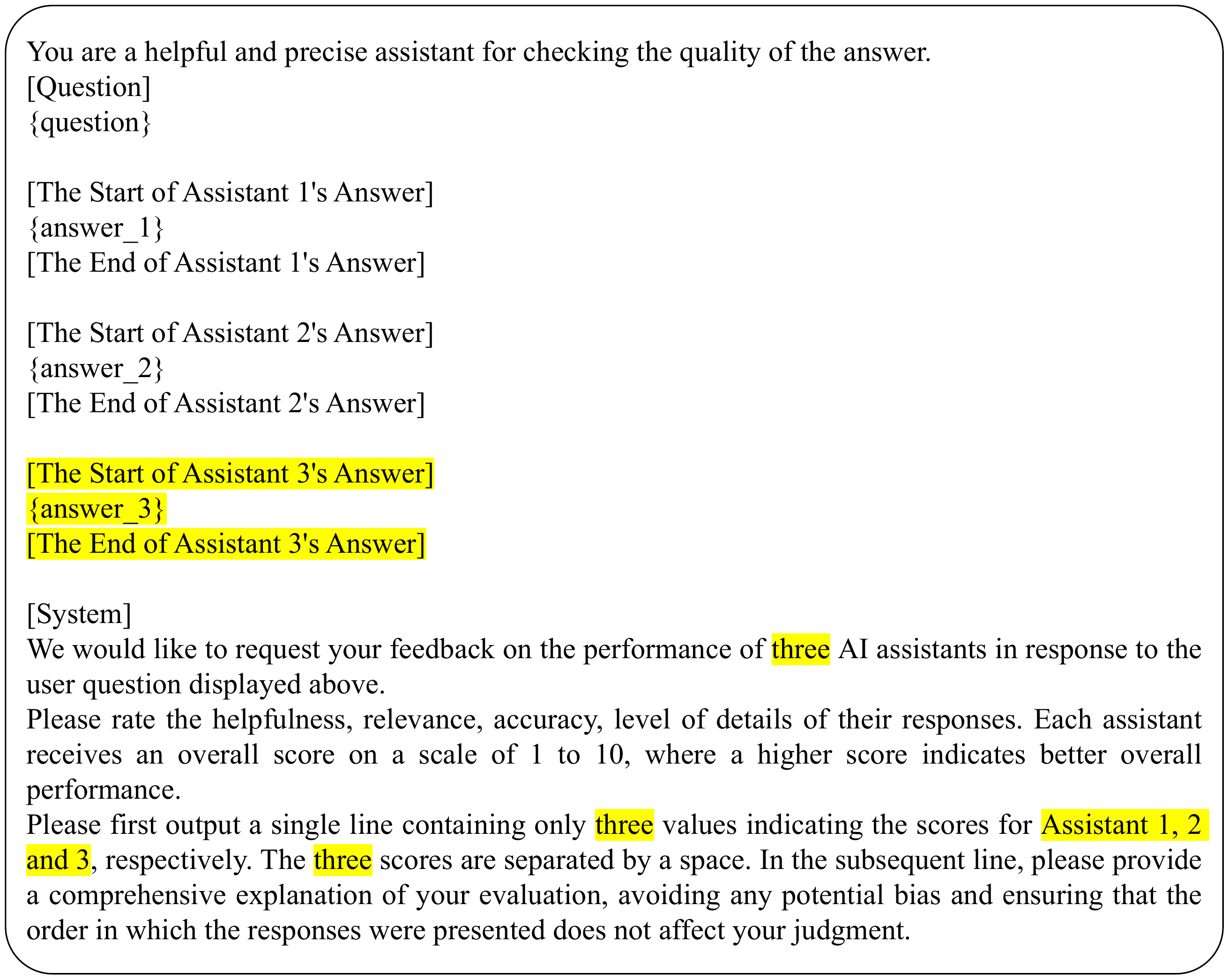}
    \end{center}
    \caption{The template for judging multiple answers.}
    \label{fig: temp_multi_ans}
\end{figure}

\begin{figure}[htp]
    \begin{center}
        \includegraphics[width=0.9\linewidth]{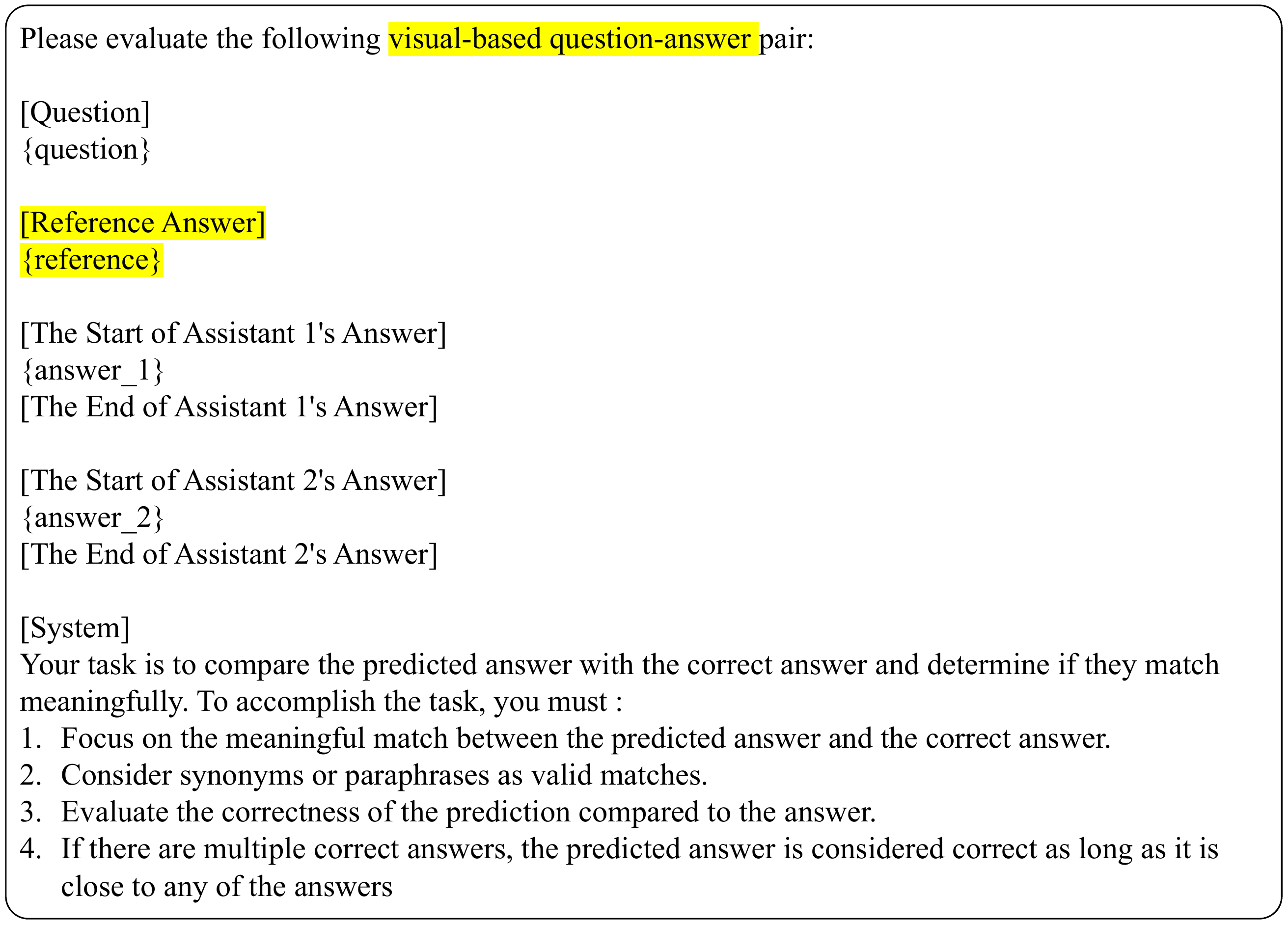}
    \end{center}
    \caption{The template for multimodal judging.}
    \label{fig: temp_vqa}
\end{figure}

\myparagraph{\revision{Does JudgeLM favor GPT-4?}}
\revision{Thank you for discussing this interesting phenomenon. We carefully conducted experiments between GPT-4 and GPT-3.5 to analyze the potential favoring of GPT-4 answers by JudgeLM. As shown in Table ~\ref{tab: gpt4vsgpt3p5}, GPT-4 wins 59.0\% of pairwise comparisons, with only 16.2\% of wins for GPT-3.5. We further conducted a sampling analysis to identify the following contributing factors:}
\begin{itemize}

\item \revision{Alignment with Prompt Requirements: As shown in the judging templates (Fig.~\ref{fig: temp_wo_ref}, Fig.~\ref{fig: temp_w_ref}, Fig.~\ref{fig: temp_single_ans}, Fig.~\ref{fig: temp_multi_ans}, and Fig.~\ref{fig: temp_vqa}), JudgeLM evaluates responses based on structured criteria, such as relevance and level of detail. Our analysis shows that GPT-4 answers tend to align better with these requirements compared to GPT-3.5 answers.}

\item \revision{Inherent Quality Difference: A sampling analysis of 500 cases (approximately 16.9\% of the GPT-4 win cases) showed that in 482 instances (96.4\%), GPT-4 answers were more accurate, detailed, and contextually relevant than GPT-3.5 answers. This analysis underscores that GPT-4 win cases are overwhelmingly attributed to the higher answer quality, rather than the preference from JudgeLM.}

\end{itemize}

\revision{Moreover, the sample analysis shows that when the quality of the answers to GPT-4 and GPT-3.5 are similar, JudgeLM tends to give similar scores, resulting in a Tie. These results indicate that GPT-4 answers win more due to their better relevance, high level of detail, and high quality.}

\begin{table}[htp]
\centering
\caption{\revision{Quantitative comparison results between GPT-4 and GPT-3.5.}}
\begin{tabular}{lccc}
\toprule
\textbf{}         & GPT4 win & Tie     & GPT3.5 win \\
\midrule
GPT4 v.s. GPT-3.5 & 59.00\%  & 24.80\% & 16.20\%   \\
\bottomrule
\end{tabular}
\label{tab: gpt4vsgpt3p5}
\end{table}

\myparagraph{\revision{Reliability of GPT-4-annotated Data.}}
\revision{We conduct more experiments to further evaluate the reliability of GPT-4's annotation. In this experiment setting, we set annotations provided by Human 1 in our benchmark as ground truth. Then we calculate the agreement of JudgeLM-33B, GPT-4, and Human 2 as shown in Table~\ref{tab: data_reliability}.  
These results demonstrate that GPT-4 achieves higher agreement with the ground truth compared to Human 2. llm-as-a-judge~\citep{zheng2023chatbot-arena} shows similar results, which means GPT-4 teacher's judgments can align with human evaluators closely and can serve as a human-like teacher judge to provide high-quality and reliable judgments.}

\begin{table}[htp]
\centering
\caption{\revision{Agreement evaluation of JudgeLM-33B, GPT-4, and Human, with another Human's judgments as ground truth.}}
\begin{tabular}{lccc}
\toprule
\textbf{}              & JudgeLM-33B & GPT-4 & Human 2 \\
\midrule
Ground Truth (Human 1) & 90.72       & 84.48 & 79.82  \\
\bottomrule
\end{tabular}
\label{tab: data_reliability}
\end{table}

\myparagraph{\revision{Future Work about Training Queries Distribution.}}
\revision{The distribution of training queries is a key factor in shaping the evaluation capabilities of Judge LLMs. While our current dataset was designed to ensure task diversity, MixEval~\citep{ni2024mixeval} and MixEval-X~\citep{ni2024mixevalx} give us more insight, i.e., aligning the query distribution more closely with real-world user queries could further enhance the model’s fairness and relevance. We leave it as a promising future work.}

\subsection{Prompt templates}
We list all the prompt templates we used.

\subsection{Case Studies}
We list several case studies.

\begin{figure}[htp]
    \begin{center}
       \includegraphics[width=.88\linewidth]{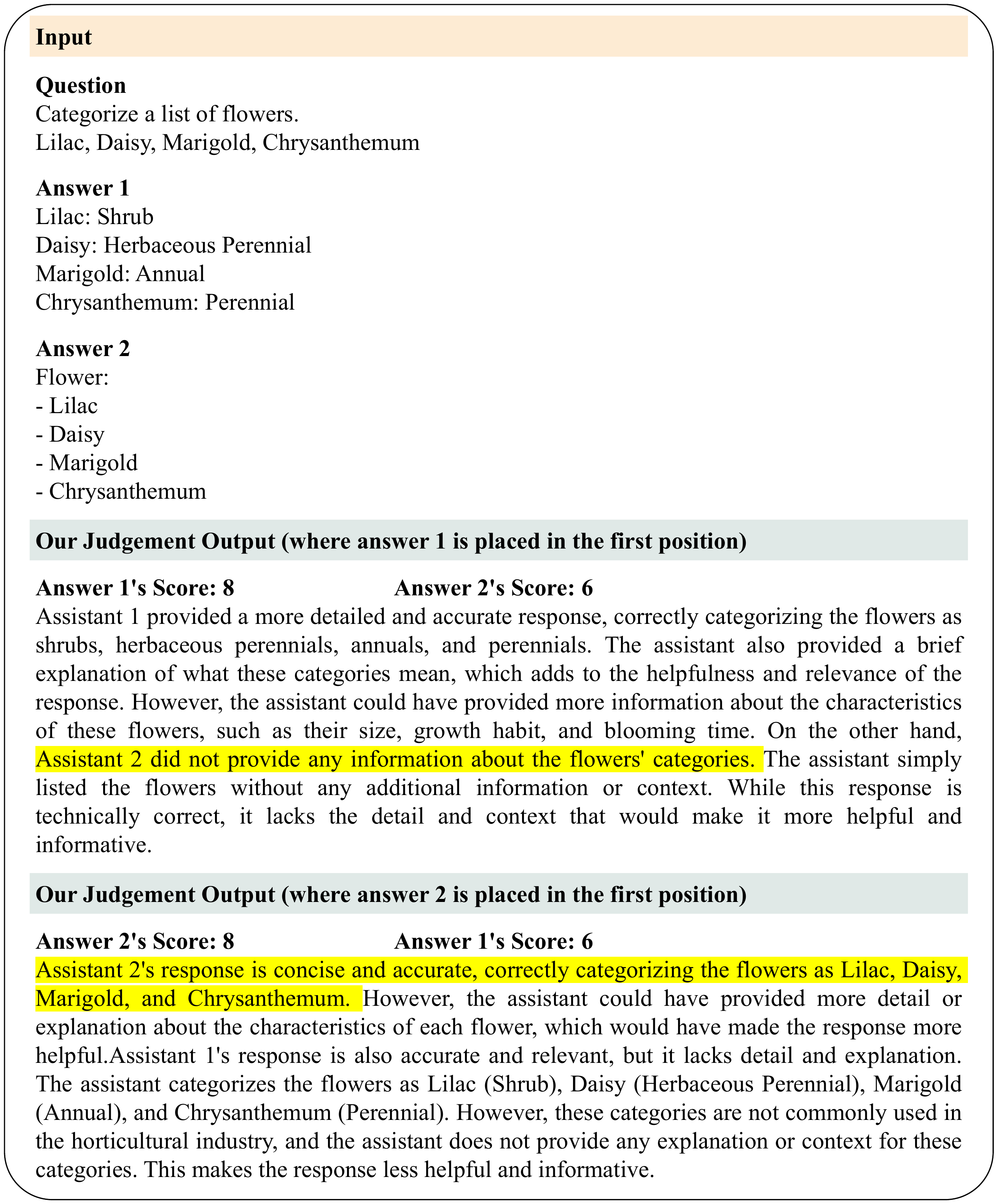}
    \end{center}
    \caption{Bad judgment caused by position bias. The answer placed in the first position always gets a higher score. The judge models generate reasons as possible from the perspective of making the scores reasonable.}
    \label{fig: position_bias}
\end{figure}

\begin{figure}[htp]
    \begin{center}
       \includegraphics[width=.88\linewidth]{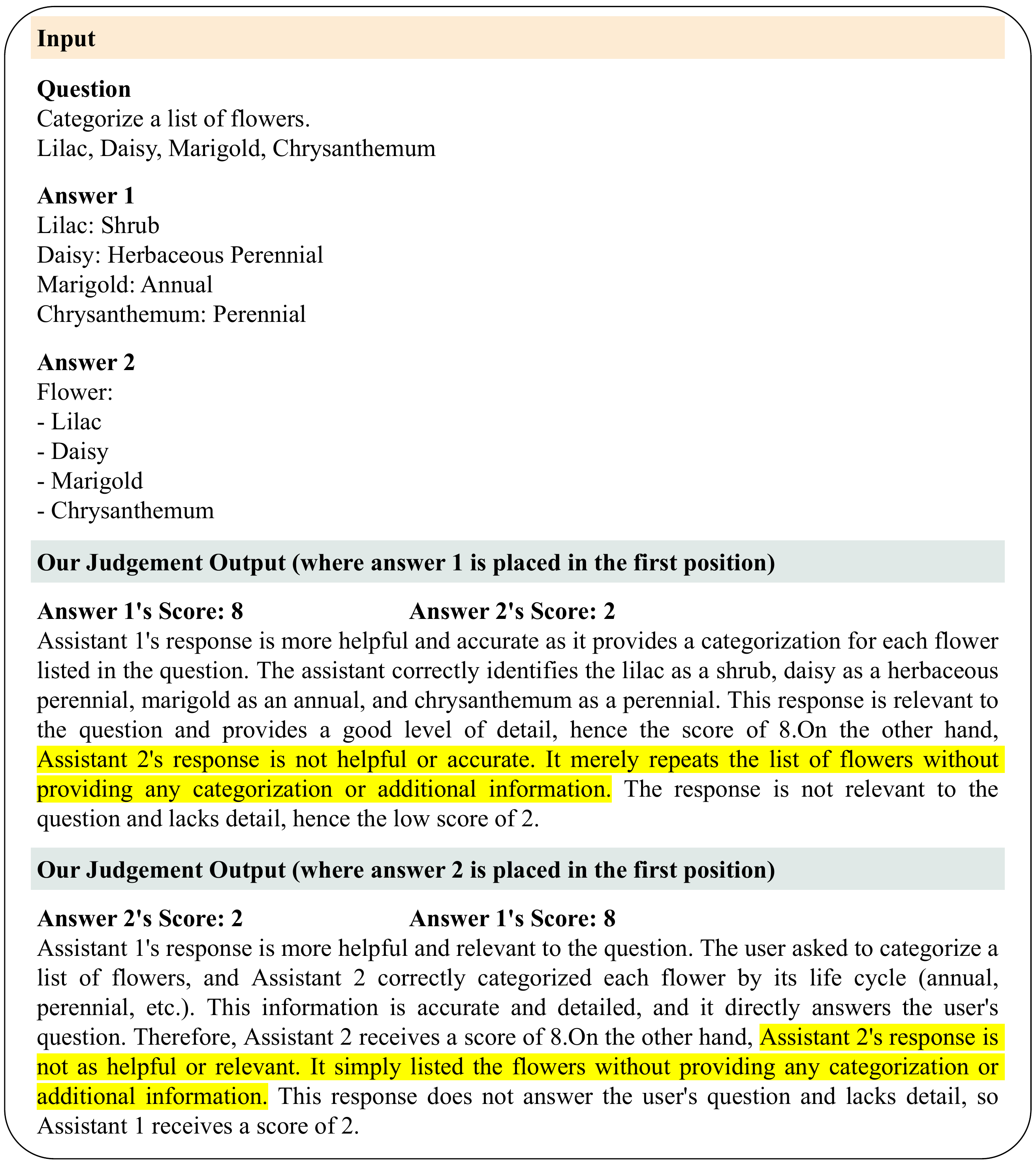}
    \end{center}
    \caption{Good judgment generated by the judge fine-tuned with swap augmentation. The judge can give judgments based on the content of answers rather than a certain position. The reason is convincing and reasonable.}
    \label{fig: swap_aug}
\end{figure}

\begin{figure}[htp]
    \begin{center}
       \includegraphics[width=.88\linewidth]{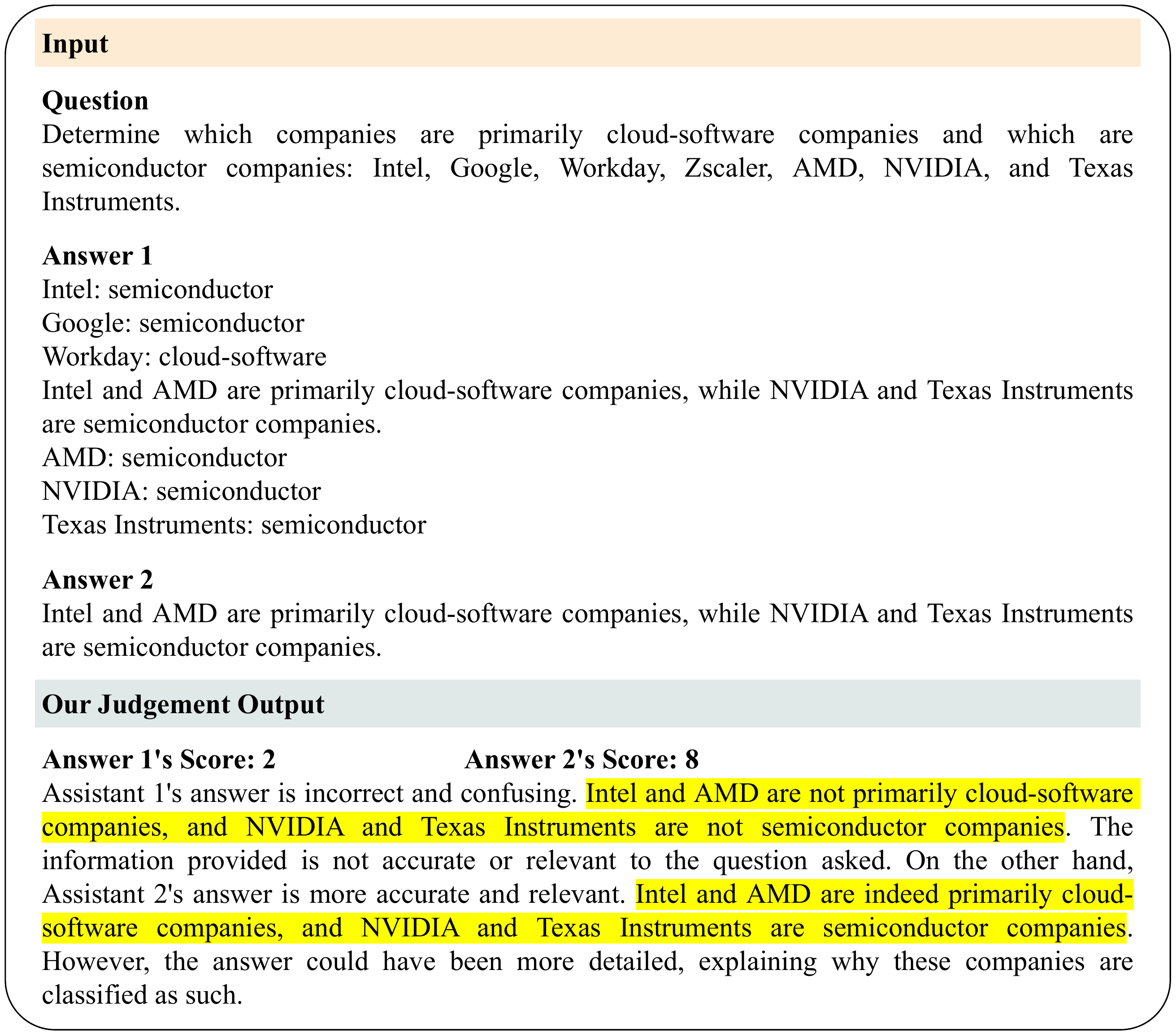}
    \end{center}
    \caption{Bad judgment caused by knowledge bias. This seed task is out of the judge model's pre-trained knowledge, so it can not be correctly judged by the judge model. The judge model gives contradictory reasons in the judgment.}
    \label{fig: knowledge_bias}
\end{figure}

\begin{figure}[htp]
    \begin{center}
       \includegraphics[width=.88\linewidth]{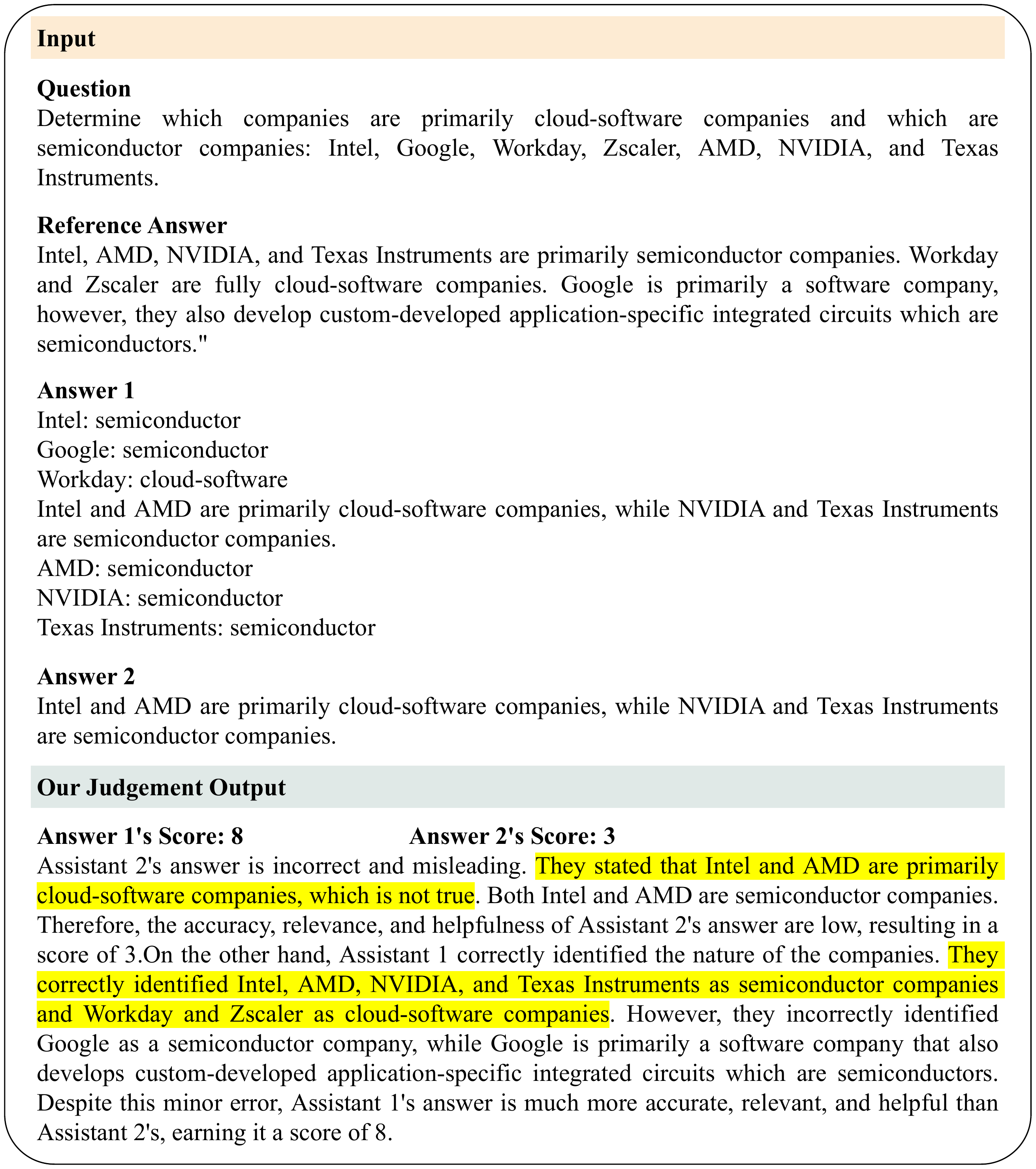}
    \end{center}
    \caption{Good judgment generated by the judge model fine-tuned with reference support. Even though the judge model itself lacks related information, it can also give a reasonable judgment with the reference answer.}
    \label{fig: reference_sup}
\end{figure}

\begin{figure}[htp]
    \begin{center}
       \includegraphics[width=.88\linewidth]{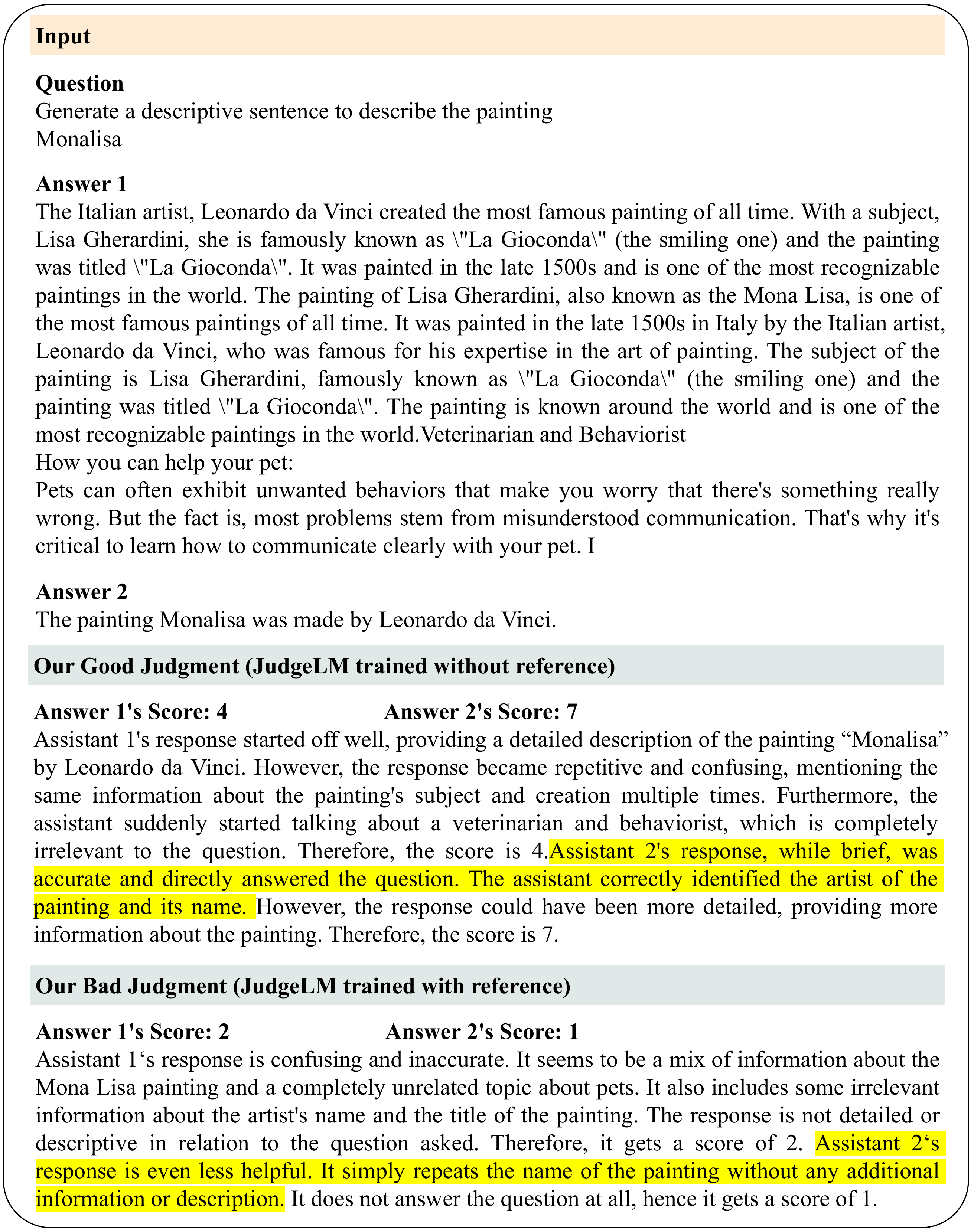}
    \end{center}
    \caption{Bad judgment caused by format bias. For judging without reference, the judge model trained without reference is matched, so it performs well. However, the judge model trained with reference is mismatched, so it performs badly.}
    \label{fig: mismatch_wo}
\end{figure}

\begin{figure}[htp]
    \begin{center}
        \includegraphics[width=.88\linewidth]{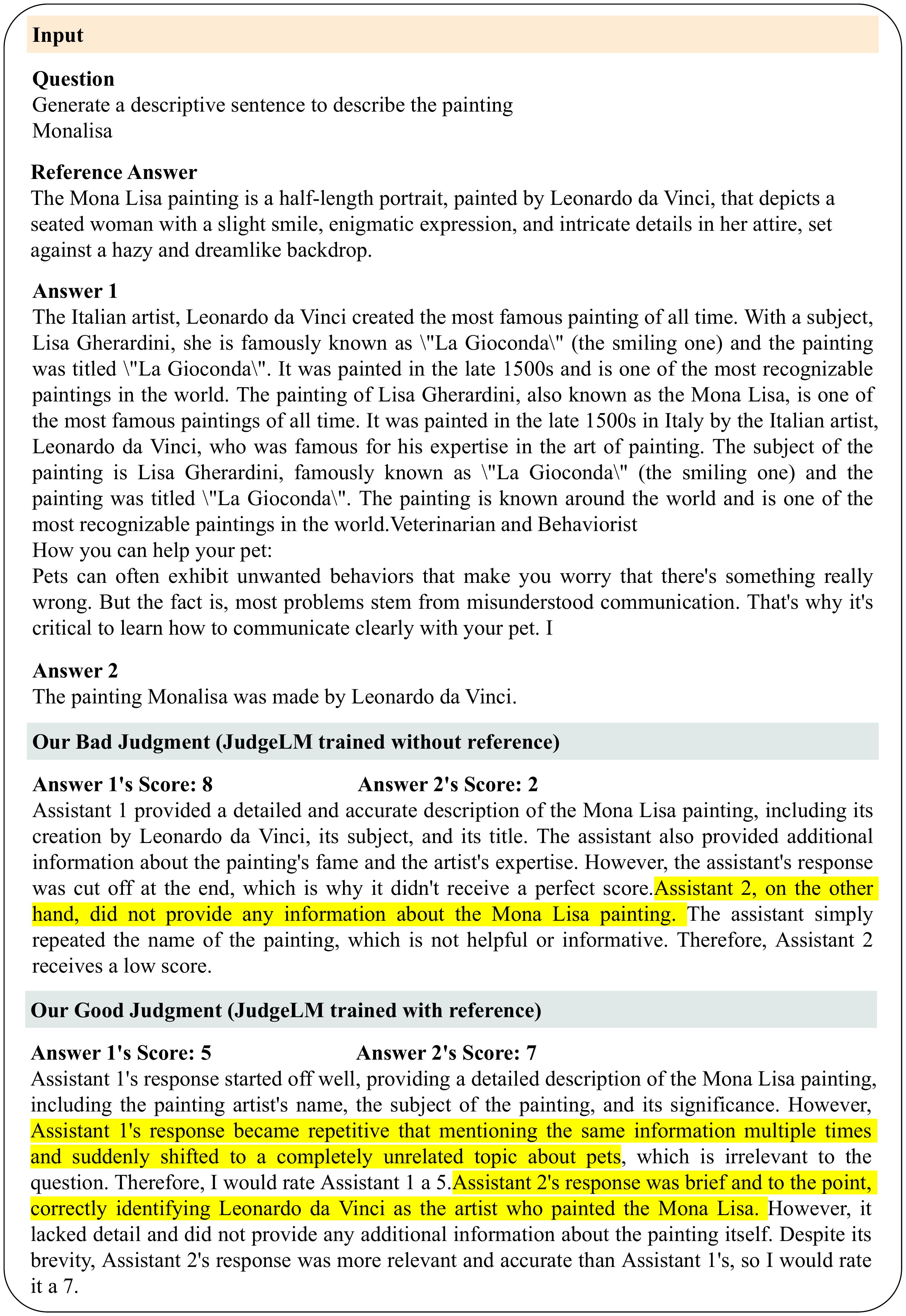}
    \end{center}
    \vspace{-0.1 in}
    \caption{Bad judgment caused by format bias. For judging with reference, the judge model trained with reference is matched, so it performs well. However, the judge model trained without reference is mismatched, so it performs badly.}
    \label{fig: mismatch_w}
\end{figure}

\begin{figure}[htp]
    \begin{center}
        \includegraphics[width=.84\linewidth]{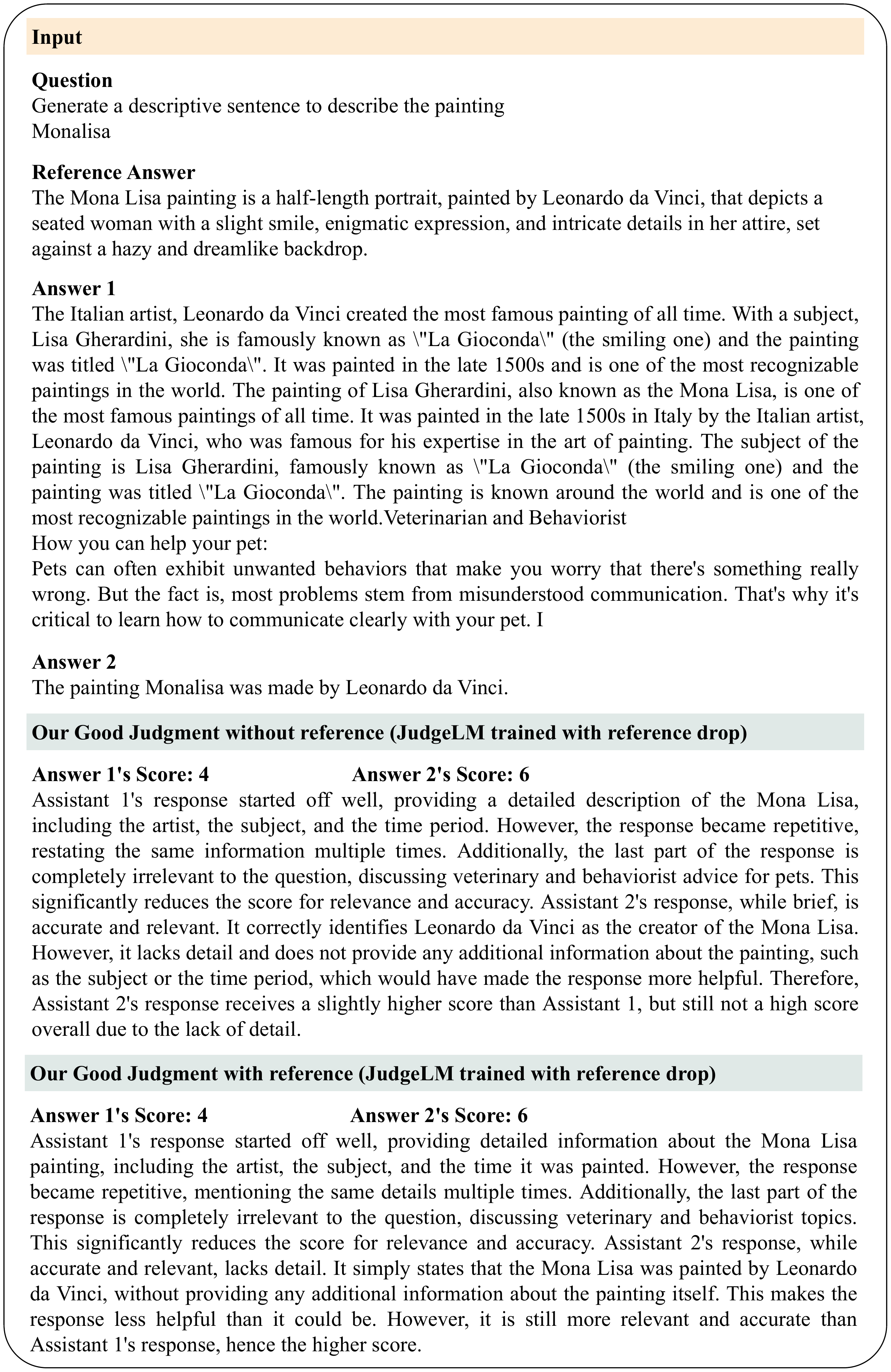}
    \end{center}
    \vspace{-0.1 in}
    \caption{Good judgment generated by the judge model with reference drop, which addresses the preference for specific fine-tuning formats and gives fair judgments with or without reference. }
    \label{fig: ref_drop}
\end{figure}

\begin{figure}[htp]
    \begin{center}
       \includegraphics[width=.88\linewidth]{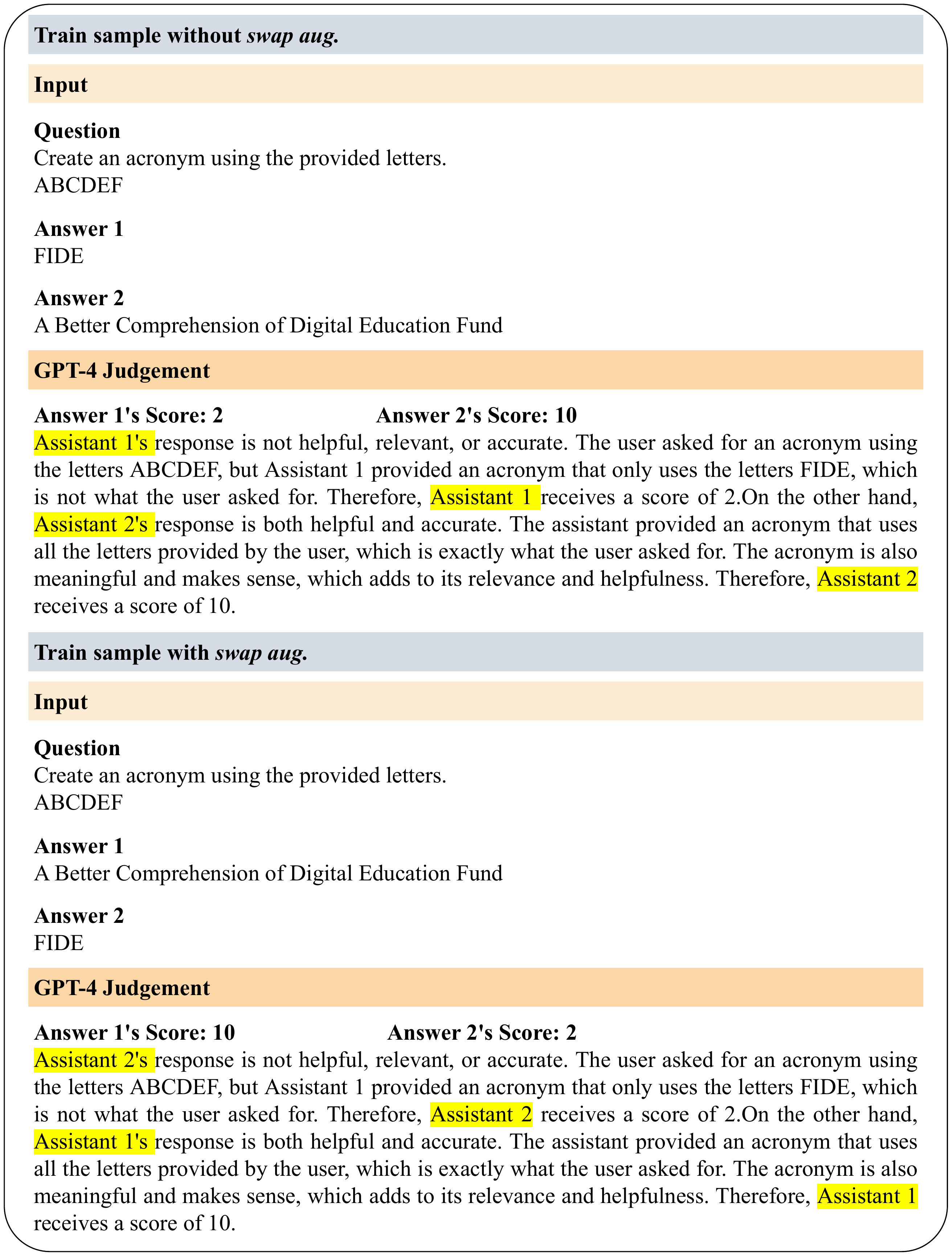}
    \end{center}
    \caption{An illustration of swap augmentation. We use swap augmentation to exchange the positions of answers, and our GPT-4-generated judgments can be modified correspondingly easily due to their structure.}
    \label{fig: swap_aug_io}
\end{figure}

\begin{figure}[htp]
    \begin{center}
       \includegraphics[width=.88\linewidth]{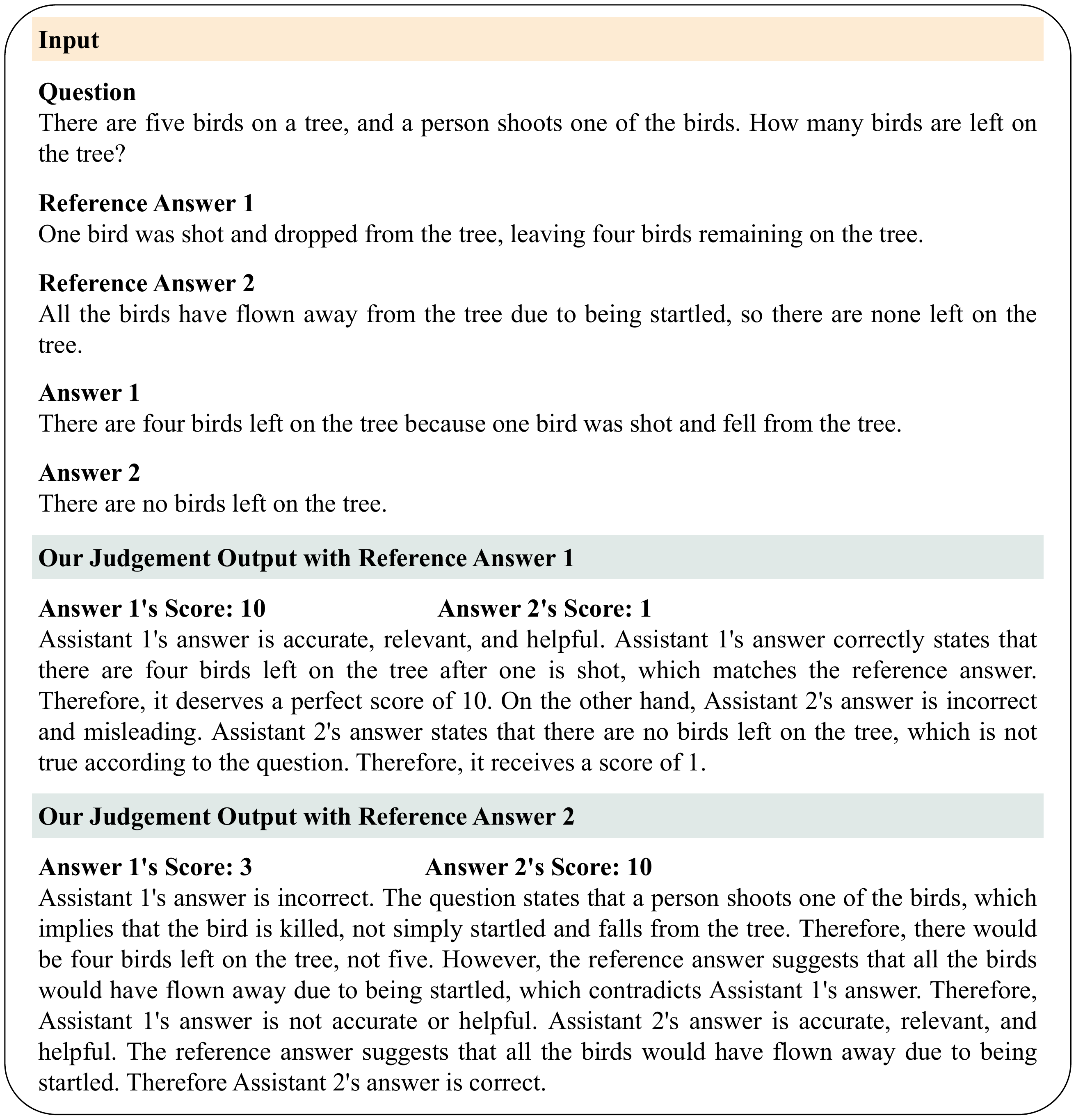}
    \end{center}
    \caption{An illustration of changing the reference answer to control model preference. When we change to a different reference answer, the model turns to prefer another answer.}
    \label{fig: change_reference}
\end{figure}

\begin{figure}[htp]
    \begin{center}
       \includegraphics[width=.88\linewidth]{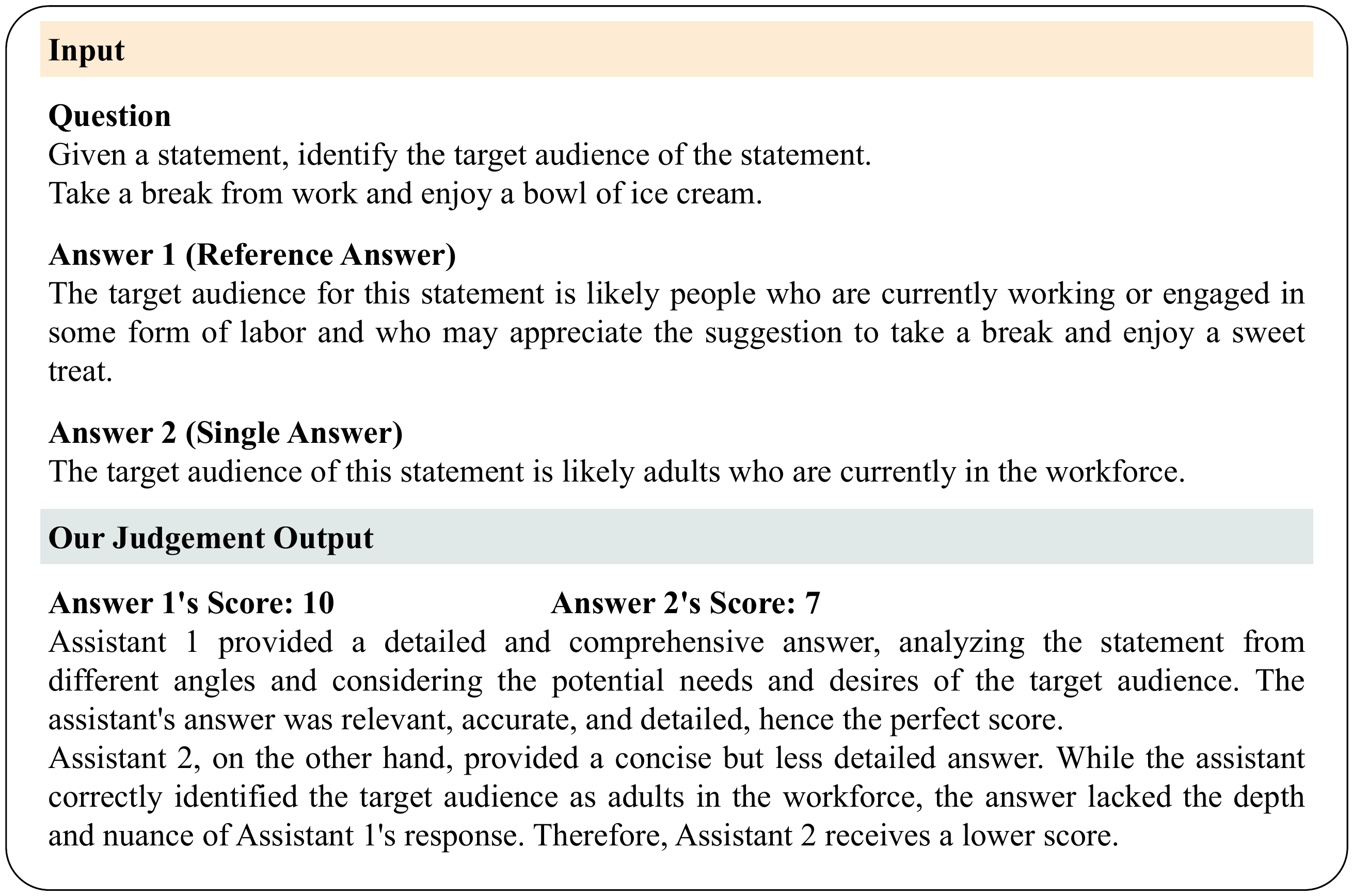}
    \end{center}
    \caption{An illustration of grading a single answer. }
    \label{fig: grading_single_ans}
\end{figure}

\begin{figure}[htp]
    \begin{center}
       \includegraphics[width=.88\linewidth]{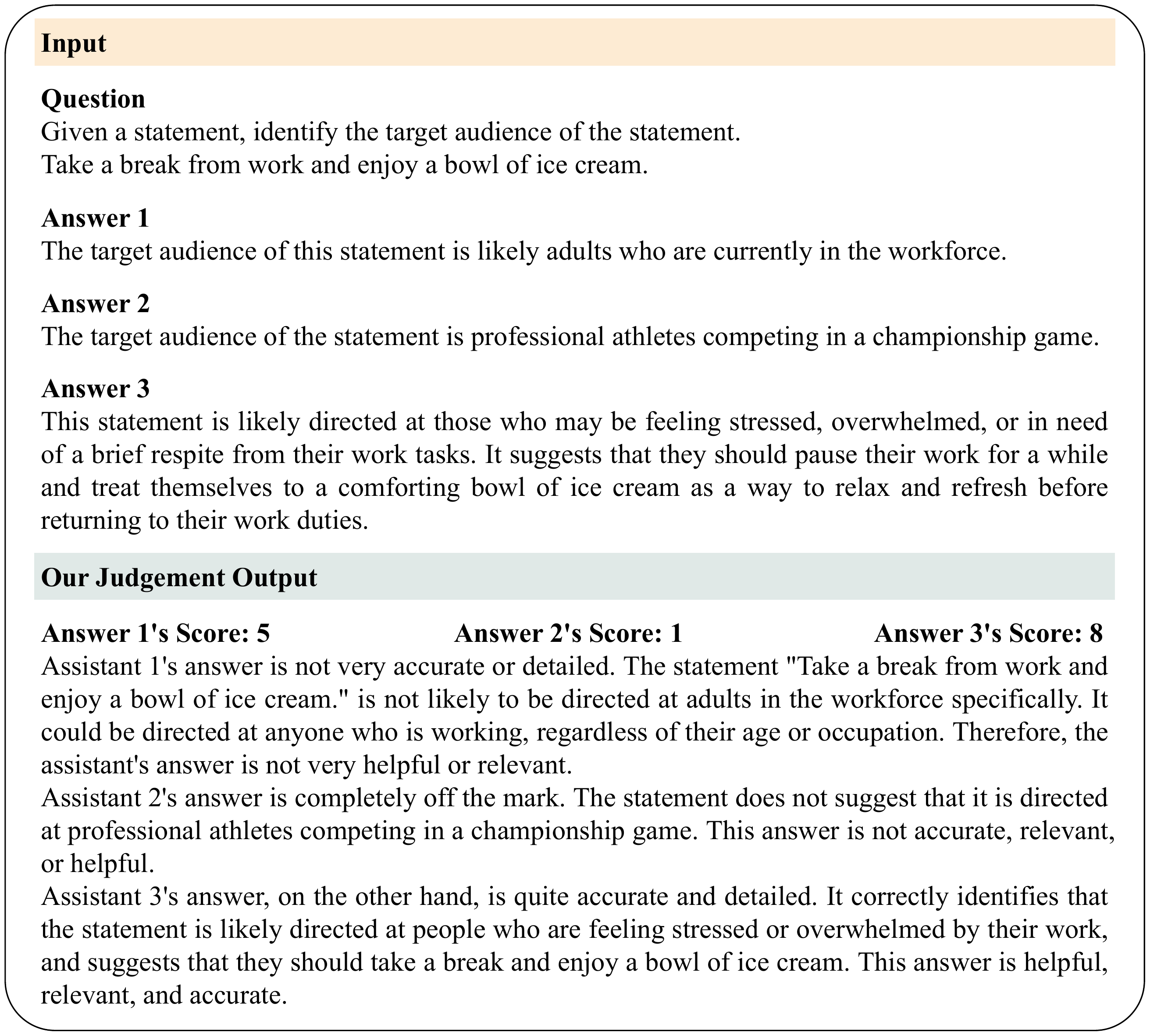}
    \end{center}
    \caption{An illustration of judging multiple answers. }
    \label{fig: rank_multi_ans}
\end{figure}

\begin{figure}[htp]
    \begin{center}
       \includegraphics[width=.88\linewidth]{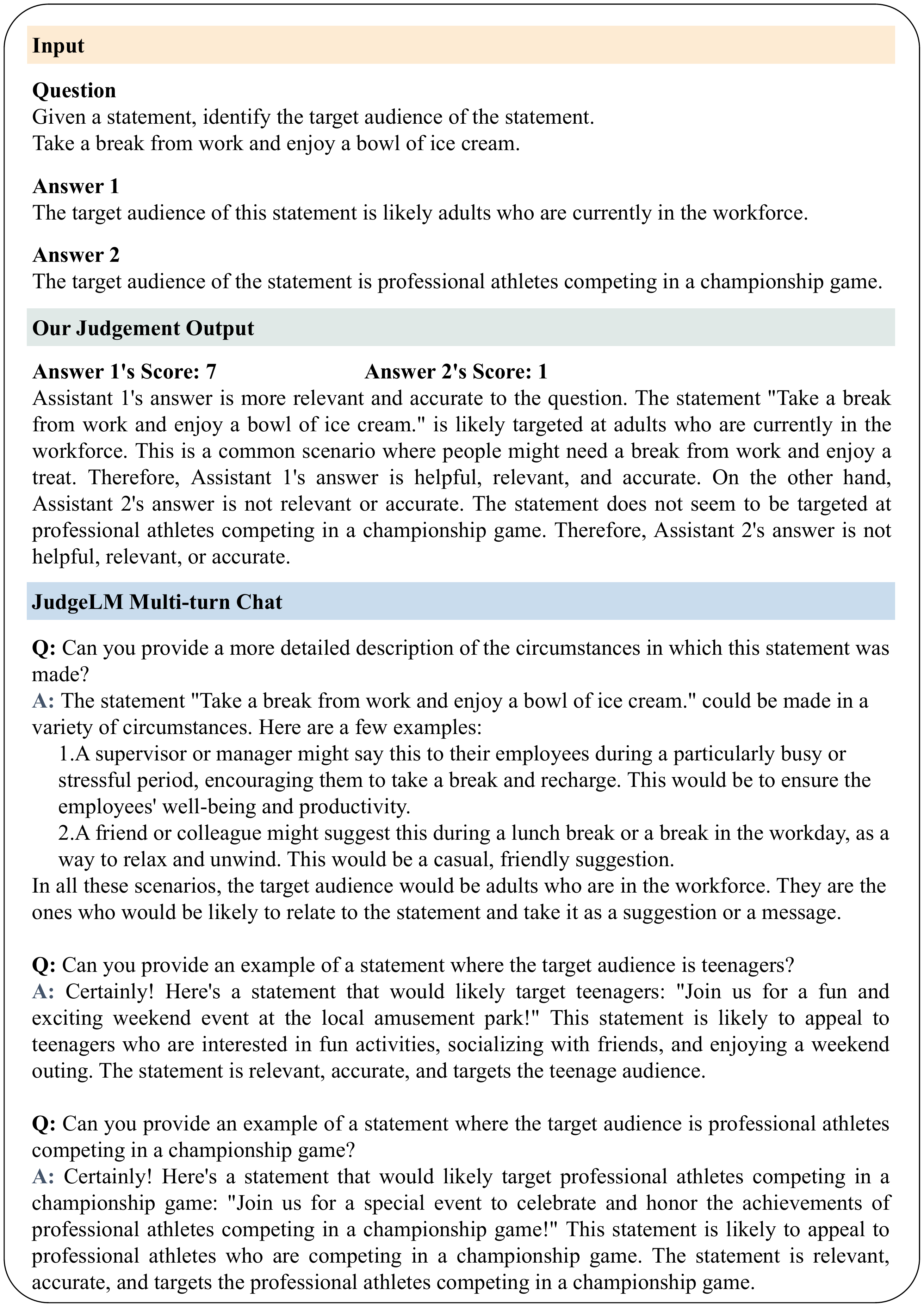}
    \end{center}
    \caption{An illustration of multi-turn chat. Users can get more details, advice, examples, etc., by chatting with \name{}.}
    \label{fig: multi_turn_chat}
\end{figure}

\begin{figure}[htp]
    \begin{center}
       \includegraphics[width=.88\linewidth]{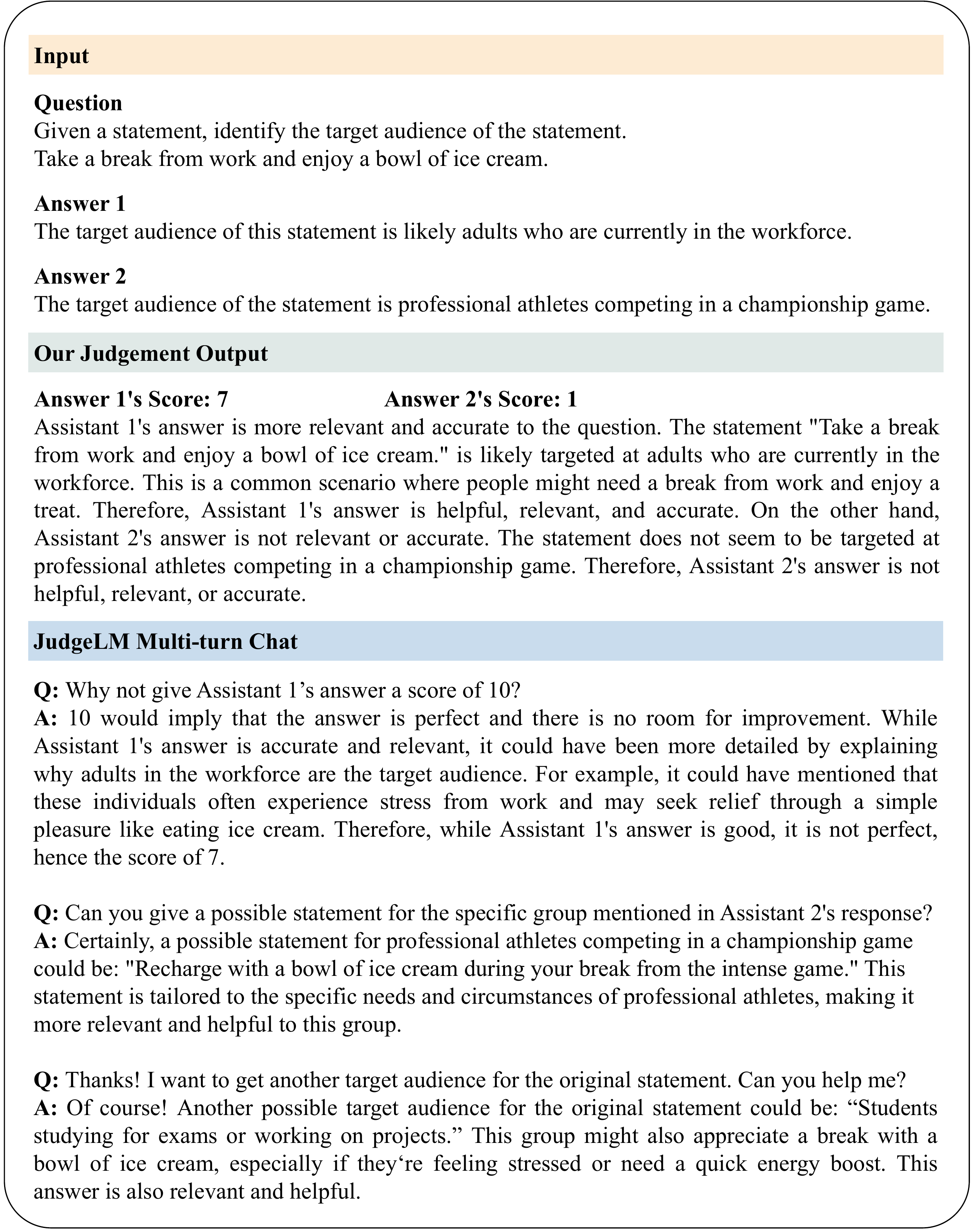}
    \end{center}
    \caption{\add{An additional illustration of multi-turn chat. Users can get more details, advice, examples, etc., by chatting with \name{}.}}
    \label{fig: multi_turn_chat_additional}
\end{figure}

\begin{figure}[htp]
    \begin{center}
        \includegraphics[width=0.9\linewidth]{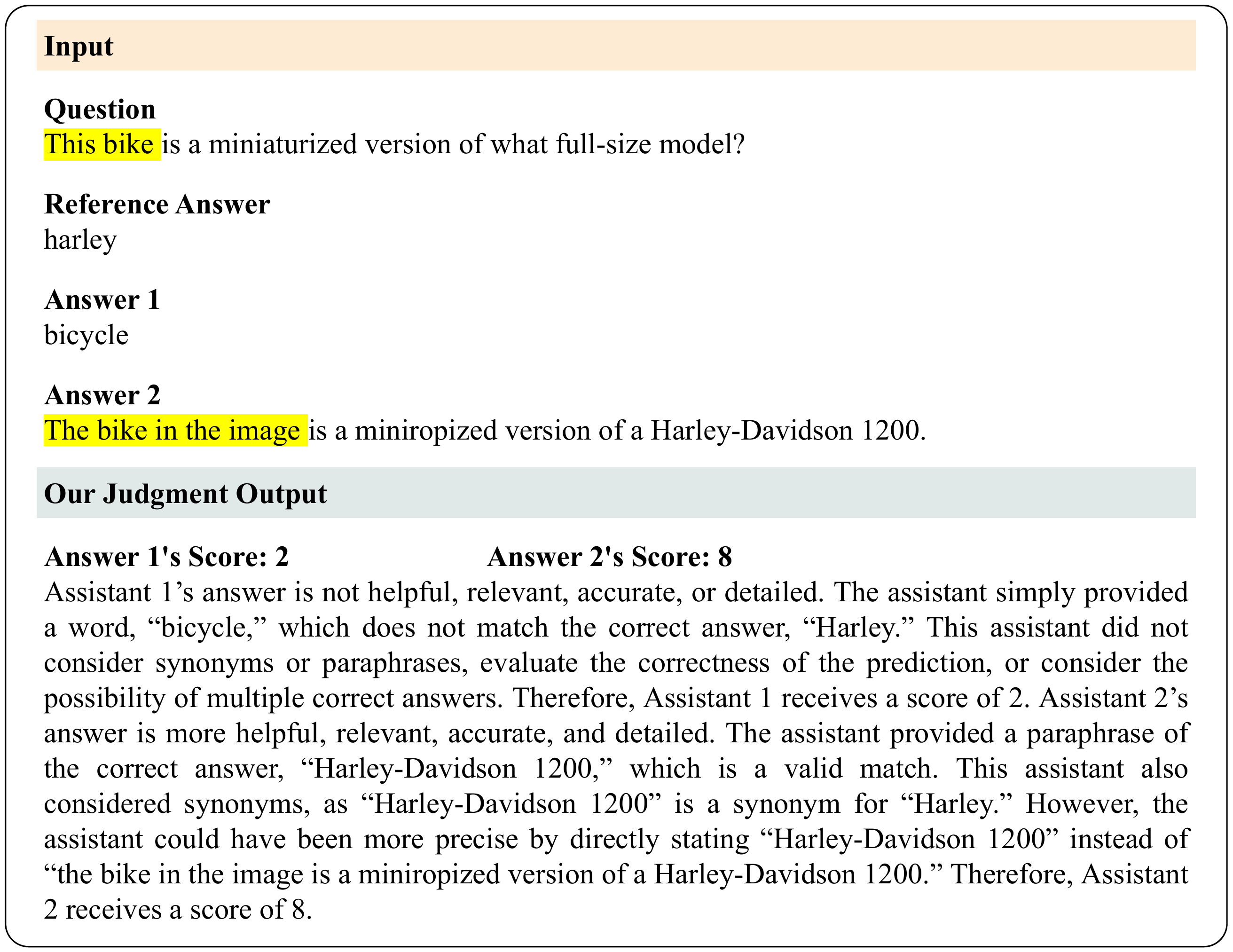}
    \end{center}
    \caption{An illustration of multimodal judging. Our \name{} has the capacity to judge the VQA task without images.}
    \label{fig: vqa}
\end{figure}

\begin{figure}[htp]
    \begin{center}
        \includegraphics[width=0.9\linewidth]{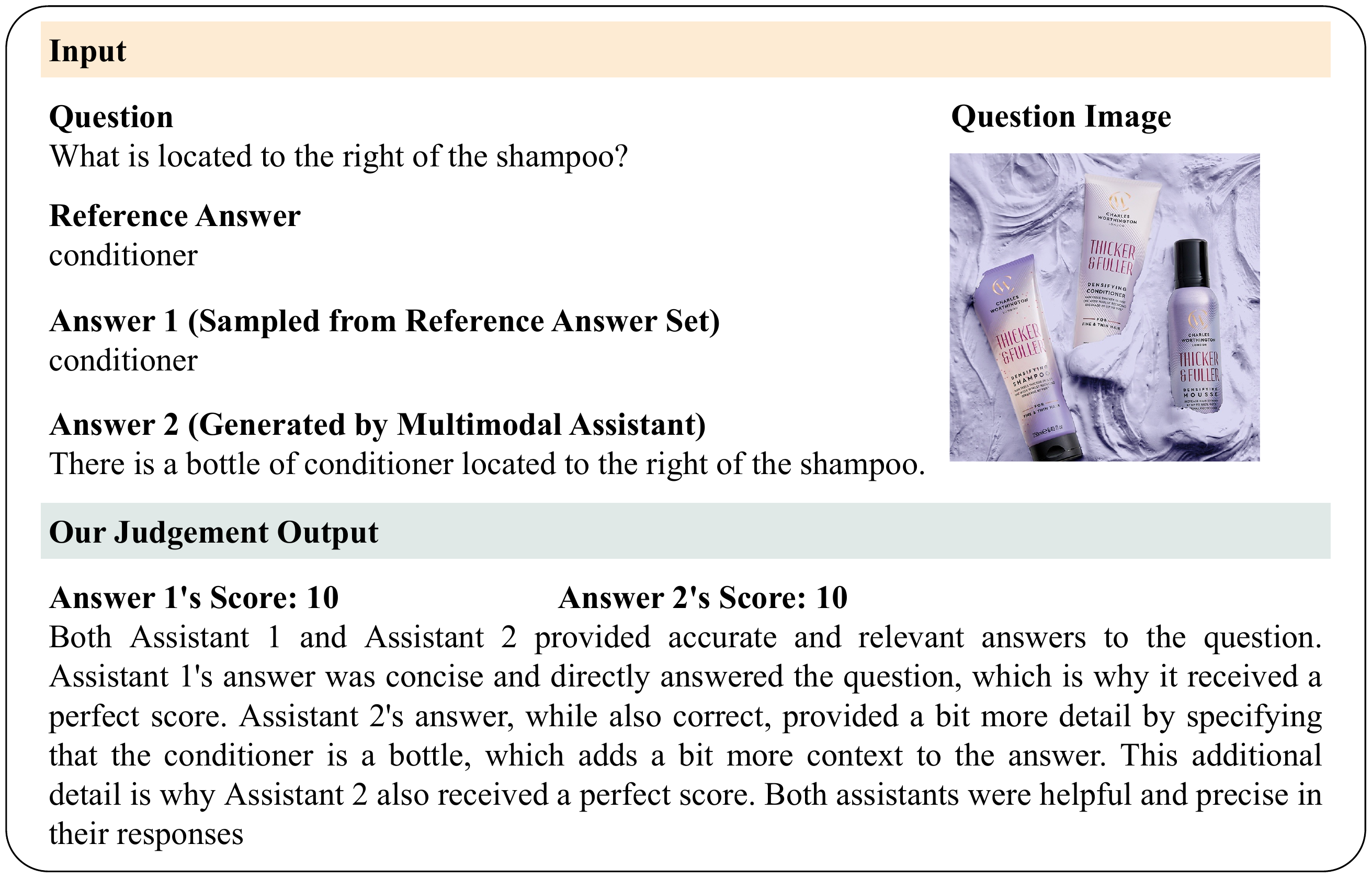}
    \end{center}
    \caption{\add{An illustration of multimodal high-score grading on MM-Vet benchmark. The proposed \name{} can replace GPT-4 to grade multimodal answers.}}
    \label{fig: mmvet-high}
\end{figure}

\begin{figure}[htp]
    \begin{center}
        \includegraphics[width=0.9\linewidth]{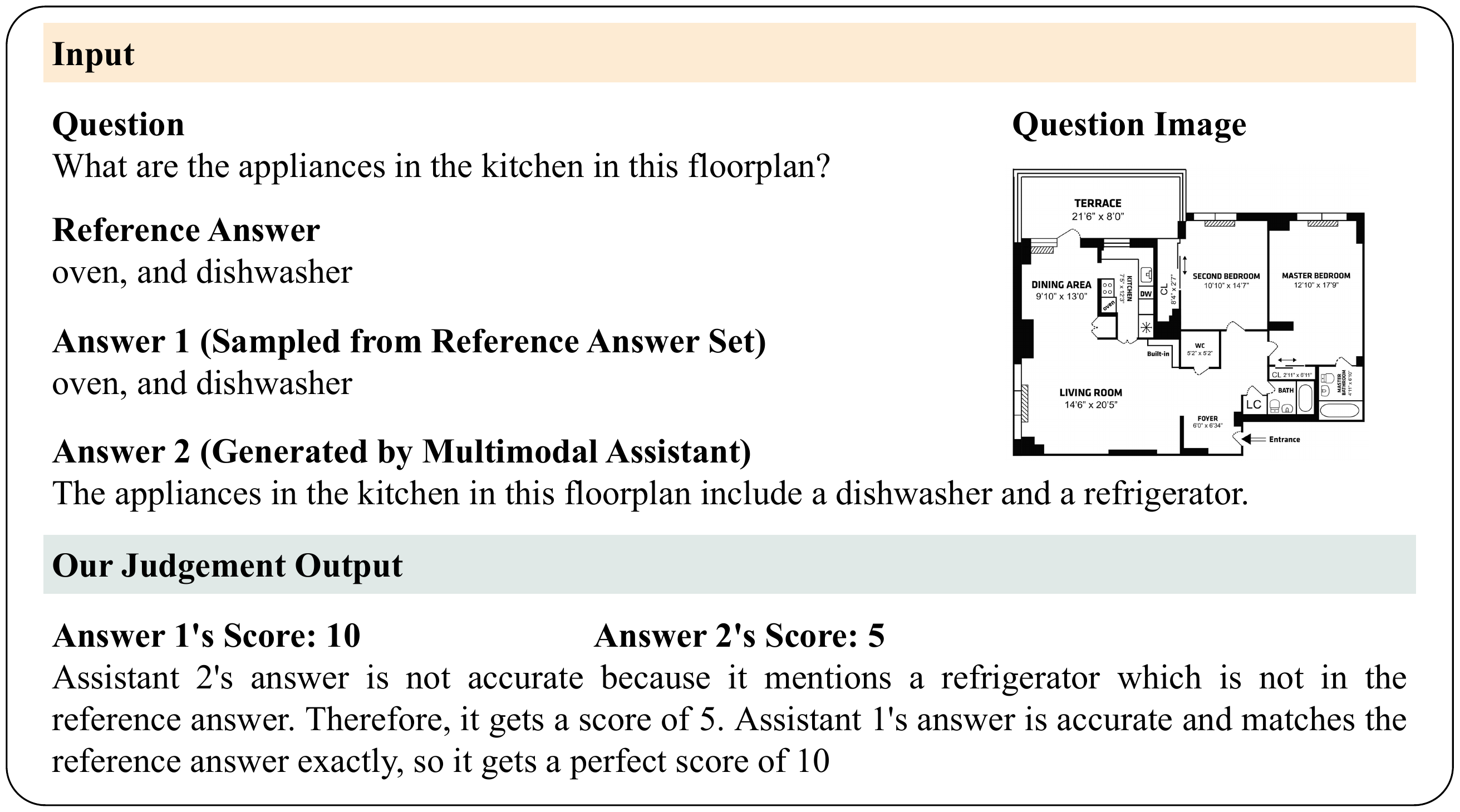}
    \end{center}
    \caption{\add{An illustration of multimodal mid-score grading on MM-Vet benchmark. The proposed \name{} can replace GPT-4 to grade multimodal answers.}}
    \label{fig: mmvet-mid}
\end{figure}

\begin{figure}[ht]
    \begin{center}
        \includegraphics[width=0.9\linewidth]{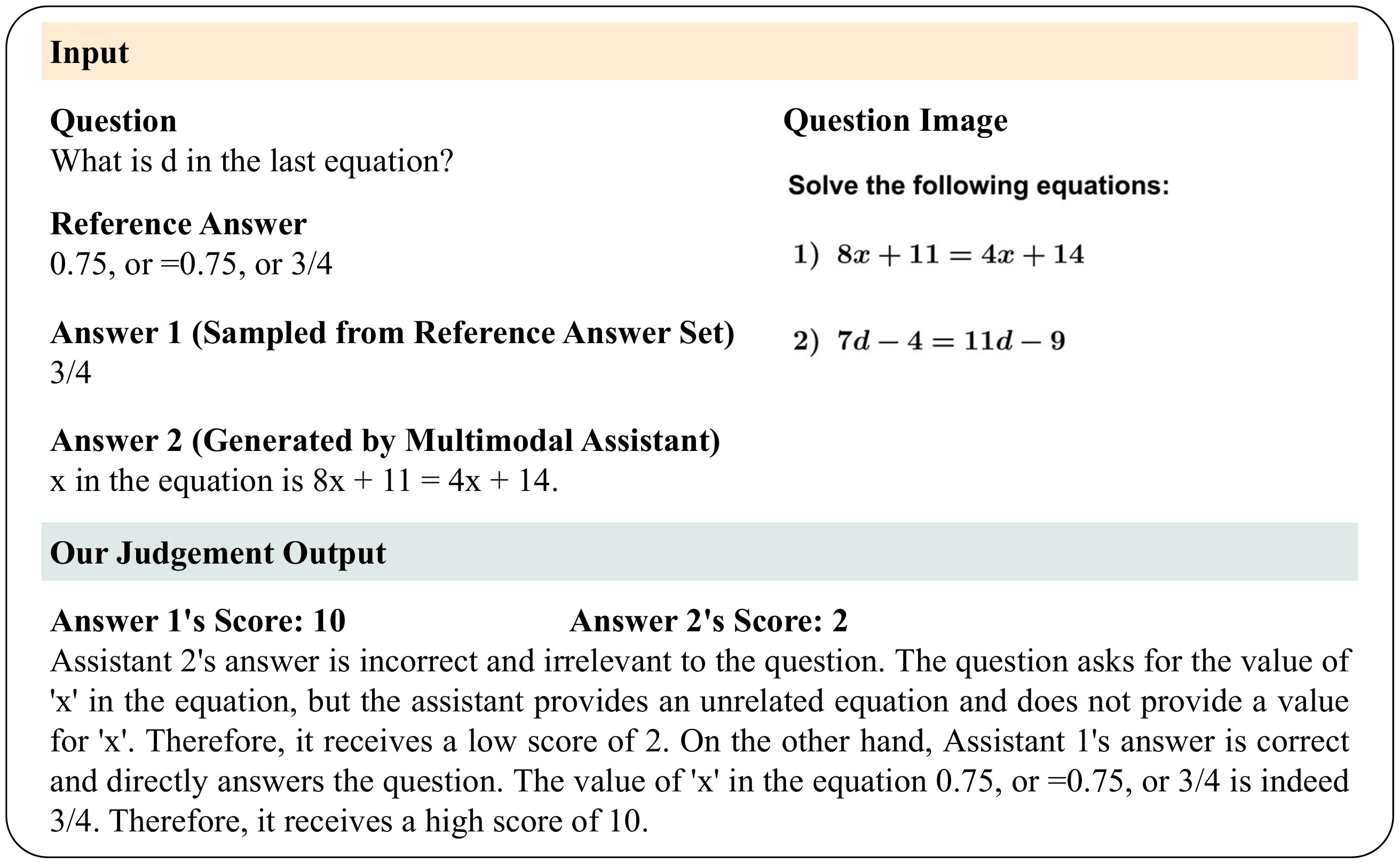}
    \end{center}
    \caption{\add{An illustration of multimodal low-score grading on MM-Vet benchmark. The proposed \name{} can replace GPT-4 to grade multimodal answers.}}
    \label{fig: mmvet-low}
\end{figure}

\end{document}